\RequirePackage{fix-cm}
\documentclass[smallextended]{svjour3}    
\smartqed 
\usepackage[margin=1in]{geometry} 
\usepackage{longtable}
\usepackage{graphicx}
\usepackage{tikz}
\usepackage{multirow}
\usepackage{{xcolor}}
\usetikzlibrary{arrows,shapes,positioning,shadows,trees}
\usepackage{threeparttable}
\usepackage{array}
\newcolumntype{C}[1]{>{\centering\arraybackslash}m{#1}}
\usepackage{amssymb}
\usepackage{pifont}
\newcommand{\cmark}{\ding{51}}%
\newcommand{\xmark}{\ding{55}}%
\usepackage{graphicx}
\usepackage{amsmath}
\usepackage{float}
\usepackage{footnote}

\AtBeginDocument{
	\setlength\abovedisplayskip{0pt}
	\setlength\belowdisplayskip{0pt}}
\tikzset{
 basic/.style = {draw, text width=2cm, drop shadow, font=\sffamily, rectangle},
 root/.style  = {basic, rounded corners=2pt, thin, align=center,
          fill=blue!30},
 level 2/.style = {basic, rounded corners=6pt, thin,align=center, fill=blue!60,
          text width=7em},
 level 3/.style = {basic, thin, align=left, fill=gray!20, text width=4em}
}
\usepackage{amssymb}
\usepackage{amssymb}
\usepackage{caption}
\usepackage{rotating}
\usepackage{tabularx}
\usepackage[labelfont=bf]{caption} 

\usepackage{ragged2e}
\usepackage{booktabs}
\usepackage{rotating}
\usepackage{lscape}
\usepackage{amsmath}
\usepackage{amssymb}
\usepackage{upquote}
\usepackage{array}
\usepackage[round]{natbib}
\usepackage[colorlinks=true,allcolors=blue]{hyperref}
\restylefloat{figure}
\journalname{Artificial Intelligence Review}

\begin{document}	
	\title{Anaphora and Coreference Resolution: A Review
	}
	
	
	\author{Rhea Sukthanker     \and
		Soujanya Poria \and
		Erik Cambria \and
		Ramkumar Thirunavukarasu
	} 
	
	\institute{Rhea Sukthanker \at
		School of Information Technology and Engineering, VIT University, Vellore, India \\
		\email{rheasukthanker@gmail.com}      
		\and
		Soujanya Poria \at
		Temasek
		Laboratories,
		Nanyang
		Technological
		University, Singapore\\
		\email{sporia@ntu.edu.sg}\and 
		Erik Cambria \at
		School of Computer Science and Engineering, Nanyang Technological University, Singapore\\
		\email{cambria@ntu.edu.sg} \and
		Ramkumar Thirunavukarasu \at
	  School of Information Technology and Engineering, VIT University, Vellore, India\\
		\email{ramkumar.thirunavukarasu@vit.ac.in}
	}
	
	\date{Received: date / Accepted: date}

	\maketitle
	
	\begin{abstract}
Entity resolution aims at resolving repeated references to an entity in a document and forms a core component of natural language processing (NLP) research. This field possesses immense potential to improve the performance of other NLP fields like machine translation, sentiment analysis, paraphrase detection, summarization, etc. The area of entity resolution in NLP has seen proliferation of research in two separate sub-areas namely: anaphora resolution and coreference resolution. Through this review article, we aim at clarifying the scope of these two tasks in entity resolution. We also carry out a detailed analysis of the datasets, evaluation metrics and research methods that have been adopted to tackle this NLP problem. This survey is motivated with the aim of providing the reader with a clear understanding of what constitutes this NLP problem and the issues that require attention.
		
	\end{abstract}
	
	\section{Introduction}
	\label{intro}
	A discourse is a collocated group of sentences which convey a clear understanding only when read together. The etymology of anaphora is $ana$ (Greek for back) and $pheri$ (Greek for to bear), which in simple terms means repetition. In computational linguistics, anaphora is typically defined as references to items mentioned earlier in the discourse or \textquotedblleft pointing back\textquotedblright{} reference as described by~\citep{mitkov1999anaphora}. The most prevalent type of anaphora in natural language is the pronominal anaphora~\citep{lappin1994algorithm}. Coreference, as the term suggests refers to words or phrases referring to a single unique entity in the world. Anaphoric and co-referent entities themselves form a subset of the broader term \textquotedblleft discourse parsing\textquotedblright{}~\citep{soricut2003sentence}, which is crucial for full text understanding.
	
	In spite of having a rich research history in the NLP community, anaphora resolution is one of the sub-fields of NLP which has seen the slowest progress thus establishing the intricacy involved in this task. Some applications of this task in NLP span crucial fields like sentiment analysis~\citep{camacsa}, summarization~\citep{steinberger2007two}, machine translation~\citep{preuss1992anaphora}, question answering~\citep{castagnola2002anaphora}, etc. Anaphora resolution can be seen as a tool to confer these fields with the ability to expand their scope from intra-sentential level to inter-sentential level.\\
	
	This paper aims at providing the reader with a coherent and holistic overview of anaphora resolution (AR) and coreference resolution€ (CR) problems in NLP. These fields have seen a consistent and steady development, starting with the earlier rule-based systems~\citep{hobbs1978resolving, lappin1994algorithm} to the recent deep learning based methodologies~\citep{wiseman2016learning, clark2016improving, clark2016deep, lee2017end, yourec}. Though there have been some thorough and intuitive surveys, the most significant ones are by~\citep{mitkov1999anaphora} for AR and~\citep{ng2010supervised} for CR. 
	
	The detailed survey on AR by~\citep{mitkov1999anaphora} provides an exhaustive overview of the syntactic constraints and important AR algorithms. It also analyzes the applications of AR in other related NLP fields. The most recent survey by~\citep{ng2010supervised} targets the research advances in the related field of CR delineating the mention-pair, entity-mention and mention-ranking models proposed till date. Both of these surveys are a great resource to gain a deeper understanding of research methodologies which have been attempted earlier for AR and CR. 
	
	The advent of neural networks in NLP has demonstrated performance strides in most of its sub-fields like POS tagging~\citep{collobert2008unified}, social data analysis~\citep{onesta}, dependency parsing~\citep{chen2014fast}, etc. and this is no different to the field of CR. Thus, this paper is fueled by the necessity for detailed analysis of state-of-the-art approaches pertaining to this field. Here, we seek to build on the earlier surveys by delineating the pioneering research methodologies proposed for these two very closely related, yet significantly different fields of research. Often the proposed methodologies differ in the evaluation metrics adopted by them, thus making comparison of their performance a major challenge. We also provide a comprehensive section on the evaluation metrics adopted, with the aim of establishing well defined standards for comparison. Another motivation factor for this survey is the requirement to establish the standard datasets and open source toolkits for researchers and off-the-shelf users, respectively.
	
	AR and CR have seen a shifting trend, from methods completely dependent on hand-crafted features to deep learning based approaches which attempt to learn feature representations and are loosely based on hand engineered features. This future trend looks very promising and, hence, we have discussed this in the comparison section. Another issue which requires to be addressed is the type of references that can occur in language and the constraints to be applied to identify the possible co-referring entities. Though state-of-the-art approaches have demonstrated a significant margin of improvement from the earlier ones, some rare types of references have gone unnoticed and, hence, demand attention. This field has faced a long history of debate with regards to comparison of different types of approaches, the appropriate metrics for evaluation, the right preprocessing tools, etc. Another topic of debate pertaining to CR is whether induction of commonsense knowledge aids the resolution process. We also aim at providing an overview of these issues and controversies. Through this survey we also aim at analyzing the application of CR in other NLP tasks with special attention to its application in sentiment analysis~\citep{chadis}, as recently this is one of the hottest topics of NLP research due to the exponential growth of social media. Finally, this survey forms building blocks for the reader to better understand this exciting field of research.
	
	\section{Types of References in Natural Language}
	\label{sec1}
	AR is a particularly challenging task because of the different forms the \textquotedblleft references\textquotedblright{} can take. Most AR and CR algorithms face the \textquotedblleft coverage\textquotedblright{} issue. This means that most algorithms are designed to target only specific types of references. Before proceeding to the current state-of-the-art research methodologies proposed for this task, it is necessary to understand the scope of this task to its entirety. In this section, we will be discussing the different semantic and syntactic forms in which the references can occur.
	\subsection{Zero Anaphora}
	\label{sec11}
	This type of anaphora is particularly common in prose and ornamental English and was first introduced by~\citep{fillmore1986pragmatically}. It is perhaps one of the most involved type of AR task which uses a gap in a phrase or a clause to refer back to its antecedent. For example, in the sentence \textquotedblleft\emph{You always have \underline{two fears}(1): \underline{your commitment}(2) versus \underline{your fear}(3)}\textquotedblright{} phrases (2) and (3) refer back (are anaphoric) to the same phrase (1). Hence, the above sentence serves as an example for a combination of zero anaphora with m-anaphors.
	\subsection{One Anaphora}
	\label{sec12}  
	In this type of anaphora, the word \textquotedblleft one\textquotedblright{} is used to refer to the antecedent. This type of anaphora, though not very common, has received sufficient attention from the research community, particularly the machine learning approach by~\citep{ng2002improving} which specifically targeted this type. The one anaphora phenomenon can be best illustrated with an example. In the sentence \textquotedblleft\emph{Since Samantha has set her eyes on \underline{the beautiful villa by the beach}(1), she just wants to buy \underline{that one} (2)}\textquotedblright{}, the phrase (2) refers back to the entity depicted by (1).
	\subsection{Demonstratives}
	\label{sec13}
	This type of reference as explained by~\citep{dixon2003demonstratives} is typically used in contexts when there is a comparison between something that has occurred earlier. For example, in the sentence \textquotedblleft\emph{This car is much more spacious and classy than \underline{that}(1)}\textquotedblright{}, phrase (1) refers to a comparison with a car that the speaker has seen earlier. This is an interesting case of AR wherein the anaphor is not specified explicitly in the text.
	\subsection{Presuppositions}
	\label{sec14}
	In this type of references, the pronouns used are someone, anybody, nobody, anyone, etc. Here, the entity resolution is complicated as there is a degree of ambiguity involved in the consideration of the noun phrases (NPs) which the referents can be corresponding to. The projection of presupposition as an AR task was first introduced by~\citep{van1992presupposition}. For example, in the sentence \textquotedblleft\emph{If there is \underline{anyone}(1) who can break the spell, it is \underline{you}(2)}\textquotedblright{}, the phrase (2) co-refers with (1). Here, the major source of ambiguity is the phrase \textquotedblleft anyone\textquotedblright.
	\subsection{Discontinuous Sets (Split anaphora)}
	\label{sec15}
	The issue of clause splitting in AR has been delineated by~\citep{mitkov2014anaphora}. In this type of anaphora, the pronoun may refer to more than one antecedents. Commonly the pronouns which refer to more than one antecedents are they, them, us, both, etc. For example, in the sentence \textquotedblleft\emph{\underline{Kathrine}(1) and \underline{Maggie}(2) love reading. \underline{They}(3) are also the members of the reader's club.}\textquotedblright{}, the pronoun (3) refers to Maggie (2) and Katherine (1) together as a single entity. Most of the prominent algorithms in anaphora and CR fail to consider this phenomenon. In this paper, we will be discussing one significantly recent research methodology which specifically focuses on this issue.
	\subsection{Contextual disambiguation}
	\label{sec16}  
	The issue of contextual disambiguation in AR has been described by~\citep{mitkov2014anaphora} in his book on AR and attempted by many like~\citep{bean2004unsupervised}. Though this problem does not semantically fit into the category of entity resolution, this issue provides an insight into how two fields in NLP, i.e., word sense disambiguation and CR serve to complement each other. For example, in the sentence \textquotedblleft\emph{The carpenter built a \underline{laminate}(1) and the dentist built \underline{one}(2) too.}\textquotedblright, though phrase (2) refers to (1) by the One Anaphora as discussed earlier, the challenge here lies in understanding how the word \textquotedblleft laminate\textquotedblright{} can actually refer to very different real world entities based on the context in which they occur, i.e., the dentist or the carpenter.
	\subsection{Pronominal anaphora}
	\label{sec17}  
	This is one of the most common and prevalent types of anaphora which occur in day-to-day speech and constitutes a significant portion of the anaphors we commonly see in web data like reviews, blog posts, etc. This type of anaphora introduced by~\citep{roberts1989modal, heim1982semantics}, has been the focus of many papers. The earliest one being the paper of~\citep{lappin1994algorithm} which aimed at pronominal resolution. There exist three types of pronominal anaphors: indefinite pronominal, definite pronominal and adjectival pronominal.
	\subsubsection{Indefinite Pronominal}
	\label{sec171}    
	In this reference type, the pronoun refers to an entity or object which is not well-defined or well-specified. An example of this type is \textquotedblleft\emph{\underline{Many}(1) of the \underline{jurors}(2) held the same opinion}\textquotedblright{}, where (2) refers back to (1), though the exact relation between the referents is ambiguous.
	\subsubsection{Definite Pronominal}
	\label{sec172} 
	This type of reference is definite since it refers to a single unique entity in the universe. For example, in the sentence
	\textquotedblleft\emph {She had seen \underline{the car}(1) which had met with an accident. \underline{It}(2) was an old white ambassador}\textquotedblright{}, pronoun (2) refers back to entity (1).
	\subsubsection{Adjectival Pronominal}
	\label{sec173}
	In this type of anaphora, there is reference to adjectival form of the entity which has occurred earlier. For example, in the sentence \textquotedblleft\emph{\underline{A kind stranger}(1) returned my wallet. \underline{Such people}(2) are rare}\textquotedblright{}, (2) refers back to (1). Thus, (1) here is an adjectival form that has been referred to by the anaphor (2). This example also serves to illustrate that adjectival noun forms can also be anaphoric.
	\subsection{Cataphora}
	\label{sec18}
	Cataphora as defined by~\citep{mitkov2002new} is said to be the opposite of anaphora. A cataphoric expression serves to point to real world entities which may succeed it. The phenomenon of cataphora is most commonly seen in \textquotedblleft poetic\textquotedblright{} English. This type is not very common in literature and, hence, most recent approaches don't focus on this issue in particular. For example, in the sentence \textquotedblleft\emph{ If \underline{she}(1) does n't show up for the examination even today, chances of \underline{Clare}(2) clearing this semester are meagre}\textquotedblright{}, (1) refers to an entity that precedes it, i.e., (2). In this paper, we will not be reviewing techniques for cataphora resolution in particular mainly because cataphora are rarely used in spoken language.
	\subsection{Inferable or Bridging Anaphora}
	\label{sec19}  
	Bridging anaphora~\citep{hou2013global} type of references in natural language are perhaps one of the most ambiguous ones. They may not explicitly seem to be pointing to an antecedent but can be said to belong to or refer to an entity mentioned at some point earlier in time. For example, in the sentence \textquotedblleft\emph{I was about to buy \underline{that exquisite dress}(1); just when I noticed a coffee stain on \underline{the lace}(2)}\textquotedblright{}, the entity that (2) refers to, though not stated explicitly, is entity (1) which can be inferred by their context.
	\section{Non-Anaphoric pronominal references}
	\label{sec2}
	A major issue which is unanimously tackled by all state-of-the-art methods is the identification and elimination of empty referents or referents which potentially do not refer back to any antecedent. The major categories of non-referential usage are: clefts, pleonastic \textquotedblleft it\textquotedblright{} and extraposition.
	\subsection{Clefts}
	\label{sec21}
	A cleft sentence is a complex sentence (one having a main clause and a dependent clause) that has a meaning that could be expressed by a simple sentence. A cleft sentence typically puts a particular constituent into focus. Clefts were introduced by~\citep{atlas1981clefts}. For example, in the sentence \textquotedblleft it\textquotedblright{} cleft is: \textquotedblleft \emph{\underline{It}(1) was Tabby who drank the milk}\textquotedblright{}, (1) does not serve to refer to an antecedent but is a potential trap for most AR systems.
	\subsection{Pleonastic \textquotedblleft It\textquotedblright}
	\label{sec22}
	This issue in AR has received a lot of attention and has been delineated by~\citep{mitkov2014anaphora}. This type of non-anaphoric referent is very common in natural language. For example, in the sentence \textquotedblleft\emph{\underline{It}(1) was raining heavily}\textquotedblright{}, (1) in spite of being a pronoun does not refer to any specific entity.
	\subsection{Extraposition}
	\label{sec23}
	Extraposition is one of the many issues in AR as described by~\citep{gundel2005pronouns}. An example of extraposition is the sentence, \textquotedblleft\emph{\underline{It} was not justified on her part to insult the waitress}\textquotedblright{}, where \textquotedblleft it\textquotedblright{} is semantically empty and serves to imply a behavioural characteristic.	
	\section{Constraints for Anaphoric resolution}
	\label{sec3}
	Most proposed approaches in AR and CR are based on some trivial syntactic and semantic constraints. Though all constraints may not be relevant for every type of referent, most methods do apply some if not all of these constraints. Syntactic approaches are solely based on these constraints and exploit them to a large extent for AR. Most statistical and machine learning approaches use these constraints in feature extraction or mention-filtering phase. Recently, there has been a growing trend towards knowledge poor AR~\citep{mitkov1998robust}. This mainly aims at reducing the level of dependency on these hand-crafted rules. Also, it is important to understand here that these constraints are not universally acceptable, i.e., some may not hold true across different languages. The constraints below are necessary but not sufficient by themselves to filter out the incorrect references. This section aims at delineating the linguistic constraints for AR.
	\subsection{Gender agreement}
	\label{sec31}  
	Any co-referring entity should agree on their gender, i.e., male, female or non-living entity. Gender is a very important constraint in AR as mentioned by~\citep{mitkov2014anaphora}. Antecedents which do not agree in terms of their gender, need not be considered further for evaluation of their correctness. This is one of the crucial constraints which serves to prune the antecedent search space to a large extent. For example, in the sentence \textquotedblleft\emph{\underline{Tom}(1) bought \underline{a puppy}(2). \underline{It }(3) is adorable}\textquotedblright{}, on application of this constraint, (1) is eliminated due to gender disagreement with (3), thus culminating in (it=puppy). The question which arises here is what happens when there are multiple antecedents satisfying gender constraint. This brings about the necessity to enforce some other syntactic and semantic constraints.
	\subsection{Number agreement}
	\label{sec32}
	An entity may corefer with another entity if and only if they agree on the basis of their singularity and plurality. This constraint is even incorporated into machine learning systems like~\citep{ng2002improving}. This constraint is necessary but not sufficient and the final resolution may be subject to the application of other constraints. For example, in the sentence \textquotedblleft\emph{ \underline{Fatima and her sisters}(1) bought \underline{groceries}(2) for \underline{the week}(3). Recently, there has been a huge hike in \underline{their}(4) prices}\textquotedblright{}, the pronominal reference (4) refers to (2) and (1). Referent (3) is successfully eliminated on the basis of number disagreement.
	\subsection{Constraints on the verbs (Selectional constraints)}
	\label{sec33}
	Human language type-casts certain verbs to certain entities. There are certain verbs which occur with only animate or living entities and some others specifically on the inanimate ones. Constraints on verbs have been exploited in many methods like~\citep{haghighi2009simple}. The sentence \textquotedblleft\emph{I sat on the \underline{tortoise}(1) with a \underline{book}(2) in my hand, assuming it to be a huge pebble and that's when \underline{it}(3) wiggled}\textquotedblright{}, for example is very difficult for a computer to interpret. Here, (3) can refer to (2) or (1). The reference (2) should be filtered out here using the animacy constraint. This constraint brings about the necessity to incorporate world knowledge into the system.
\subsection{Person agreement}
	\label{sec34}
	Linguistics has three persons namely the first (i.e., I, me), second (i.e., you) and third (i.e., he, him, she, it, they). This feature has been exploited by many approaches like~\citep{lappin1994algorithm}. The co-referent nouns or entities must agree with respect to their person. For example, in the sentence \textquotedblleft\emph{\underline{John and Sally}(1) are siblings. It's amazing how significantly different \underline{they}(2) are from each other}\textquotedblright{}, (2) refers to (1) as they agree with respect to their person. In case the pronoun (2) had been \textquotedblleft we\textquotedblright{} this possible antecedent would have been eliminated.
	\subsection{Grammatical role}
	\label{sec35}
	Any given sentence can be decomposed to its subject, verb and object part and these roles of the words in the sentence can aid AR as mentioned by~\citep{kennedy1996anaphora}. Entities occurring in the subject portion of a sentence are given a higher priority than the entity in object position. For example, in the sentence \textquotedblleft\emph{\underline{Kavita}(1) loves shopping. She goes shopping with \underline{her sister}(2) every weekend. \underline{She}(3) often buys stuff that she may never use}, (3) refers to (1) and not to (2) as (1) being the subject has more priority or salience over (2).
	\subsection{Recency}
	\label{sec36}
	As mentioned in~\citep{carbonell1988anaphora} recency is an important factor of consideration in AR. Entities introduced recently have more salience than entities which have occurred earlier in the discourse. For example, in the sentence \textquotedblleft\emph{I have two dogs. \underline{Steve}(1), a grey hound, is a guard dog. \underline{Bruno}(2) who is a Labrador is pampered and lazy. Sally often takes \underline{him}(3) for a stroll}\textquotedblright{, (3) may refer to (1) or (2) syntactically. To resolve this ambiguity this constraint gives more salience to (2) over (3) due to the entity's recency.
	\subsection{Repeated mention}
	\label{sec37}
	Repeated mention forms a feature of many system like the statistical method of~\citep{ge1998statistical}. Entities which have been introduced repeatedly in the context or have been the main focus or topic of the earlier discourse are given a higher priority than the rest. For example, in the sentence \textquotedblleft\emph{\underline{Katherine}(1) is an orthopaedic surgeon. Yesterday she ran into \underline{a patient}(2), she had not been in contact with her since ages. \underline{She}(3) was amazed at her speedy recovery}\textquotedblright{}, the referent (3) refers to (1) and not (2) because Katherine (1) here is an entity that has been in focus in prior discourse and, hence, is more salient.
	\subsection{Discourse structure}
	\label{sec38}
	The preference of one entity over another can also be due to the structural idiosyncrasies of the discourse like parallelism. These phenomenon are discussed by~\citep{carbonell1988anaphora} in their paper and form a crucial component of Centering Theory. For example, in the sentence \textquotedblleft\emph{\underline{Aryan}(1) passed a note to \underline{Shibu}(2) and \underline{Josh}(3) gave \underline{him }(4) a chocolate}\textquotedblright{}, (4) refers to (2) and not (1) due to the discourse structure involved here. Though the occurrence of this type of discourse is tough to spot and disambiguate, if exploited appropriately this can increase the precision factor involved in the CR to a large extent.
	\subsection{World Knowledge}
	\label{sec39}
	This is the constraint that has a very wide scope and generally cannot be completely incorporated into any system. In spite of this, attempts to incorporate this behavior in CR system has been made by~\citep{rahman2011coreference}. Though syntax does play a role in entity resolution, to some extent world knowledge or commonsense knowledge does function as a critical indicator. Commonsense concepts like \textquotedblleft cat-meows\textquotedblright{} and \textquotedblleft dog-barks\textquotedblright{} cannot be resolved only by studying the syntactic properties. One obvious example where syntax by itself fails to identify the appropriate antecedent are the sentences\\ \emph{The city councilmen (1) refused the demonstrators (2) a permit because they (3) advocated violence.}\\ 
	\emph{The city councilmen (1) refused the demonstrators (2) a permit because they (3) feared violence.}
	
	As cited by some researchers~\citep{levesque2011winograd}, here world knowledge needs to be incorporated to disambiguate \textquotedblleft they\textquotedblright{}. In the first sentence (3) refers to (2) and in the second sentence (3) refers to (1). What results in this transition here is only the change of the verb involved in the discourse.
	\section {Evaluation metrics in CR}
	\label{sec4}
	There are a number of metrics which have been proposed for the evaluation of CR task. Here, we delineate the standard metrics used to evaluate the task.
	\subsection{Bagga and Baldwin's B-cubed metric}
	\label{sec41}
	This metric proposed by~\citep{bagga1998algorithms} begins by computing a precision and recall for each individual mention and, hence, takes weighted sum of these individual precision and recalls. Greedy matching is undertaken for evaluation of chain-chain pairings.
	\begin{equation}
	FinalPrecision=\sum_{i=1}^{N}w_{i}*Precision
	\end{equation}
	\begin{equation}
	FinalRecall=\sum_{i=1}^{N}w_{i}*Recall
	\end{equation}
	Where N= Number of entities in the document and $w_{i}$ is the weight assigned to entity i in the document. Usually the weights are assigned to $1/N$.  
	\subsection{MUC- Link based F-measure}
	\label{sec42}
	This metric, proposed during the 6th Message Understanding Conference by~\citep{vilain1995model} considers a cluster of references as linked references, wherein each reference is linked to at most two other references. MUC metric primarily measures the number of link modifications required to make the result-set identical to the truth-set. 
	\begin{equation}
	partition(c,s)=\mbox{\{}s|s \in S \mbox{ \& } s\in c\neq\phi\mbox{\}}
	\end{equation}
	The MUC Precision value is calculated as follows:
	\begin{equation}
	MUCPrecision(T,R)=\sum_{r\in R} \frac{|r|-|partition(r,T)|}{|r|-1}
	\end{equation}
	Where, $|partition(r,T)|$ is the number of clusters within truth T that the recall cluster \emph{r} intersects with.
	The MUC Recall value is calculated as follows:
	\begin{equation}
	MUCRecall(T,R)=\sum_{t\in T}\frac{|t|-|partition(t,R)|}{|t|-1}
	\end{equation}
	Where, $|partition(t,R)|$ represents the number of clusters within the result R that truth set $|t|$ intersects with.
	\subsection{CEAF: Constrained Entity Alignment F-measure}
	\label{sec43}
	This metric proposed by~\citep{luo2005coreference} is used for entity-based similarity identification. It uses similarity measures to first create an optimal mapping between result clusters and truth clusters. Using this mapping, CEAF leverages self-similarity to calculate the precision and recall. This similarity measure is computed using the following equations:
	\begin{equation}
	\phi_{1}(T,R)=
	\begin{cases}
	1 & \mbox{if R=T} \\
	0 & \mbox{otherwise}
	\end{cases}\end{equation}
	\begin{equation}
	\phi_{2}(T,R)=
	\begin{cases}
	1, & \mbox{if R}\cap T\neq\phi \\
	0, & \mbox{otherwise}
	\end{cases}\end{equation}
	\begin{equation}
	\phi_{3}(T,R)=|R\cap T|\end{equation}
	\begin{equation}
	\phi_{4}(T,R)=2.\frac{|R\cap T|}{|R|+|T|}
	\end{equation}
	The function $m(r)$ takes in a cluster \emph{r} and returns the true cluster \emph{t} that the result cluster \emph{r} is mapped to with constraint that one cluster can be mapped to at most one result cluster.
	\begin{equation}
	CEAF_{\phi_{i}}Precision(T,R)=max_{m}\frac{\sum_{r\in R}\phi_{i}(r,m(r))}{\sum_{r\in R}\phi_{i}(r,r)}
	\end{equation}  
	\begin{equation}
	CEAF_{\phi_{i}}Recall(T,R)=max_{m}\frac{\sum_{r\in R}\phi_{i}(r,m(r))}{\sum_{r\in R}\phi_{i}(t,t)}
	\end{equation}
	\subsection{ACE-Value}
	\label{sec44}
	The ACE evaluation score~\citep{doddington2004automatic}, proposed during the Automatic Content Extraction Conference is also based on optimal matching between the result and the truth like CEAF. The difference between the two is that ACE's precision and recall is calculated using true positives, false positives, false negatives amongst the predicted co-referent entities. Another difference is that ACE does not normalize its precision and recall values unlike the CEAF metric.
	\subsection{CoNLL score}
	\label{sec45}
	This score is calculated as the average of the B-cubed score, MUC score and the CEAF score. This is the score used by the CoNLL-2012 shared task by~\citep{pradhan2012conll} which is based on CR in the OntoNotes corpus. Thus, the CoNLL score is calculated as shown in the equation below.
	\begin{equation}
	CoNLL=\frac{(MUC_{F1} + \mbox{B-Cubed}_{F1} + CEAF_{F1})}{3}
	\end{equation}
  \subsection{BLANC metric}
  \label{sec46}
   BLANC~\citep{recasens2010semeval, pradhan2014scoring} is a link-based metric that adapts the Rand index~\citep{rand1971objective} CR evaluation. Given that $C_k$ is the key entity set and $C_r$ is the response entity set, the BLANC Precision and Recall is calculated as follows. $R_{c}$ and $R_{n}$ refer to recall for coreference links and non-coreference links, respectively. Precision is also defined with a similar notation.
\begin{equation}
R_{c}=\frac{|C_{k}\cap C_{r}|}{|C_{k}|} 
\end{equation}
\begin{equation}
 P_{c}=\frac{|C_{k}\cap C_{r}|}{|C_{r}|}
\end{equation}
\begin{equation}
R_{n}=\frac{|N_{k}\cap N_{r}|}{|N_{k}|} 
\end{equation}
\begin{equation}
P_{c}=\frac{|N_{k}\cap N_{r}|}{|N_{r}|}
\end{equation}
\begin{equation}
Recall=\frac{R_{c}+R_{n}}{2}
\end{equation}
\begin{equation}
Precision=\frac{P_{c}+P_{n}}{2}
\end{equation}
The mention identification effect delineated by (Moosavi and Strube 2016) affects BLANC metric very strongly and, hence, this metric is not widely adopted.
\subsection{LEA metric}
\label{sec47}
Link-based Entity-Aware (LEA) metric proposed by~\citep{moosavi2016coreference} aims at overcoming the mention identification effect of the earlier coreference evaluation metrics which makes it impossible to interpret the results properly. LEA considers how important the entity is and how well is it resolved. LEA metric is dependent on two important terminologies which are \emph{importance} and \emph{resolution-score}. Importance is dependent on size of the entity and Resolution-Score is calculated using link similarity. The link function returns the total number of possible links between \emph{n} mentions of an entity \emph{e}. 
\begin{equation}
\text{importance}(e_{i})=|e_{i}|
\end{equation}
\begin{equation}
\text{resolution-score}(k_{i})=\sum_{r_j \in R}\frac{link(k_{i}\cap r_{j})}{link(k_{i})}
\end{equation}
\begin{equation}
Recall=\frac{\sum_{k_{i}\in K}importance(k_{i})* \sum_{r_j \in R}\frac{link(k_{i}\cap r_{j})}{link(k_{i})}}{\sum_{k_{z}\in K}importance(k_{z})}
\end{equation}
\begin{equation}
Precision=\frac{\sum_{r_{i}\in R}importance(r_{i})* \sum_{k_{j} \in K}\frac{link(r_{i}\cap k_{j})}{link(r_{i})}}{\sum_{r_{z}\in R}importance(r_{z})}
\end{equation}
In the above equations, $r_{i}$ represents the result set and $k_{i}$ represents the key set or the gold set.
	\subsection{Comparison of evaluation metrics}
	\label{sec48}
	The MUC-score which was one of the earliest metric to be proposed for CR has some drawbacks as pointed out by~\citep{luo2005coreference}. Being link based MUC score ignores singleton-mention entities, since no link can be found in the entities. It also fails to distinguish the different qualities of system outputs and favors system producing fewer entities. Thus, in some cases MUC-score may result in higher F-measure for worse systems. B-cubed metric which was MUC-score's successor aimed at fixing some of the problems associated with the metric. However, an important issue associated with B-cubed is that it is calculated by comparing entities containing the mention and, hence, an entity can be used more than once. The BLANC metric~\citep{recasens2011blanc} is also quite flawed because it considers the non-coreference links which increase with increase in gold mentions, thus giving rise to the mention identification effect. ACE metric which is very closely related to CEAF metric is not very interpretable. CEAF metric solves the interpretability issue of the ACE-metric and the drawbacks of MUC F1 score and B-cubed F1 score. However, CEAF metric has problems of its own too. As mentioned by~\citep{denis2009global} CEAF ignores all correct decisions of unaligned response entities that may lead to un-reliable results. A recent paper which particularly targets this flaw~\citep{moosavi2016coreference} discusses the issues with the existing metrics and proposes a new link aware metric (LEA metric). The figure below represents the different types of metrics proposed till date.

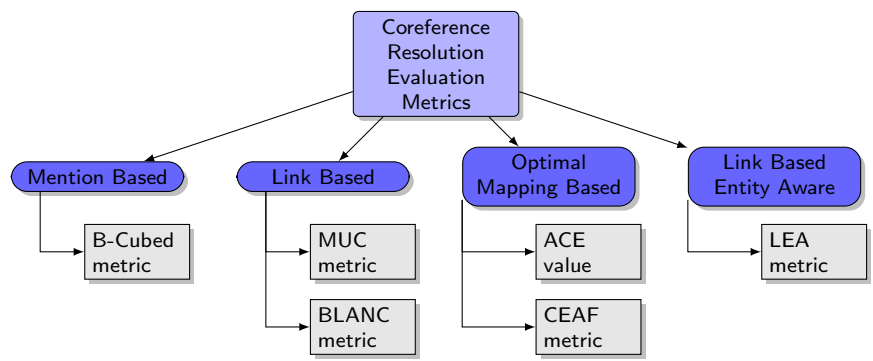
\begin{figure}[H]
	\centering
\begin{tikzpicture}[
 level 1/.style={sibling distance=30
mm},
 edge from parent/.style={->,draw},
 >=latex]

\node[root] {Coreference Resolution Evaluation Metrics}
 child {node[level 2] (c1) {Mention Based}}
 child {node[level 2] (c2) {Link Based}}
 child {node[level 2] (c3) {Optimal Mapping Based}}
 child {node[level 2] (c4) {Link Based Entity Aware}};
\begin{scope}[every node/.style={level 3}]
\node [below of = c1, xshift=15pt] (c11) {B-Cubed metric};

\node [below of = c2, xshift=15pt] (c21) {MUC metric};
\node [below of = c21] (c22) {BLANC metric};

\node [below of = c3, xshift=15pt] (c31) {ACE value};
\node [below of = c31] (c32) {CEAF metric};
\node [below of = c4, xshift=15pt] (c41) {LEA metric};
\end{scope}
\foreach \value in {1}
 \draw[->] (c1.195) |- (c1\value.west);

\foreach \value in {1,2}
 \draw[->] (c2.195) |- (c2\value.west);

\foreach \value in {1,2}
 \draw[->] (c3.195) |- (c3\value.west);

\foreach \value in {1}
 \draw[->] (c4.195) |- (c4\value.west);
\end{tikzpicture}
\caption{Evaluation Metrics}
\label{fig:1}
\end{figure}
	\section{Comparison between Anaphora and Coreference Resolution}
	\label{sec7}
	AR is an intra−linguistic terminology, which means that it refers to resolving entities used within the text with a same "sense". Also, these entities are usually present in the text and, hence, the need of world-knowledge is minimal. CR, on the other hand, has a much broader scope and is an extra-linguistic terminology. Co-referential terms may have completely different \textquotedblleft senses\textquotedblright{} and yet, by definition, they refer to the same extra linguistic entity. Coreference treats entities in a way more similar to how we understand discourse, i.e., by treating each entity as a unique entity in real time.
	
	The above explanation elicits that AR is a subset of CR. However, this claim though commonly made fails in some cases as stated by~\citep{mitkov2001outstanding} in his example: \emph{Every speaker had to present his paper}. Here, if \textquotedblleft his\textquotedblright{} and \textquotedblleft every speaker\textquotedblright{} are said to co-refer (i.e., considered the same entity), the sentence is interpreted as \textquotedblleft Every speaker had to present Every speaker's paper\textquotedblright{} which is obviously not correct. Thus, \textquotedblleft his\textquotedblright{} here is an anaphoric referent and not coreferential, hence demarcating the two very similar but significantly different concepts. This is a typical case of the bound variable problem in entity resolution. Hence, the often made claim that AR is a type of CR, fails in this case.
	
	Some researchers also claim that coreference is a type of AR. However, this can often be seen as a misnomer of the term \textquotedblleft anaphora\textquotedblright{}, which clearly refers to something that has occurred earlier in the discourse. CR, on the other hand, spans many fields like AR, cataphora resolution, split antecedent resolution, etc. For example: \emph{If \underline{he}(1) is unhappy with your work, \underline{the CEO}(2) will fire you}. Here, the first reference is not anaphoric as it does not have any antecedent, but (1) is clearly coreferent with (2). What we see here is the occurrence of the "cataphora" phenomenon. Thus, this claim too fails to capture these phenomenon adequately. Though these two concepts have a significant degree of overlap, they are very different and can be represented by the chart below.
	\begin{figure}[H]
		\centering
		\includegraphics[width=0.6\linewidth,height=8.4cm]{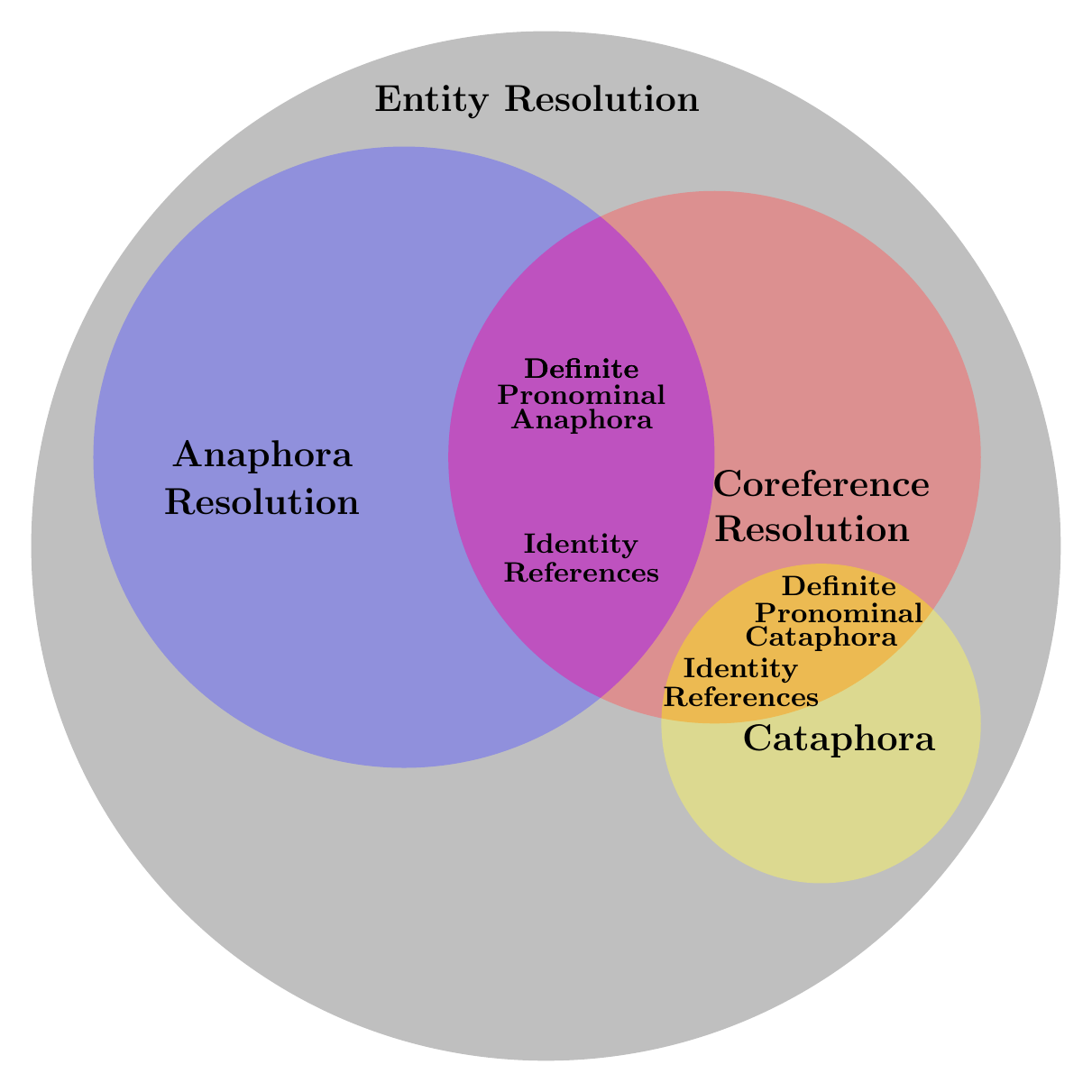}
		\caption{Vein Diagram of Entity Resolution}
		\label{fig:2}
	\end{figure}
	There is a clear need for redefinition of the CR problem. We find that the standard datasets like CoNLL 2012~\citep{pradhan2012conll} fail to capture the problem to its entirety. To address the fuzziness involved in the terminologies used in entity resolution, we suggest that the datasets created for the task explicitly specify the coreference type they have considered for annotation and the ones they have not. We also insist that future entity resolution (CR, AR, etc.) models also perform exhaustive error analysis and clearly state the types of references which their algorithms fail to resolve. This will serve two purposes: first it will help the future researchers focus their efforts on specific types of references like co-referent event resolution which most models fail to resolve and secondly, this will also help surface some clear issues in the way we currently define the task
\section{Coreference and Anaphora Resolution Datasets}
\label{sec5}
The datasets predominantly used for the task of CR differ based on a number of factors like their domain, their annotation schemes, the types of references which are labelled, etc. Thus, it is crucial to develop a clear understanding of the AR and CR datasets before proceeding to the research methodologies pertaining to the task which use these datasets either for training or for deriving effective rules. Every new dataset for AR and CR was introduced with the aim of addressing the issues with the earlier ones.

Though there are myriad datasets available for this task there are some major ones which have been widely popular for evaluation purposes. The three main corpora created targeting this task were the MUC, the ACE and the OntoNotes corpora. The MUC-6~\citep{grishman1996message} and MUC-7~\citep{chinchor1998overview} have been typically prepared by human annotators for training, dry run text and formal run test usage. The MUC datasets which were the first corpora of any size for CR, are now hosted by Linguistic Data Consortium. These dataset contains 318 annotated Wall Street Journal (WSJ) Articles mainly based on North American news corpora. The co-referring entities are tagged using SGML tagging based on the MUC format. The evaluation on this dataset is carried out using the MUC scoring metric. Now that larger resources containing multi-genre documents are available these datasets are not widely used any more except for comparison with baselines. The ACE corpus~\citep{doddington2004automatic} which was the result of series of evaluations from 2000 to 2008 is labelled for different languages like English, Chinese and Arabic. Though initial version of this corpus was based on news-wire articles like MUC the later versions also included broadcast conversations, web-log, UseNet and conversational telephonic speech. Thus, this dataset is not genre specific and is heterogeneous. ACE is mainly annotated for pronominal AR and is evaluated using the ACE-score metric.

 The Task-1 of SemEval 2010 defined by~\citep{recasens2010semeval} was CR. This dataset is made freely available for the research community. The annotation format of the SemEval Dataset is similar to the CoNLL dataset and it is derived from the OntoNotes 2.0 corpus. SemEval 2010 CR task can be seen as a predecessor of the CoNLL-2012 shared task. The CoNLL 2012 shared task~\citep{pradhan2012conll} targeted the modeling of CR for multiple languages. It aimed at classifying mentions into equivalence classes based on the entity they refer to. This task was based on the OntoNotes 5.0 dataset which mainly contains news corpora. This dataset has been widely used recently and is freely available for research purposes.
 
Unlike the datasets discussed earlier, the GNOME corpus by~\citep{poesio2004discourse} contains annotated text from the museum, pharmaceutical and tutorial dialogue domains and, hence, is very useful for cross-domain evaluation of AR and CR algorithms. Since this corpus was mainly developed for study of centering theory, it focuses on \textquotedblleft utterance\textquotedblright{} labelling. The GNOME corpus is not freely distributed. The Anaphora Resolution and Underspecification (ARRAU) corpus~\citep{poesio2008anaphoric} is a corpus labelled for anaphoric entities and maintained by LDC. This corpus is a combination of TRAINS~\citep{gross1993trains, heeman1995trains}, English Pear~\citep{watson1981pear}, RST~\citep{carlson2003building} and GNOME datasets. It is labelled for multi-antecedent anaphora, abstract anaphora, events, actions and plans. It is labelled using the MMAX2 format which uses hierarchical XML files for each document, sentence and markable. This is a multi-genre corpora, based on news corpora, task oriented dialogues, fiction, etc. The ARRAU guidelines were also adapted for the LIVEMEMORIES corpus~\citep{recasens2010ancora} for AR in the Italian Language.

In addition to the above corpora, there were some corpora which were created for task-specific CR. The ParCor dataset by~\citep{guillou2014parcor} is mainly for parallel pronoun CR across multiple languages. It is based on the genre of TEDx talks and Bookshop publications. The corpora is annotated using the MMAX2 format. This parallel corpus is available in two languages German and English and mainly aims at CR for machine translation.
The Character Identification Corpus is a unique corpus by~\citep{chen2016character} which contains multi-party conversations (TV show transcripts) labelled with their speakers. This dataset is freely available for research and is annotated using the popular CoNLL format. This dataset like GNOME is extremely useful for cross-domain evaluation. This corpus was introduced mainly for the task of character-linking in multi-party conversations. The task of Event CR has also received significant amount of attention. The NP4E corpora~\citep{hasler2009coreferential} is labelled for corefering events in the texts. This corpus is based on terrorism and security genres and is annotated in the MMAX2 format. Another event coreference dataset is the Event Coreference Bank (ECB$+$)~\citep{cybulska2014guidelines} dataset for topic-based event CR. The dataset is available for download and is annotated according to the ECB$+$ format.

The GUM corpus~\citep{zeldes2017gum} is another open source multilayer corpus of richly annotated web texts. It contains conversational, instructional and news texts. It is annotated in the CoNLL format.
The WikiCoref dataset~\citep{ghaddar2016wikicoref}, maps the CR task to Wikipedia. The dataset mainly consists of 30 annotated Wikipedia articles. Each annotated entity also provides links to FreeBase knowledge repository for the mentions. The dataset is annotated in OntoNotes schema using the MaxNet tagger and is freely available for download. GUM and WikiCoref are both mainly based on Wikipedia data. These datasets aimed to address two main issues in CR datasets: domain adaptation and world knowledge induction.

In this article, our main focus is the study of the CR task in English, but it is very interesting to note that there are datasets available to address this issue in multiple languages. The ACE corpora~\citep{doddington2004automatic} and the CoNLL-2012~\citep{pradhan2012conll} shared task in addition to English are also labelled for the Chinese and Arabic languages. The SemEval 2010 Task-1~\citep{recasens2010semeval} also provides datasets for CR in Catalan, Dutch, German, Italian and Spanish languages. The ParCor~\citep{guillou2014parcor} corpus is also labelled for German language. The AnCora-Co~\citep{recasens2010ancora} corpus has also been labelled for coreference in Spanish and Catalan.
\begin{table}[H]
	\centering
	\caption{AR and CR datasets Comparison}
	\label{my-label}
	\setlength{\tabcolsep}{15pt}
	\renewcommand{\arraystretch}{2.5}
	\begin{tabular}{lllll}
		\hline
		Dataset & \multicolumn{1}{l}{Multi-lingual} & \multicolumn{1}{l}{Multi-Domain} & \multicolumn{1}{l}{\begin{tabular}[l]{@{}c@{}}Intra-Document \\  Annotation\end{tabular}} & \multicolumn{1}{l}{\begin{tabular}[l]{@{}c@{}}Inter-Document \\  Annotation\end{tabular}} \\ \hline
		CoNLL-2012 & \cmark & \cmark & \cmark & \xmark \\ 
		ECB+ & \xmark & \xmark & \cmark & \cmark \\ 
		SemEval 2010 & \cmark & \cmark & \cmark & \xmark \\ 
		ARRAU & \xmark & \cmark & \cmark & \xmark \\ 
		CIC & \xmark & \xmark & \cmark & \xmark \\ 
		MUC 6 \& 7 & \xmark & \xmark & \cmark & \xmark \\ 
		ParCor & \cmark & \cmark & \cmark & \xmark \\ 
		GUM & \xmark & \cmark & \cmark & \xmark \\ 
		NP4E & \xmark & \xmark & \cmark & \cmark \\ 
		ACE & \cmark & \cmark & \cmark & \xmark \\ 
		WikiCoref & \xmark & \cmark & \cmark & \xmark \\ 
		GNOME & \xmark & \cmark & \cmark & \xmark \\ \bottomrule
	\end{tabular}
\end{table}
	\begin{landscape}
 	
 	\setlength\tabcolsep{1pt}
 	\footnotesize

\begin{table}[htbp]
	\centering
	\caption{English Coreference and Anaphora Resolution Datasets}
	\label{table:1}
	\begin{tabular}{llllll}
		\toprule
		Dataset & Source Corpora & Statistics & Genre & Annotation Scheme & Availability \\
		\hline 
		\multicolumn{1}{l}{CoNLL-2012~\citep{pradhan2012conll}} & \multicolumn{1}{l}{OntoNotes 5.0 Corpus} & \multicolumn{1}{l}{\begin{tabular}[l]{@{}l@{}}Train docs:2802\\ Test docs:348\\ Dev docs:343\\ Total docs: 3493\end{tabular}} & \multicolumn{1}{l}{\begin{tabular}[l]{@{}l@{}}News, conversational\\ telephone speech, \\ web-logs, UseNet \\ newsgroups, talk shows\end{tabular}} & \multicolumn{1}{l}{CoNLL format} & \multicolumn{1}{l}{\begin{tabular}[l]{@{}l@{}}Freely available\\ through LDC\end{tabular}} \\ 
		\multicolumn{1}{l}{ECB+~\citep{cybulska2014guidelines}} & \multicolumn{1}{l}{Google News} & \multicolumn{1}{l}{Total docs:982} & \multicolumn{1}{l}{News} & \multicolumn{1}{l}{ECB+ format} & \multicolumn{1}{l}{Freely available} \\ 
		\multicolumn{1}{l}{SemEval 2010~\citep{recasens2010semeval}} & \multicolumn{1}{l}{English: OntoNotes 2.0} & \multicolumn{1}{l}{\begin{tabular}[l]{@{}l@{}}Train docs:229\\ Test docs:85\\ Dev docs:39\end{tabular}} & \multicolumn{1}{l}{\begin{tabular}[l]{@{}l@{}}News, conversational \\ telephone speech, web-logs, \\ UseNet newsgroups, \\ talk shows\end{tabular}} & \multicolumn{1}{l}{CoNLL format} & \multicolumn{1}{l}{\begin{tabular}[l]{@{}l@{}}Freely available \\ through LDC\end{tabular}} \\ 
		\multicolumn{1}{l}{ARRAU 2~\citep{poesio2008anaphoric}} & \multicolumn{1}{l}{\begin{tabular}[l]{@{}l@{}}TRAINS, English Pear,\\ RST, GNOME\end{tabular}} & \multicolumn{1}{l}{Total docs:552} & \multicolumn{1}{l}{\begin{tabular}[l]{@{}l@{}}News (RST), \\ task-oriented dialogues \\ (TRAINS), fiction \\(PEAR) and medical\\ leaflets (GNOME)\end{tabular}} & \multicolumn{1}{l}{MMAX2 format} & \multicolumn{1}{l}{\begin{tabular}[l]{@{}l@{}}Available by payment \\ through LDC\end{tabular}} \\ 
		\multicolumn{1}{l}{\begin{tabular}[l]{@{}l@{}}CIC~\citep{chen2016character}\end{tabular}} & \multicolumn{1}{l}{\begin{tabular}[l]{@{}l@{}}Dialogue Scripts of \\Friends TV Show (Season 1\\ and 2), and The Big Bang\\ Theory TV Show (Season 1)\end{tabular}} & \multicolumn{1}{l}{\begin{tabular}[l]{@{}l@{}}Train docs(Episodes+ Scenes): 478\\ Dev docs(Episodes+ Scenes):51 \\ Test docs(Episodes+Scenes):77\end{tabular}} & \multicolumn{1}{l}{TV Show Dialogues} & \multicolumn{1}{l}{CoNLL format} & \multicolumn{1}{l}{\begin{tabular}[l]{@{}l@{}}Freely available \\ for download\end{tabular}} \\
		\multicolumn{1}{l}{MUC 6 \& 7~\citep{grishman1996message}} & \multicolumn{1}{l}{\begin{tabular}[l]{@{}l@{}}MUC 6:WSJ corpus, \\ MUC 7: NY Times Corpu\end{tabular}} & \multicolumn{1}{l}{\begin{tabular}[l]{@{}l@{}}MUC 6-Train docs:30, Test docs:30,\\ Total docs:60\\ MUC 7-Train docs:30, Test docs:20,\\ Total docs:50\end{tabular}} & \multicolumn{1}{l}{News} & \multicolumn{1}{l}{\begin{tabular}[l]{@{}l@{}}MUC SGML tagging\\ format\end{tabular}} & \multicolumn{1}{l}{\begin{tabular}[l]{@{}l@{}}Available by payment\\ through LDC\end{tabular}} \\ 
		\multicolumn{1}{l}{ParCor~\citep{guillou2014parcor}} & \multicolumn{1}{l}{Multilingual SMT Corpora} & \multicolumn{1}{l}{Total docs:19} & \multicolumn{1}{l}{\begin{tabular}[l]{@{}l@{}}EU Bookshop and TED Talks\end{tabular}} & \multicolumn{1}{l}{MMAX2} & \multicolumn{1}{l}{Freely Available} \\ 
		\multicolumn{1}{l}{GUM~\citep{zeldes2017gum}} & \multicolumn{1}{l}{\begin{tabular}[l]{@{}l@{}}Wikinews, WikiHow,\\ WikiVoyage, Reddit,\\ Wikipedia\end{tabular}} & \multicolumn{1}{l}{Total docs:101} & \multicolumn{1}{l}{\begin{tabular}[l]{@{}l@{}}News (narrative) \\ Interview (conversational) \\ How-to (instructional) \\ Travel guide (informative),\\ Academic Writing,\\ Biographies, Fiction,\\ Forum Discussions\end{tabular}} & \multicolumn{1}{l}{\begin{tabular}[l]{@{}l@{}}Richly annotated \\ with multiple layers \\ of annotation like\\ RST, CoNLL, \\WebAnno, ISO \\date/time,\\ Dependencies etc\end{tabular}} & \multicolumn{1}{l}{Freely Available} \\ 
		\multicolumn{1}{l}{NP4E~\citep{hasler2009coreferential}} & \multicolumn{1}{l}{Reuters Corpus} & \multicolumn{1}{l}{\begin{tabular}[l]{@{}l@{}}Total docs: \\(NP+Event Coreference)\\ 104+20=124\end{tabular}} & \multicolumn{1}{l}{\begin{tabular}[l]{@{}l@{}}News in domain of \\ Terrorism/Security\end{tabular}} & \multicolumn{1}{l}{\begin{tabular}[l]{@{}l@{}}Available in NP4E \\defined annotation and\\ MMAX format\end{tabular}} & \multicolumn{1}{l}{Freely Available} \\ 
		\multicolumn{1}{l}{ACE 2007~\citep{doddington2004automatic}} & \multicolumn{1}{l}{\begin{tabular}[l]{@{}l@{}}News articles from: \\New York Times, \\Cable News Network, etc.\end{tabular}} & \multicolumn{1}{l}{Total docs: 599} & \multicolumn{1}{l}{\begin{tabular}[l]{@{}l@{}}Weblogs, Broadcast \\Conversation, Broadcast \\News, News Groups\end{tabular}} & \multicolumn{1}{l}{ACE format} & \multicolumn{1}{l}{\begin{tabular}[l]{@{}l@{}}Available by payment\\ through LDC\end{tabular}} \\ 
		\multicolumn{1}{l}{WikiCoref~\citep{ghaddar2016wikicoref}} & \multicolumn{1}{l}{Wikipedia} & \multicolumn{1}{l}{Total docs:30} & \multicolumn{1}{l}{\begin{tabular}[l]{@{}l@{}}People, Organization, \\ Human made Object,\\ Occupation\end{tabular}} & \multicolumn{1}{l}{CoNLL format} & \multicolumn{1}{l}{Freely Available} \\ 
		\multicolumn{1}{l}{GNOME corpus~\citep{poesio2004discourse}} & \multicolumn{1}{l}{\begin{tabular}[l]{@{}l@{}}Museum:ILEX, SOLE \\corpora Pharmaceuticals:\\ICONOCLAST corpora,\\ and Dialogues: \\Sherlock corpus\end{tabular}} & \multicolumn{1}{l}{Total docs:5} & \multicolumn{1}{l}{\begin{tabular}[l]{@{}l@{}}Museum, Pharmaceuticals,\\ Tutorial Dialogues\end{tabular}} & \multicolumn{1}{l}{GNOME format} & 
		\multicolumn{1}{l}{Not Available} \\ 
		\bottomrule
	\end{tabular}
\end{table}
\end{landscape}
The preceding datasets are labelled on multiple text genres. Recently, there has also been a surge of interest in the area of domain specific CR, particularly biomedical CR. This can be attributed to the BioNLP-2011~\citep{kim2011overview} CR task which was built on the GENIA corpus and contains Pubmed abstracts. There are mainly two lines of research in the biomedical CR task: annotation of abstract and full-text annotations.

In abstract annotation, biomedical abstracts are labelled with co-referent entity types. These datasets mainly use annotation scheme like MUC-7 and restrict to labelling of only biomedical entity types. The MedCo$^{1}$ corpus described by~\citep{su2008coreference} consists of coreference annotated Medline abstracts from GENIA dataset. 
The Protein Coreference resolution task was a part of BioNLP-2011 shared task~\citep{kim2011overview}. The dataset for the task was a combination of three resources: MedCo coreference annotation~\citep{su2008coreference}, Genia event annotation~\citep{kim2008corpus}, and Genia Treebank~\citep{tateisi2005syntax} all of which were based on the GENIA corpus by~\citep{kim2003genia}. This task focused on resolution of names of proteins. Medstract is a large corpus of medline abstracts and articles labelled for CR which was introduced by~\citep{pustejovsky2002medstract}. The coherence and anaphora module of this dataset focuses on resolution of biologically relevant sortal terms (proteins, genes as well as pronominal anaphors). It is mainly concerned with two types of anaphora namely pronominal and sortal anaphora. DrugNerAR corpus proposed by~\citep{segura2009drugnerar} aims at resolving anaphoras for extraction drug-drug interactions in pharmacological literature. It is derived from the DrugBank corpus which contains 4900 drug entries. This corpus was created by extracting 49 structured and plain unstructured and plain documents which were randomly taken from field interactions and, hence, annotated for nominal and pronominal anaphora.

There are a lot of benefits associated with using full-text instead of abstracts for biomedical CR. Though such fully annotated biomedical texts are not very accessible, there are three very interesting projects which aim at creating this type of corpora. The Corlando Richly Annotated Full-Text or CRAFT corpus~\citep{cohen2010structural} contains 97 full-length open access biomedical journal articles that have been annotated both semantically and syntactically to serve as a resource for the BioNLP community. Unlike the other corpora created for CR in biomedical literature this corpus is drawn from diverse biomedical disciplines and are marked up to their entirety. The FlySlip corpus~\citep{gasperin2008statistical} was introduced with the aim of addressing the issues associated with the earlier BioNLP Corpora which mainly considered only short abstracts. Since anaphora is a phenomenon that develops through a text, this paper posited that short abstracts are not he best resources to work with. The domain of this corpora is fruit fly genomics and it labels direct and indirect sortal anaphora types. The HANNAPIN corpus~\citep{batista2011building} was a successor of CRAFT corpus which also annotates full biomedical articles for CR. The annotated 20 full-text covers several semantic types like proteins, enzymes, cell lines and pathogens, diseases, organisms, etc. This resource is openly available for researchers.
\begin{table}[H]
	\caption{Comparison of Biomedical coreference datasets}
	\label{tab:1}    
	\setlength{\tabcolsep}{10pt}
	\renewcommand{\arraystretch}{2.3}
	\centering
	\begin{tabular}{lllll}
		\hline\noalign{\smallskip}
		Dataset & Type &Statistics & Annotation & Availability\\
		\noalign{\smallskip}\hline\noalign{\smallskip}
		MEDSTRACT& Abstract Annotation & 100 abstracts& MUCCS &publicly available\\
		MEDCo-A& Abstract Annotation & 1999 abstracts & MUCCS & publicly available\\
		MEDCo-B& Full-Text Annotation & 43 full texts& MUCCS & currently unavailable\\
		FlySlip & Full-Text Annotation & 5 full texts& FlySlip scheme & publicly available\\
		CRAFT & Full-Text Annotation & 97 full texts& OntoNotes & currently unavailable \\
		DrugNERAr & Full-Text Annotation & 49 full texts & MUCCS & publicly available\\
		HANNAPIN & Full-Text Annotation &20 full texts& MEDCo-scheme & publicly available \\
		\noalign{\smallskip}\hline
	\end{tabular}
\end{table}
.
\section{Reference Resolution Algorithms}
\subsection{Rule-based entity resolution}
Reference resolution task in NLP has been widely considered as a task which inevitably depends on some hand-crafted rules. These rules are based on syntactic and semantic features of the text under consideration. Which features aid entity resolution and which do not has been a constant topic of debate. There have also been studies conducted specifically targeting this issue~\citep{bengtson2008understanding, moosavi2017lexical}. Thus, most of the earlier AR and CR algorithms were dependent on a set of hand-crafted rules and, hence, were \textquotedblleft knowledge rich\textquotedblright{}.

Hobb's na\"ive algorithm~\citep{hobbs1978resolving} was one of the first algorithm to tackle AR. This algorithm used a rule-based, left to right breadth-first traversal of the syntactic parse tree of a sentence to search for an antecedent. The Hobb's algorithm also used world knowledge based selectional constraints for antecedent elimination. The rules and selectional constraints were used to prune the antecedent search space till the algorithm converged to a single antecedent. This algorithm was manually evaluated on different corpora like fiction and non-fiction books and news magazines.

 Another knowledge-rich algorithm was the Lappin and Leass algorithm~\citep{lappin1994algorithm} for pronominal AR. This algorithm was based on the salience assignment principle. This algorithm maintained a discourse model consisting of all potential antecedent references corresponding to a particular anaphor. Each antecedent was assigned a salience value based on a number of features. The salience categories were recency, subject emphasis, existential emphasis, accusative emphasis, indirect object emphasis, non-adverbial emphasis and head noun emphasis. The strategy followed here was to penalize or reward an antecedent based on its syntactic features. The algorithm started with generation of a list of possible antecedents extracted using the syntactic and semantic constraints mentioned earlier. Then, a salience value was assigned to each antecedent. The salience was calculated as a sum over all the predetermined salience values corresponding to the salience category satisfied. The antecedent with the maximum salience value was proposed as the appropriate antecedent. The Lappin and Leass algorithm also incorporated a signal attenuation mechanism wherein the influence or salience of an antecedent was halved on propagation to next sentence in the discourse and was evaluated on a dataset consisting of five computer science manuals.
 
 The earliest attempt at exploiting discourse properties for pronoun resolution was the BFP algorithm~\citep{brennan1987centering}. This algorithm motivated the centering theory. The centering theory~\citep{grosz1995centering} was a novel algorithm used to explain phenomenon like anaphora and coreference using discourse structure. In centering theory, the center was defined as an entity referred to in the text which linked multiple \textquotedblleft utterances\textquotedblright{} or sentences in the discourse. Forward looking centers were defined as set of centers that were referred to in an utterance. The backward looking center was defined as a single center belonging to the intersection of the sets of forward centers of the current and the preceding utterance. This algorithm started with creation of all possible anchors, i.e., pairs of forward centers and backward entities. The ordering of the centers was done according to their prominence and their position in the utterance. The backward looking center was defined as the current topic and the preferred center was the potential new topic. The three major phases in center identification were: center continuation, where same center was continued for the next sentence in discourse, center retaining, wherein there was a possible indication for shift of the center and center-shifting wherein there was a complete shift in the center involved. As summarized by~\citep{kibble2001reformulation} there were two key rules governing centering theory. The \textbf{Rule 1} stated that the center of attention was the entity that was most likely to be pronominalized and \textbf{Rule 2} stated that there was a preference given to keep the same entity as the center of attention. Apart from these rules various discourse filters were also applied to filter out good and bad anchors and the remaining good ones were ranked according to their transition type. The centering algorithm was evaluated on Hobb's datasets and some other Human-Keyword task oriented databases. There were many modifications proposed on centering theory and the most significant one was the Left Right Centering theory~\citep{tetreault2001corpus,tetreault1999analysis}. This was based on the observation in Psycholinguistic research that listeners attempted to resolve an anaphor as soon as they heard it. LRC~\citep{tetreault1999analysis} first attempted to find the antecedent in the current utterance itself and if this does not work it proceeds to process the previous utterances in a left to right fashion. Another modification on LRC, i.e., LRC-F~\citep{tetreault2001corpus} also encoded information about the current subject into the centering theory.
 
 Though most of the rule-based algorithms were knowledge rich, there were some~\citep{baldwin1997cogniac,harabagiu2001text, haghighi2009simple, lee2013deterministic, zeldes2016annotation} that aimed at reducing the level of dependency of rules on external knowledge. These were categorized as the \textquotedblleft knowledge-poor algorithms\textquotedblright. CogNIAC~\citep{baldwin1997cogniac} was a high precision coreference resolver with limited resources. This early method moved a step closer to how human beings resolve references. Take, for example, the sentence: \emph{Charles (1) went to the concert with Ron (2) and he hurt his (3) knee on the way back.} Here, the resolution of (3) is an intricate task for a human being due to inevitable requirement of knowledge beyond the discourse. Thus, CogNIAC was based on the simple but effective assumption that there existed a sub class of anaphora that did not require general purpose reasoning. Thus, if an anaphoric reference required external world resources in its resolution CogNIAC simply did not attempt its resolution. Here, CogNIAC could be considered to be analogous to a human who recognizes knowledge intensive resolutions and makes a decision on when to attempt resolution. CogNIAC was evaluated on myriad datasets like narratives and newspaper articles and in scenarios with almost no linguistic preprocessing to partial parsing. The core rules defining CogNIAC were picking a unique or single existent antecedent in current or prior discourse, the nearest antecedent for a reflexive anaphor, picking exact prior or current string match for possessive pronoun, etc. Adhering to the these core rules or presuppositions, the CogNIAC's algorithm proceeded to resolve pronouns from left to right in the given text. Rules were followed in an orderly fashion and once a given rule was satisfied and antecedent match occurred no further rules are attempted. On the other hand, if none of the rules were satisfied CogNIAC left the anaphor unresolved. Two additional constraints were deployed during the evaluation phase of CogNIAC. These two constraints were picking the backward center which is also the antecedent as the target solution and the second one was picking the most recent potential antecedent in the text. CogNIAC was evaluated on multiple datasets like narratives and newspaper articles.
 
 Apart from the methods discussed earlier which were a combination of salience, syntactic, semantic and discourse constraints, attempts have also been made to induce world knowledge into the CR systems. The COCKTAIL system~\citep{harabagiu1999knowledge}, basically a blend of multiple rules, was one such system which took a knowledge-based approach to mining coreference rules. It used WordNet for semantic consistency evidence and was based on structural coherence and cohesion principles. It was evaluated on the standard MUC 6 CR dataset. 
 
 Another rule-based algorithm which took a knowledge-based approach to entity resolution specifically for pronominal AR was the rule-based algorithm by~\citep{liang2004automatic} for automatic pronominal AR. In this algorithm, WordNet ontology and heuristic rules were deployed to develop an engine for both intra-sentential and inter-sentential antecedent resolution. This algorithm started with parsing each sentence in the text, POS tagging and lemmatizing it. These linguistic features were stored in an internal data structure. This global data structure was appended with some other features like base nouns, number agreement, person name identification, gender, animacy, etc. This model also constructed a finite state machine with the aim of identifying the NPs. The parsed sentence was then sequentially checked for anaphoric references and pleonastic it occurrences. The remaining mentions were considered as possible candidates for antecedents and were heuristically evaluated using a scoring function. The toolkit was extensively evaluated on reportage, editorials, reviews, religion, fiction, etc.
 
As the research in CR started to shift towards machine learning algorithms which used classification and ranking it slowly became clear that to beat the machine learning systems, rules had to be ordered according to their importance. A rule-based CR baseline which gained wide acclaim was the deterministic CR system by Haghini and Klein (H and K model)~\citep{haghighi2009simple}, who proposed a strong baseline by modularizing syntactic, semantic and discourse constraints. In spite of its simplicity it outperformed all the unsupervised and most of the supervised algorithms proposed till then. This algorithm first used a module to extract syntactic paths from mentions to antecedents using a syntactic parser. It then proceeded by eliminating some paths based on deterministic constraints. After this, another module was used evaluate the semantic compatibility of headwords and individual names. Compatibility decisions were made from compatibility lists extracted from corpora. The final step was the elimination of incompatible antecedents and selection of the remaining antecedents so as to minimize the tree distance. This algorithm was evaluated on multiple versions of the ACE corpus and the MUC-6 dataset and achieved significant improvements in accuracy. 

The H and K model~\citep{haghighi2009simple} motivated the use of \textquotedblleft successive approximations\textquotedblright{} or multiple hierarchical sieves for CR. The current version of Stanford CoreNLP deterministic CR system is a product of extensive investigations conducted on deciding the precise rules to govern the task of CR. This system was an outcome of three widely acclaimed papers~\citep{raghunathan2010multi,lee2011stanford,lee2013deterministic}. Though rule-based systems have lost their popularity in favor of deep learning algorithms, it is very interesting to understand how this multi-sieve based approach for CR improved over time.
The work of~\citep{raghunathan2010multi} was motivated by the hypothesis that a single function over a set of constraints or features did not suffice for CR as lower precision features could often overwhelm higher precision features. This multi-sieve approach proposed a CR architecture based on a sieve that applied tiers of deterministic rules ordered from high precision to lowest precision one be one. Each sieve built on the result of the previous cluster output. The sieve architecture guaranteed that the important constraints were given higher precedence. This algorithm had two phases. The first one was the mention processing phase wherein the mentions were extracted, sorted and pruned by application of myriad constraints. The second phase was the multi-pass sieve phase which used multiple passes like string match, head match, precise constraints like appositives, shared features like gender, animacy, number, etc. This system was evaluated on the same datasets as the H and K model~\citep{haghighi2009simple} and outperformed most of the baselines. 

An extension of the multi-sieve approach~\citep{raghunathan2010multi} was presented at the CoNLL 2011 shared task~\citep{pradhan2011conll}. The major modifications made to the earlier system were addition of five more sieves, a mention detection module at the beginning and, finally, a post-processing module at the end to provide the result in OntoNotes format. This system was ranked first in both the open and closed tracks of the task. A more detailed report and more extensive evaluation of this system was also reported by Lee et al.~\citep{lee2013deterministic}, who delineated the precise sieves applied using an easy to understand and intuitive example. Like the earlier system~\citep{raghunathan2010multi} this approach also incorporated shared global entity-level information like gender, number and animacy into the system to aid CR. The figure below shows the composition of different sieves used in this deterministic system.
\begin{figure}[H]
	\centering
	\includegraphics[width=12cm]{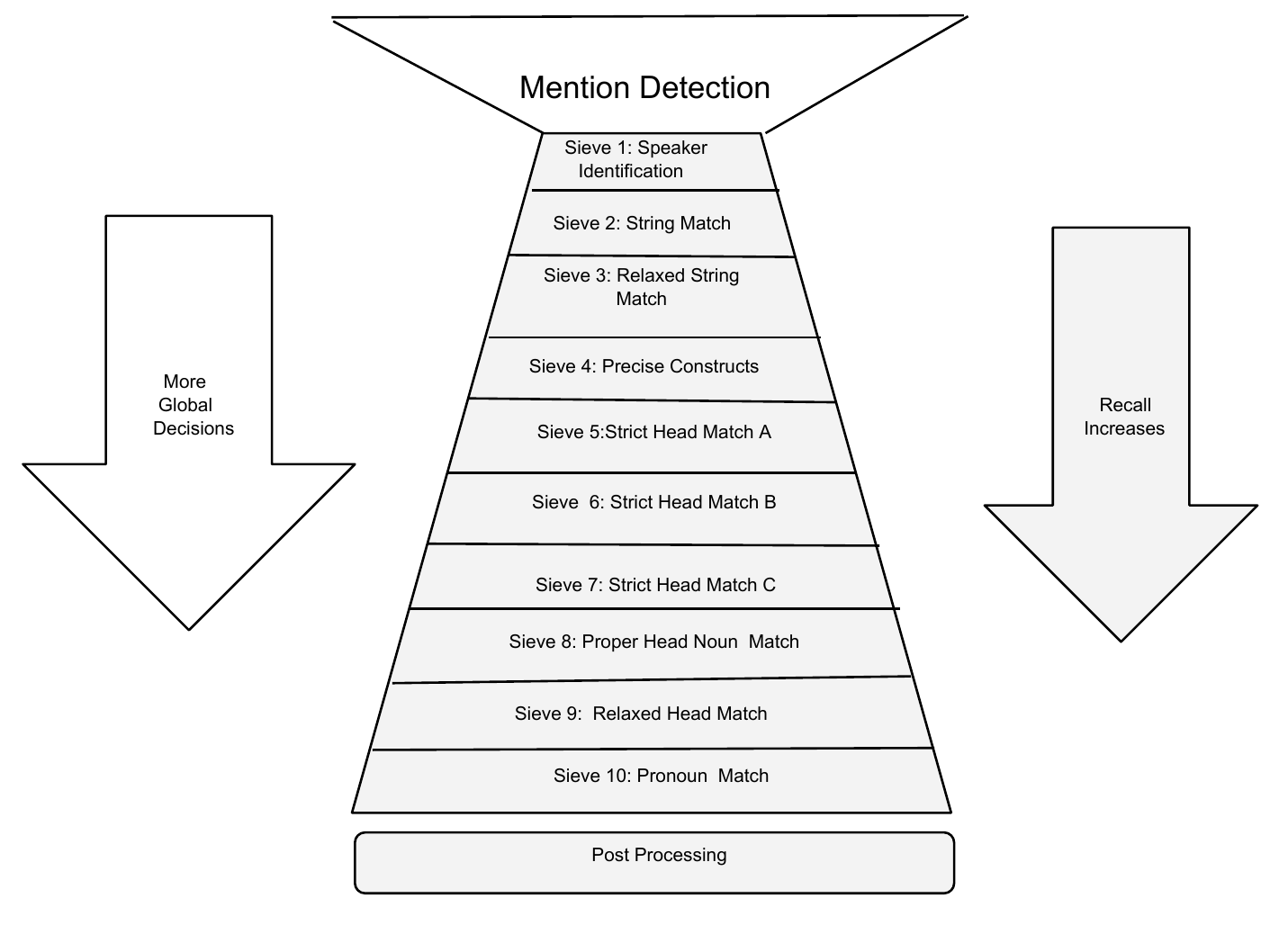}
	\caption{ Coreference Resolution Sieve~\citep{lee2013deterministic}}
\end{figure}
The shifting trend of CR research from rule-based systems to deep learning systems has come at the cost of loss of the ability of the CR systems to adapt to different coreference phenomenon and border definitions, when there is no access to large training data in the desired target scheme~\citep{zeldes2016annotation}. A recent rule-based algorithm~\citep{zeldes2016annotation} also used dependency syntax as input. It aimed at targeting the coreference types which were not annotated by the CoNLL 2012 shared task like cataphora, compound modifier, i-within-i, etc. This system was called Xrenner and was evaluated on two very different corpora, i.e., the GUM corpus and the WSJ. It used semantic and syntactic constraints and rules for antecedent filtering. Xrenner was compared with other well-known rule-based systems like Stanford CoreNLP~\citep{lee2013deterministic} and Berkeley systems~\citep{durrett2013easy} on the datasets mentioned earlier and outperformed all of them. Xrenner raised a very important question of the domain-adaptation problem of learning-based systems. 

Hobbs algorithm~\citep{hobbs1978resolving} was a syntax-based algorithm while centering theory~\citep{grosz1995centering} was discourse-based algorithm. Lappin and Leass~\citep{lappin1994algorithm} algorithm, on the other hand, can be seen as a hybrid since it was both syntax- and discourse-based. In addition, it also made use of knowledge resources and morphological and semantic information to rank the possible antecedents. These three algorithms were amongst the earliest algorithms for AR and, hence, the evaluation metrics and datasets used for their evaluation were not the standardized ones. This makes comparison of these algorithms with the recent rule-based algorithms extremely difficult. Also, the Hobb's algorithm~\citep{hobbs1978resolving} was hand evaluated and, hence, was prone to human errors. The datasets used for evaluation of this algorithm also raise a concern. As pointed out by contemporaries, in most cases the resolution of the antecedent was trivial. Another issue with the algorithm is that it was based on the assumption that the correct syntax parse of the sentence always exits. Nonetheless, this algorithm is still highly regarded as a strong baseline given its simplicity and ease of implementation. 

The Lappin and Leass algorithm~\citep{lappin1994algorithm} which is still highly regarded in AR research also had some drawbacks. First is that Lappin and Leass algorithm was mainly aimed only at pronominal AR which only forms a small subset of the AR task. Another drawback is that the Lappin and Leass algorithm is highly knowledge driven. This dependency on knowledge resources can become very problematic especially when the required knowledge resources were not accessible. Another loophole of this algorithm is the weight assignment scheme for the different salience categories. These weights are decided by extensive corpus experiments. Hence, the fundamental question which arises here is are these values corpus-dependent. The weight initialization stage can be very problematic when trying to adapt this algorithms to other corpora. The Lappin and Leass algorithm (RAP) was compared with the Hobb's algorithm for intra-sentential and intra-sentential case on a dataset of computer manuals. The Hobb's algorithm outperformed the RAP algorithm for the inter-sentential case (87\% vs 74\%) while the RAP algorithm outperformed the Hobb's algorithm for the intra sentential case (89\% vs 81\%). Overall, RAP algorithm outperformed Hobb's algorithm by 4\%. In spite of Lappin and Leass algorithm's success, it is important to bear in mind that this algorithm was tested on the same genre as its development set, while the genre used by Hobbs for the development of his own algorithm was very different from the test set. High Precision systems like CogNIAC~\citep{baldwin1997cogniac} aimed at circumscribing the heavy dependency of the RAP algorithm on external knowledge resources by taking a knowledge-minimalistic approach. CogNIAC achieved performance at-par with Hobb's algorithm in the \textquotedblleft resolve-all\textquotedblright{} setting. In spite of this CogNIAC had issues of its own. Its rules were defined only for specific reference types and it was mainly useful for systems which required high precision resolution at the cost of a low recall. As a result of this, the defined rules performed well on the narratives dataset but CogNIAC failed to meet a high precision when evaluated on the MUC-6 corpus.

The centering theory~\citep{grosz1995centering,walker1998centering} was a discourse-based algorithm that took a psycholinguistic approach to AR. Centering theory was an attractive algorithm for researchers mainly because the discourse information it requires could be obtained from structural properties of utterances alone. Thus, eliminating the need for any extra-linguistic semantic information. One possible disadvantage of CT was its preference for inter-sentential references as compared to intra-sentential references. In some ways we can even consider that Lappin and Leass algorithm incorporated centering theory's discourse constraint and modified it by assigning weights to these discourse phenomenon. A manual comparison of Hobb's algorithm~\citep{hobbs1978resolving} and CT-based algorithm by~\citep{walker1989evaluating} showed that the two performed equally over a fictional dataset of 100 utterances, but the Hobb's algorithm outperformed CT for news paper articles domain (89\% vs 79\%) and task domain (51\% vs 49\%).
In spite of this spurge in interest in this field with the methods discussed earlier, it is important to note one important thing. The evaluation standards of these algorithms were very inconsistent~\citep{mitkov2001towards} and this slowly started to change with the evaluation guidelines laid by the MUC~\citep{grishman1996message}, ACE~\citep{doddington2004automatic} and CoNLL~\citep{pradhan2011conll} corpora.

Another widely accepted and extensively evaluated rule-based system was the coreference system by Haghini and Klein~\citep{haghighi2009simple}. This system was evaluated on multiple standard datasets like the MUC and ACE corpora. This simple but effective algorithm was purely based on syntax and had well-defined antecedent pruning rules. Instead of weighting the salience categories like Lappin and Leass, this algorithm defined rules which were successively applied starting with the most important ones. This algorithm formed the first effective baseline comparison of rule-based CR approaches. Its main strength was its simplicity and effectiveness. The idea of defining rules was further developed and delineated more intuitively using a novel sieve architecture for CR~\citep{raghunathan2010multi}. Overtime there were a couple of additions and modifications made to this architecture to improve its performance~\citep{lee2013deterministic}, result of which is the current version of the best performing rule-based system of Stanford CoreNLP. This coreference system is extremely modular and new coreference models can be easily incorporated into it. Overall, we observe that the rules and constraints deployed became even more fine-grained as the CR research took pace. This was mainly because the focus of the task started to shift towards CR which has a much broader scope than AR. 
\begin{landscape}
	\setlength\tabcolsep{1pt}
	\footnotesize
	\begin{threeparttable}[htbp]
		\centering
		\caption{Rule-based entity resolution algorithm}
		\label{table:4}
		\bgroup
		\def\arraystretch{1.2}
		
		\begin{tabular}{lllll}
			\hline
			Algorithm & Dataset & Evaluation Metric & Metric Value & Algorithm Rule Types \\ 
			\hline
			~\citep{hobbs1978resolving} & \begin{tabular}[c]{@{}l@{}}Fictional, Non-fictional,\\ Books, Magazines, Part of \\ Brown Corpus\end{tabular} & Hobb's metric*& \begin{tabular}[c]{@{}l@{}}88.3\% (without selectional constraints)\\ 91.7\% (with selectional constraints)\end{tabular} &\begin{tabular}[c]{@{}l@{}}Syntax-based rules+\\Selectional rules\end{tabular} \\ \hline
			\begin{tabular}[c]{@{}l@{}}~\citep{lappin1994algorithm}\end{tabular} & \begin{tabular}[c]{@{}l@{}}Five Computer \\ Science Manuals\end{tabular} & Hobb's metric & \begin{tabular}[c]{@{}l@{}}74\% (inter-sentential)\\ 89\% (intra-sentential)\end{tabular} & \begin{tabular}[c]{@{}l@{}}Hybrid of: Syntax Rules+ \\Discourse Rules +Morphological+\\Semantic\end{tabular} \\ \hline
			~\citep{walker1998centering} & \begin{tabular}[c]{@{}l@{}}2 of the fiction and \\ non fiction books \\ same as Hobb's+ 5 \\ Human-keyword and task-oriented \\ and task oriented databased\end{tabular} & Hobb's metric & Overall: 77.6\% &\begin{tabular}[c]{@{}l@{}}Discourse-based rules \\and constraints \end{tabular} \\ \hline
			\multirow{2}{*}{~\citep{baldwin1997cogniac}} & Narratives & \multirow{2}{*}{\begin{tabular}[c]{@{}l@{}}Precision and Recall\end{tabular}} & P:92\% R:64\% & \multirow{2}{*}{Discourse rules+Syntax rules} \\ \cline{2-2} \cline{4-4}
			& MUC-6 & & P:73\% R:75\% & \\ \hline
			~\citep{liang2004automatic}& \begin{tabular}[c]{@{}l@{}}Random Texts from \\ Brown Corpora\end{tabular} & Hoobs's metric & Overall:77\% & \begin{tabular}[c]{@{}l@{}}Semantic constraints+Discourse\\constraints+ Syntactic Constraints\end{tabular} \\ \hline
			\multirow{4}{*}{~\citep{haghighi2009simple}} & ACE 2004 Roth-dev & \multirow{4}{*}{\begin{tabular}[c]{@{}l@{}}MUC,$B^{3}$, CEAF \\ \\   (F1 values)\end{tabular}} & MUC:75.9, $B^{3}$:77.9,CEAF:72.5 & \multirow{4}{*}{\begin{tabular}[c]{@{}l@{}}Syntactic rules+\\Semantic rules\end{tabular}} \\ \cline{2-2} \cline{4-4}
			& ACE 2004 Culotta-test & & MUC:79.6,$B^{3}$:79,CEAF:73.3 & \\ \cline{2-2} \cline{4-4}
			& MUC-6 Test & & MUC:81.9, $B^{3}$:75.0,CEAF:72 & \\ \cline{2-2} \cline{4-4}
			& ACE 2004 nwire & & MUC:76.5, $B^{3}$:76.9, CEAF:71.5 & \\ \hline
			\multirow{4}{*}{~\citep{raghunathan2010multi}} & ACE 2004 Roth-dev & \multirow{4}{*}{MUC, $B^{3}$} & MUC: 78.6, $B^{3}$:80.5 & \multirow{4}{*}{\begin{tabular}[c]{@{}l@{}}Syntactic rules+\\Semantic rules(minimal)\end{tabular}} \\ \cline{2-2} \cline{4-4}
			& ACE 2004 Culotta-test & & MUC:75.8, $B^{3}$:80.4 & \\ \cline{2-2} \cline{4-4}
			& MUC-6 Test & & MUC:77.7, $B^{3}$:73.2 & \\ \cline{2-2} \cline{4-4}
			& ACE 2004 nwire & & MUC:78.1, $B^{3}$:78.9 & \\ \hline
			\multirow{4}{*}{~\citep{lee2013deterministic}} & ACE 2004 Culotta-test & \multirow{3}{*}{MUC, $B^{3}$} & MUC:75.9, $B^{3}$:81 & \multirow{4}{*}{Syntactic rules+Semantic rules} \\ \cline{2-2} \cline{4-4}
			& ACE 2004 nwire & & MUC:79.6, $B^{3}$:80.2 & \\ \cline{2-2} \cline{4-4}
			& MUC6-Test & & MUC:78.4, $B^{3}$:74.4 & \\ \cline{2-4}
			& CoNLL 2012 & MUC, $B^{3}$, CEAF, CoNLL & \begin{tabular}[c]{@{}l@{}}MUC:63.72, $B^{3}$:52.08, \\ CEAF:47.79, CoNLL:60.13\end{tabular} & \\ \hline
			\multirow{2}{*}{~\citep{zeldes2016annotation}} & GUM corpus & \multirow{2}{*}{MUC, $B^{3}$, CEAF, CoNLL} & \begin{tabular}[c]{@{}l@{}}MUC:55.95, $B^{3}$:49.09, \\ CEAFe: 44.47, CoNLL:49.84\end{tabular} & \multirow{2}{*}{Syntactic Rules} \\ \cline{2-2} \cline{4-4}
			& Wall Street Journal Corpus & & \begin{tabular}[c]{@{}l@{}}MUC:49.23, $B^{3}$:41.52, \\ CEAFe:41.13, CoNLL: 43.96\end{tabular} & \\ \hline

		\end{tabular}

		\egroup
		\begin{tablenotes}\footnotesize
			\item[*]$\text{Hobbs metric}=\frac{\text{Number of correct resolutions}}{\text{Total No. of Resolutions attempted}}$
		\end{tablenotes}
	\end{threeparttable}
\end{landscape}

 \subsection{Statistical and machine learning based entity resolution}
The field of entity resolution underwent a shift during the late nineties from heuristic- and rule-based approaches to learning-based approaches. Some of the early learning-based and probabilistic approaches for AR used decision trees~\citep{aone1995evaluating}, genetic algorithms~\citep{mitkov2002new,mitkov2001towards} and Bayesian rule~\citep{ge1998statistical}. These approaches set the foundation for the learning-based approaches for entity resolution which improved successively over time and, finally, outperformed the rule-based algorithms. This shift was mainly because of the availability of tagged coreference corpora like the MUC and ACE corpora. The research community of CR expanded from linguists to machine learning enthusiasts. Learning-based coreference models can be classified into three broad categories of mention-pair, entity-mention and ranking model.

The mention-pair model treated coreference as a collection of pairwise links. It used a classifier to make a decision whether two NPs are co-referent. This stage was followed by the stage of reconciling the links using methods like greedy partitioning or clustering to create an NP partition. This idea was first proposed for pronoun resolution~\citep{aone1995evaluating,mccarthy1995using} in the early nineties using the decision tree classifier~\citep{quinlan1986induction} and is still regarded as a simple but very effective model. The mention pair model had three main phases each of which acquired significant research attention. It is important to note here that every phase of the mention-pair model was independent and improvement in the performance of one stage did not necessarily imply improvement in accuracy of the subsequent phases.

The first phase of the mention-pair model was the creation of training instances. Since most entities in the text were non-coreferent, the aim of training instance creation was to reduce the skewness involved in the training samples. The most popular algorithm for mention instance creation was the Soon et al.~'s heuristic mention creation method~\citep{soon2001machine}. Soon's method created a positive instance between a NP \emph{A1} and its closest preceding antecedent \emph{A2} and a negative instance by pairing \emph{A1} with each of the NPs intervening between \emph{A1} and \emph{A2}. It only considered annotated NPs for instance creation. A modification on this approach~\citep{ng2002improving} enforced another constraint that a positive instance between a non-pronominal instance \emph{A1} and antecedent \emph{A2} could only be created if \emph{A2} was non-pronominal too. Other modifications on Soon's instance creation~\citep{yang2003coreference,strube2002influence} used number, gender agreement, distance features for pruning of incorrect instances. There have also been some mention creation systems~\citep{harabagiu2001text,ng2002combining} which learnt a set of rules with the aim of excluding the hard training instances whose resolution was difficult even for a human being. 

The second phase of mention-pair models was the training of a classifier. Decision trees and random forests were widely used as classifiers~\citep{aone1995evaluating,mccarthy1995using,lee2017scaffolding} for CR. In addition, statistical learners~\citep{berger1996maximum,ge1998statistical}, memory learners like Timbl~\citep{daelemans2004timbl} and rule-based learners~\citep{cohen1999simple} were also widely popular.

The next phase of the mention pair model was the phase of generating an NP partition. Once the model was trained on an annotated corpus it could be tested on a test-set to obtain the coreference chains. Multiple clustering techniques were deployed to tackle this task. Some of the most prominent ones were best-first clustering~\citep{ng2002improving}, closest-first clustering~\citep{soon2001machine}, correlational clustering~\citep{mccallum2005conditional}, Bell Tree beam search~\citep{luo2005coreference} and graph partitioning algorithms~\citep{nicolae2006bestcut,mccallum2003object}. 
In the closest first clustering~\citep{soon2001machine} all possible mentions before the mention under consideration were processed from right to left, processing the nearest antecedent first. Whenever the binary classifier returned true the two references were linked together and the further antecedents were not processed. Further, the references could be clustered using transitivity between the mentions. A modification on this approach~\citep{ng2002improving} linked the current instance instead with the antecedent which is classified as true and has the maximum likelihood, i.e., the best antecedent. Though this method had an overhead of processing all possible antecedent before conclusively deciding on one, the state-of-the-art model~\citep{lee2017end} also uses a version of this clustering albeit by restricting the search-space of the antecedent. Another kind of clustering deployed to generate the NP partition was the correlational clustering algorithm~\citep{mccallum2005conditional}. This algorithm measured the degree of inconsistency incurred by including a node in a partition and making repairs. This clustering type was different from the ones discussed earlier as the assignment to the partition was not only dependent on the distance measure with one node but on a distance measurement between all the nodes in a partition. For example, this clustering type avoided assigning the reference \emph{she} to a cluster containing \emph{Mr. Clinton} and \emph{Clinton}. Though the classifier could have predicted that \emph{Clinton} is an antecedent of \emph{she} this link was avoided by using correlational clustering. Another variant of clustering algorithms used graph-partitioning. The nodes of the graph represented the mentions and the edge weights represented the likelihood of assignment of the pairs. Bell trees~\citep{luo2005coreference} were also used for creating an NP partition. In a Bell Tree, the root node was the initial state of the process which consisted of a partial entity containing the first mention of the document. The second mention was added in the next step by either linking to the existing entity or starting a new entity. The second layer of nodes was created to represent possible outcomes and subsequent mentions are added to the tree in a similar manner. The process was mention synchronous and each layer of the tree nodes was created by adding one mention at a time.

Another direction of research in mention pair models attempted at combining the phases of classification and effective partitioning using Integer Linear Programming~\citep{denis2007joint,finkel2008enforcing}. As posited by Finkel and Manning~\citep{finkel2008enforcing} this task was suitable for integer linear programming (ILP) as CR required to take into consideration the likelihood of two mentions being coreferent during two phases: pair-wise classification and final cluster assignment phase. The ILP models first trained a classifier over pairs of mentions and encode constraints on top of probability outputs from pairwise classifiers to extract the most probable legal entity assignments. The difference between two ILP models mentioned earlier was that the former does not enforce transitivity while the latter encodes the transitivity constraint while making decisions. However, the ILP systems had a disadvantage that ILP is an NP-hard problem and this could create issues when the length of the document decreased. Another recently proposed model which eliminated the classification phase entirely was the algorithm by Fernandes et al.~\citep{fernandes2012latent}. Their model had only two phases of mention detection and clustering. The training instances were a set of mentions \emph{x} in the document and the correct co-referent cluster \emph{y}. The training objective was a function of the cluster features (lexical, semantic, syntactic, etc.). This algorithm achieved an official CoNLL score of 58.69 and was one of the best performing systems in closed track of CoNLL 2012 shared-task.
\begin{table}[H]
	\centering
	\caption{Mention pair variants}
	\label{table5}
	\resizebox{\textwidth}{!}{
		\begin{tabular}{lllllll}
			\toprule
			\multirow{2}{*}{Algorithm} & \multirow{2}{*}{NP Partitioning Algorithm} & \multirow{2}{*}{Learning Algorithm} & \multirow{2}{*}{Dataset} & \multicolumn{3}{l}{Performance metrics} \\ 
			
			& & & & Accuracy & MUC & $B^3$ \\
			\hline 
			\multirow{2}{*}{\begin{tabular}[c]{@{}l@{}}~\citep{mccarthy1995using}\end{tabular}} & \multirow{2}{*}{\begin{tabular}[c]{@{}l@{}}Used Symmetricity and \\ Transitivity to link\\ mentions\end{tabular}} & \multirow{2}{*}{Decision Tree C4.5} & \begin{tabular}[c]{@{}l@{}}English \\ Joint Venture \\ Articles\end{tabular} & 86.5 & - & - \\ \cline{4-7} 
			& & & MUC-6 & - & 47.2 & - \\ \hline
			\multirow{2}{*}{~\citep{soon2001machine}} & \multirow{2}{*}{Closest first clustering} & \multirow{2}{*}{Decision Tree C5} & MUC-6 & - & 62.6 & - \\ \cline{4-7} 
			& & & MUC-7 & - & 60.4 & - \\ \hline
			\multirow{2}{*}{~\citep{ng2002improving}} & \multirow{2}{*}{Best first clustering} & RIPPER & MUC-6 & - & 70.4 & - \\ \cline{3-7} 
			& & Decision Tree & MUC-7 & - & 63.4 & - \\ \hline
			~\citep{bengtson2008understanding} & Best first clustering & \begin{tabular}[c]{@{}l@{}}Averaged Perceptron\\ Algorithm\end{tabular} & \begin{tabular}[c]{@{}l@{}}ACE-Culotta\\ test\end{tabular} & - & 75.8 & 80.8 \\ \hline
			\multirow{3}{*}{~\citep{denis2007joint}} & \multirow{6}{*}{\begin{tabular}[c]{@{}l@{}}Global inference with \\ Integer Linear Programming\end{tabular}} & \multirow{3}{*}{\begin{tabular}[c]{@{}l@{}}Maximum Entropy\\ Model\end{tabular}} & ACE-BNEWS & - & 69.2 & - \\ \cline{4-7} 
			& & & ACE-NPAPER & - & 72.5 & - \\ \cline{4-7} 
			& & & ACE-NWIRE & - & 67.5 & - \\ \cline{1-1} \cline{3-7} 
			\multirow{3}{*}{~\citep{finkel2008enforcing}} & & \multirow{3}{*}{Logistic Classifier} & MUC-6 & - & 68.3 & 64.3 \\ \cline{4-7} 
			& & & ACE-NWIRE & - & 61.1 & 73.1 \\ \cline{4-7} 
			& & & ACE-BNEWS & - & 67.1 & 74.5 \\ \hline
			\begin{tabular}[c]{@{}l@{}}~\citep{mccallum2005conditional}\end{tabular} & Graph Partitioning Algorithm & \begin{tabular}[c]{@{}l@{}}Conditional Random\\ Fields over hidden Markov\\ models\end{tabular} & MUC-6 & - & 73.42 & - \\ \hline
			~\citep{nicolae2006bestcut}
			& Graph Partitioning & \begin{tabular}[c]{@{}l@{}}Maximum Entropy\\ Model\end{tabular} & MUC-6 & - & 89.63 & - \\ \hline
			\begin{tabular}[c]{@{}l@{}}~\citep{mccallum2003object}\end{tabular} & Correlational Clustering & \begin{tabular}[c]{@{}l@{}}Conditional Random\\ Fields over hidden \\ markov models\end{tabular} & MUC-6 & -& 91.59 & - \\ \bottomrule
		\end{tabular}}
	\end{table}
In spite of being a widely used model even today for CR, there were some fundamental deficits with the mention-pair model. The first one was the constraint of transitivity which was enforced did not always hold good. This meant that if an entity \emph{A} referred to entity \emph{B} and entity \emph{B} referred to entity \emph{C} it was not always true that \emph{A} co-referred with \emph{C}, e.g., consider the case when \emph{she} is predicted antecedent of 
\emph{Obama} and \emph{Obama} is predicted antecedent of \emph{he}, but since \emph{he} is not co-referent with \emph{she} by violation of gender constraint, transitivity condition should not be enforced here. This flaw was mainly because the decisions made earlier by the co-reference classifier were not exploited to correct future decisions. The information from only two NP's here \emph{Obama} and \emph{he} did not suffice to make an informed decision that they are co-referent, as the pronoun here was semantically empty. In addition, the NP \emph{Obama} was itself ambiguous and could not be assigned any semantic feature like gender. Another disadvantage of the mention-pair model was that it only determined how good an antecedent was with respect to the anaphoric NP and not how good it was with respect to other antecedents available. The entity-mention models and the mention-ranking models were proposed with the aim of overcoming the disadvantages of the mention-pair models.

The entity mention model for CR focuses on a single underlying entity of each referent in discourse. This genre of algorithms was motivated by the fact that instead of making coreference decisions independently for every mention-antecedent pair it was necessary to exploit the past coreference decisions to inform the future ones. The entity mention model aimed at tackling this \textquotedblleft expresiveness\textquotedblright{} issue~\citep{ng2010supervised} with the mention-pair model by attempting to classify whether an NP was coreferent with a preceding partially formed cluster instead of an antecedent. Thus, the training instances for the classifier were modified to a pair of NP \emph{N} and cluster \emph{C} and a label depicting whether the assignment of the NP to the partial cluster was positive or negative. Instances were represented as cluster-level features instead of pair wise features. The cluster-level features, e.g., gender, number, etc. were defined over subsets of clusters using the \textquotedblleft ANY\textquotedblright{}, \textquotedblleft ALL\textquotedblright{}, \textquotedblleft MOST\textquotedblright{}, etc. predicates. Entity mention model was evaluated by many researchers~\citep{yang2004improving,luo2005coreference}. The former evaluated the entity-mention model in comparison to mention pair model on the ACE datasets using the decision tree classifier and inductive logic programming. The results for the entity-mention model as compared to the mention-pair model showed a slight decrease in performance using C4.5 classifier and a marginal increase in performance using inductive logic programming. The \textquotedblleft ANY\textquotedblright{} constraint to generate cluster-level features was also encoded by the Bell Tree algorithm~\citep{luo2005coreference}. However, even in this case the performance of the entity mention model was not at par with the mention-pair model. The major reason for this was that it was extremely difficult to define cluster-level features for the entity-mention model. Most of the referents did not contribute anything useful to the cluster features because they were semantically empty (e.g., pronouns). Another model which attempted using features defined over clusters for CR was the first order probabilistic model by~\citep{culotta2007first}. Most recent models~\citep{clark2015entity,clark2016improving} also attempt at learning cluster-level features. 

Mention-pair models faced an issue that they used a binary classifier to decide whether an antecedent was coreferent with the mention. The binary classifier could only provide a \textquotedblleft YES\textquotedblright{} or \textquotedblleft NO\textquotedblright{} result and failed to provide an intuition on how good one antecedent was compared to the other antecedent. The ranking models circumvented this flaw by ranking the mentions and choosing the best candidate antecedent. Ranking was considered to be a more natural way to predict the coreference links as it captured the competition between different antecedents. Some proposed models to realize this purpose were the tournament models and the twin candidate model by~\citep{yang2008twin}. On a closer observation, the earlier rule-based approaches~\citep{hobbs1978resolving,lappin1994algorithm} also used constraints or sieves in a hierarchical manner starting with the most crucial ones to converge to the best antecedent. Hence, they too in principle ranked the antecedents using constraints which were ordered by their importance. A particularly prominent work which incorporated mention-ranking was the algorithm by Dennis and Baldridge~\citep{denis2008specialized} who replaced the classification function by a ranking loss. Another mention ranking model which used only surface features~\citep{durrett2013easy} and deployed a log-linear model for antecedent selection, outperformed the Stanford system~\citep{lee2011stanford} which was the winner of CoNLL 2011 shared task~\citep{pradhan2011conll} by a margin of 3.5\% and the IMS system~\citep{bjorkelund2012data} which was the then best model for CR by a margin of 1.9\%. 

In spite of its wide spread popularity, the mention rankers were still not able to effectively exploit past decisions to make current decisions. This motivated the \textquotedblleft cluster ranking\textquotedblright{} algorithms. The cluster ranking approaches aimed at combining the best of the entity-mention models and the ranking models. Recent deep learning models~\citep{clark2016improving} have also used a combination of mention ranker and cluster ranker for CR. Another issue with the mention-ranking model was that it did not differentiate between anaphoric and non-anaphoric NP's. The recent deep learning based mention ranking models~\citep{clark2016deep,clark2016improving,wiseman2015learning,wiseman2016learning} overcome this flaw by learning anaphoricity jointly with mention ranking. One of the earlier machine learning approaches which aimed at achieving this was the work of~\citep{rahman2009supervised}.

Until recently, the best performing model on the CoNLL 2012 shared task was an entity centric model~\citep{clark2015entity}. Like other machine learning approaches, it also was feature rich. Defining features for mentions and especially for clusters is a very challenging task. Also, the extraction of the features is a time consuming task. This slowly started to change with the introduction of deep learning for NLP.

\subsection{Deep learning models for CR}
Since its inception, the aim of entity resolution research has been to reduce the dependency on hand-crafted features. With the introduction of deep learning in NLP, words could be represented as vectors conveying semantic dependencies~\citep{mikolov2013distributed,pennington2014glove}. This gave an impetus to approaches which deployed deep learning for entity resolution~\citep{wiseman2015learning,wiseman2016learning,clark2016deep,clark2016improving,lee2017end}.

 The first non-linear mention ranking model~\citep{wiseman2015learning} for CR aimed at learning different feature representations for anaphoricity detection and antecedent ranking by pre-training on these two individual subtasks. This approach addressed two major issues in entity resolution: the first being the identification of non-anaphoric references which are abound in text and the second was the complicated feature conjunction in linear models which was necessary because of the inability of simpler features to make a clear distinction between truly co-referent and non-coreferent mentions. This model handled the above issues by introducing a new neural network model which took only raw un-conjoined features as inputs and attempted to learn intermediate representations.
 
The algorithm started with liberal mention extraction using the Berkeley Coreference resolution system~\citep{durrett2013easy} and sought to capture relevant aspects of the task better using representation learning. The authors proposed an extension to the original mention-ranking model using a neural network model, for which the scoring function is defined as: 
\begin{equation}
s(x,y)\triangleq 
\begin{cases}
u^{T}g\Bigg(\begin{bmatrix}
h_{a}(x) \\
h_{p}(x,y) \\

\end{bmatrix}\Bigg)+u_{0}& \text{if $y\neq\epsilon$} \\
v^{T}h_{a}(x)+v_{0} & \text{if $y=\epsilon$} \\
\end{cases}
\end{equation}\\
\begin{equation}
h_a(x)\triangleq tanh(W_a \theta_a (x)+b_a)
\end{equation}
\begin{equation}
h_p(x,y)\triangleq tanh(W_p \theta_p(x,y)+b_p)
\end{equation}
Hence, $h_a$ and $h_p$ represented the feature representations which were defined as non-linear functions on mention and mention-pair features $\theta_a$ and $\theta_b$, respectively, and the function g's two settings were a linear function $g_1$ or a non-linear (tanh) function $g_2$, on the representations. The only raw features defined were $\theta_a$ and $\theta_p$. According to the model, $C'(x)$ corresponded to the cluster the mention belongs to or $\epsilon$ if the mention was non-anaphoric. $y_n^{l}$ corresponded to the highest scoring antecedent in cluster $C'(x)$ and was ${\epsilon}$ if x was non-anaphoric
The neural network was trained to minimize the slack rescaled latent-variable loss which the authors' define as:
\begin{equation}
L(\theta)=\sum_{n=1}^{N} max_{\hat{y} \in \mathcal{Y}(x_n)}\Delta(x_n,\hat{y})(1+s(x_n,\hat{y})-s(x_n,y_n^l))
\end{equation}
$\Delta$ was defined as a mistake specific cost function. The full set of parameters to be optimized was {$W$,$u$,$v$,$W_a$,$W_p$,$b_a$,$b_p$}. $\Delta$ could take on different values based on the type of errors possible in a CR task~\citep{durrett2013easy}, i.e., false link(FL), false new(FN) and wrong link(WL), error types.

The subtask of anaphoricity detection aimed at identifying the anaphors amongst the extracted mentions. Generally the extracted mentions were non-anaphoric, thus this subtask served as an important step to filter out the mentions which needed further processing for antecedent ranking. The pre-trained parameters from this task were used for initializing weights of the antecedent ranking task. The antecedent ranking task was undertaken after filtering the non anaphoric mentions from the antecedent discovery process. The scoring procedure followed was similar to one discussed earlier.

The model was trained on two sets of BASIC~\citep{durrett2013easy} and modified BASIC$+$ raw features. The baseline model used for anaphoricity prediction was an L1-regularized SVM using raw and conjoined features. The baseline model used for subtask two was the neural network based non-linear mention ranking model using the margin-based loss. The proposed neural network based model outperformed the two baseline models for both of the subtasks. The full model (g1 and g2) also achieved the best $F_1$ score with improvement of 1.5 points over the best reported models and 2 over the best mention ranking system(BCS). It outperformed all the state-of-the-art models (as of 2014)~\citep{durrett2013easy,bjorkelund2012data}. 

The first non-linear coreference model which proved that coreference task could benefit from modeling global features about entity clusters~\citep{wiseman2016learning} augmented the neural network based mention-ranking model~\citep{wiseman2015learning} by incorporating entity-level information produced by a recurrent neural network (RNN) running over the candidate antecedent-cluster. This model modified the scoring function of the antecedent ranking model by adding a global scoring term to it. The global score aimed to capture how compatible the current mention was with the partially formed cluster of the antecedent. The clusters were represented using separate weight sharing RNNs which sequentially consumed the mentions being assigned to each cluster. The idea was to capture the history of previous decisions along with the mention-antecedent compatibility. The Clark and Manning algorithm which was proposed roughly during the same time~\citep{clark2016improving} instead defined a significantly different cluster ranking model to induce global information. 

This approach was based on the idea of incorporation of entity-level information, i.e., features defined over clusters of mention pairs. The architecture of this neural network consisted of mainly four sub-parts which were – the mention-pair encoder which passes features (described later) through a feed-forward neural network (FFNN) to produce distributed representations of mentions, a cluster-pair encoder which uses pooling over mention pairs to produce distributed representations of cluster pairs, a mention ranking model to mainly pre-train weights and obtain scores to be used further in cluster ranking and the cluster ranking module to score pairs of clusters by passing their representations through a single-layer neural network.
\begin{figure}[H]
	\centering
	\caption{The Mention-pair and the Cluster-pair encoder~\citep{clark2016improving}}
	\begin{minipage}{0.5\textwidth}
		\centering
		\includegraphics[width=1.2\linewidth]{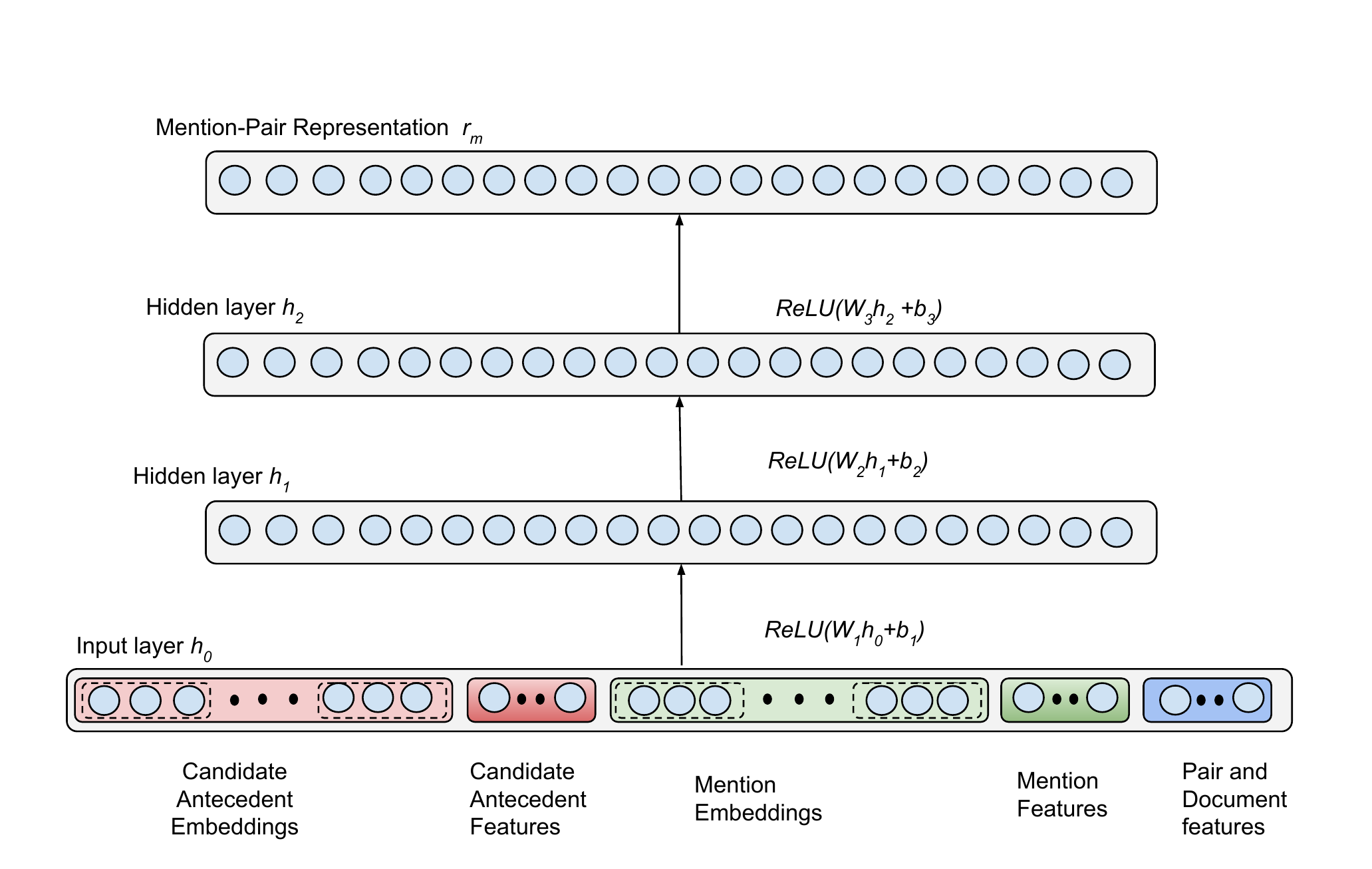}
	\end{minipage}%
	\begin{minipage}{0.5\textwidth}
		\centering
		\includegraphics[width=1.1\linewidth]{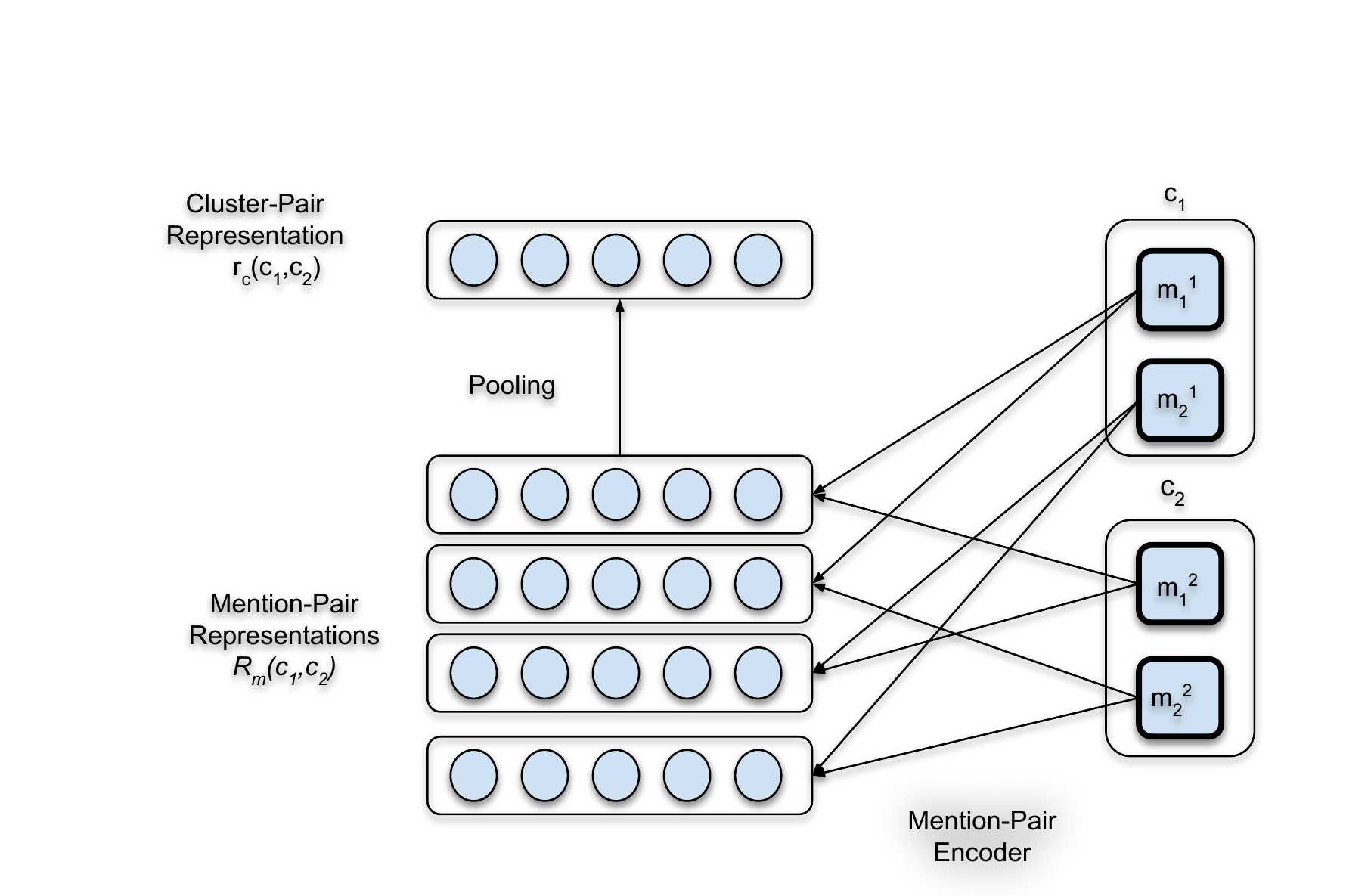}
	\end{minipage}
\end{figure}
The features used for the entire model were: the average of the embeddings of words in each mention, binned distance between the mentions, head word embedding, dependency parent, first word's, last word's and two preceding word's embedding and average of 5 preceding and succeeding words of the mention, the type of mention, position of mention, sub-mention, mention-length, document genre, string match, etc. These features were concatenated into an input vector and fed into a FFNN consisting of three fully-connected hidden rectified linear layers. The output of the last layer was the vector representation of the mention pair. The cluster pair encoder, given the two clusters of mentions $c_i$ ={$m^{i}_1$, $m^{i}_2$, ..., $m^{i}_{|ci|}$} and $c_j$ = {$m^{j}_1$,$ m^{j}_2$, ..., $m^{j}_{|cj|}$}, produces a distributed representation $r_c$($c_i$, $c_j$)$\in$ $R^{2d}$. This matrix was constructed by using max and average pooling over the mention-pair representations. Next, a mention-pair model was trained on the representations produced by the mention pair encoder which servers the purpose of pre-training weights for the cluster ranking task and to provide a measure for coreference decisions. This mention ranking model was trained on the slack rescaled objective~\citep{wiseman2015learning} discussed earlier. The final stage was cluster ranking which used the pre-trained weights of the mention ranking model to obtain a score by feeding the cluster representations of the cluster encoder to a single-layered fully-connected neural network.
The two available actions based on scores were merge (combine two clusters) and pass (no action). During inference, the highest-scoring (most probable) action was taken at each step. This ensemble of cluster ranking beat the earlier state-of-the-art approaches achieving an F1 score of 65.39 on the CoNLL English task and 63.66 on the Chinese task.

Another algorithm which complemented the earlier work~\citep{clark2016deep} attempted at effectively replacing the heuristic loss functions which complicated training, with the reinforce policy gradient algorithm and reward-rescale max-margin objective. This approach exploited the immense importance of independent actions in mention ranking models. The independence of actions implied that the effect of each action on the final result was different thus making this scenario a suitable candidate for reinforcement learning. This model used neural mention ranking model~\citep{clark2016improving} described earlier as the baseline and replaced the heuristic loss with reinforcement learning based loss functions. Reinforcement learning was utilized here to provide a feedback on different set of actions and linkages performed by the mention ranking models. Thus, the model could optimize its actions in such a way that the actions were performed to maximize the reward (called the reward rescaling algorithm). The reward rescaling algorithm achieves an average $F_1$ score of 65.73 and 63.4 on the $CEAF_{\theta4}$ and $B^{3}$ metric respectively on the CoNLL 2012 English Test Data, thus beating the earlier systems. This algorithm was novel because it avoided making costly mistakes in antecedent resolution which could penalized the recall value. On the other hand, unimportant mistakes are not penalized as heavily. The approach is novel with regards to it being the first one to apply reinforcement learning to CR. The most challenging task in this algorithm was the assignment of reward costs which could be corpus specific.

The state-of-the-art model is an end-to-end CR system which outperformed the previous approaches in spite of being dependent on minimal features. This end-to-end neural model~\citep{lee2017end} is jointly modeled mention detection and CR. This model began with the construction of high-dimensional word embeddings to represent the words of an annotated documents. The word embeddings used were a concatenation of Glove, Turian and character embeddings. The character embeddings were learnt using a character-level convolutional neural network (CNN) of three different window sizes. The vectorized sentences of the document were fed into a bidirectional long short-term memory (LSTM) network to learn effective word representations. Next, all the possible mentions in a document were extracted and represented as a one dimensional vector. This mention representation was a conjugation of the start word embedding, head word embedding, end word embedding and some other mention-level features. The head word embedding was learnt using attention mechanism over the entire mention span. The mention representation $g_i$ was defined as:
\begin{equation}
g_{i}=[x^{*}_{START(i)},x^{*}_{END(i)},x\sp{\prime}_{i},\phi(i)]
\end{equation}
where $x\sp{\prime}_{i}$ represented an attention-weighted sum of the word vectors in span i and $x^{*}_{START(i)}$ and $x^{*}_{END(i)}$ are the span boundaries. The approach pruned candidates greedily for training and evaluation and considered only spans of maximum width ten. The mentions were scored using a FFNN and only a fraction of the top scoring spans were preserved further for CR. 
\begin{figure}[H]
	\centering
	  \caption{Bi-LSTM to encode sentences and Mention scoring~\citep{lee2017end}}
		\centering
		\includegraphics[width=\linewidth]{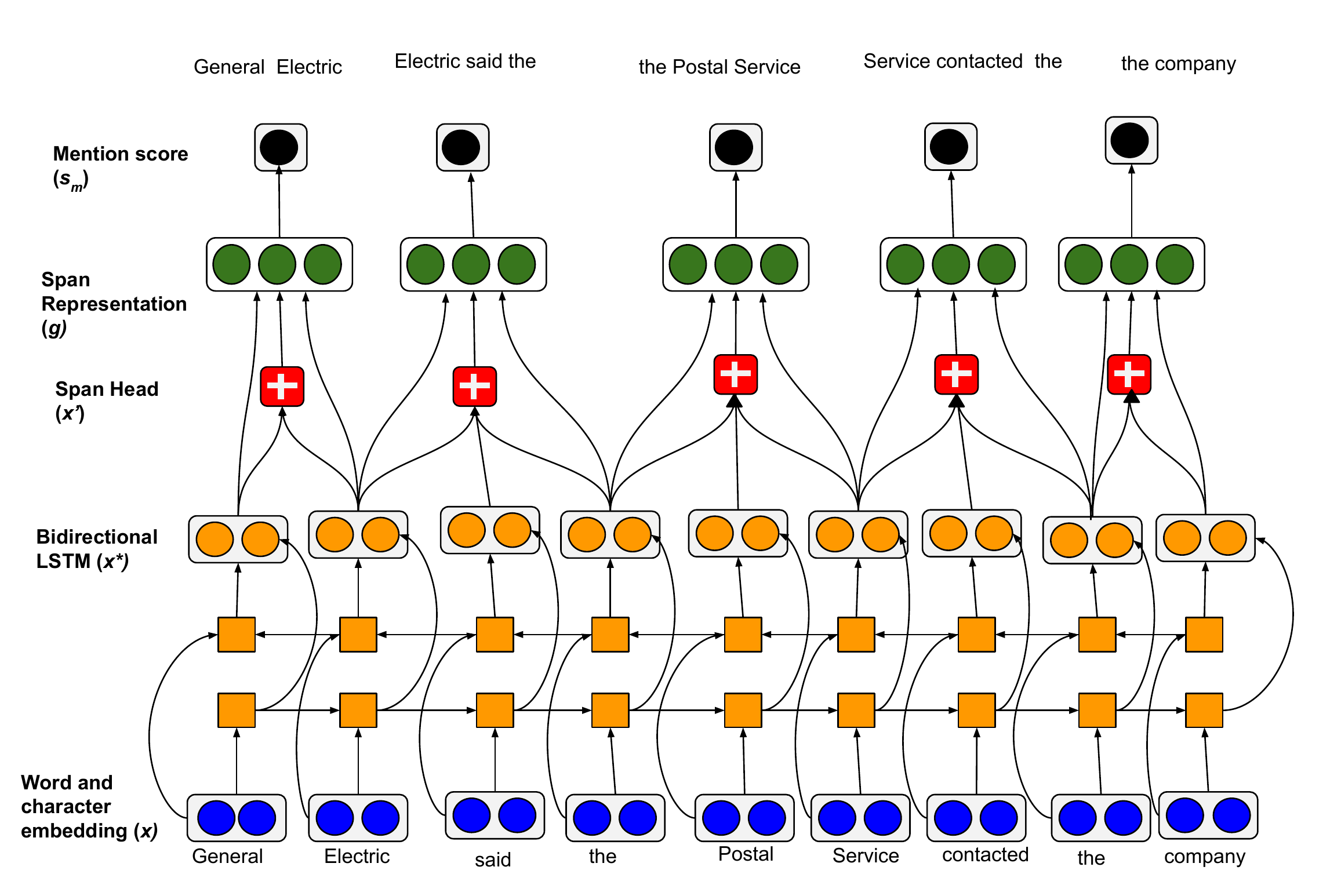}
\end{figure}
These top scoring mentions served as input to the CR model. The preceding 250 mentions were considered as the candidate antecedents. The scores of the mention-antecedent pairs were computed using the equation below. The mention-antecedent pair representation was a concatenation of individual mention representations $g_{i}$ and $g_{j}$, the similarity between the two mentions $g_{i}\circ g_{j}$ and pairwise features $\phi(i,j)$ representing speaker and distance features. The final scoring function optimized is a sum of the of the two individual mention scores of the candidate mentions and the mention-antecedent pair score represented by the equation below.
\begin{equation}
s_{a}(i,j)=w_{a}\cdot FFNN_{a}([g_{i},g_{j},g_{i}\circ g_{j},\phi(i,j)])
\end{equation}
\begin{figure}[H]
	\centering
	\caption{Antecedent Scoring~\citep{lee2017end}}
	\includegraphics[width=0.7\linewidth]{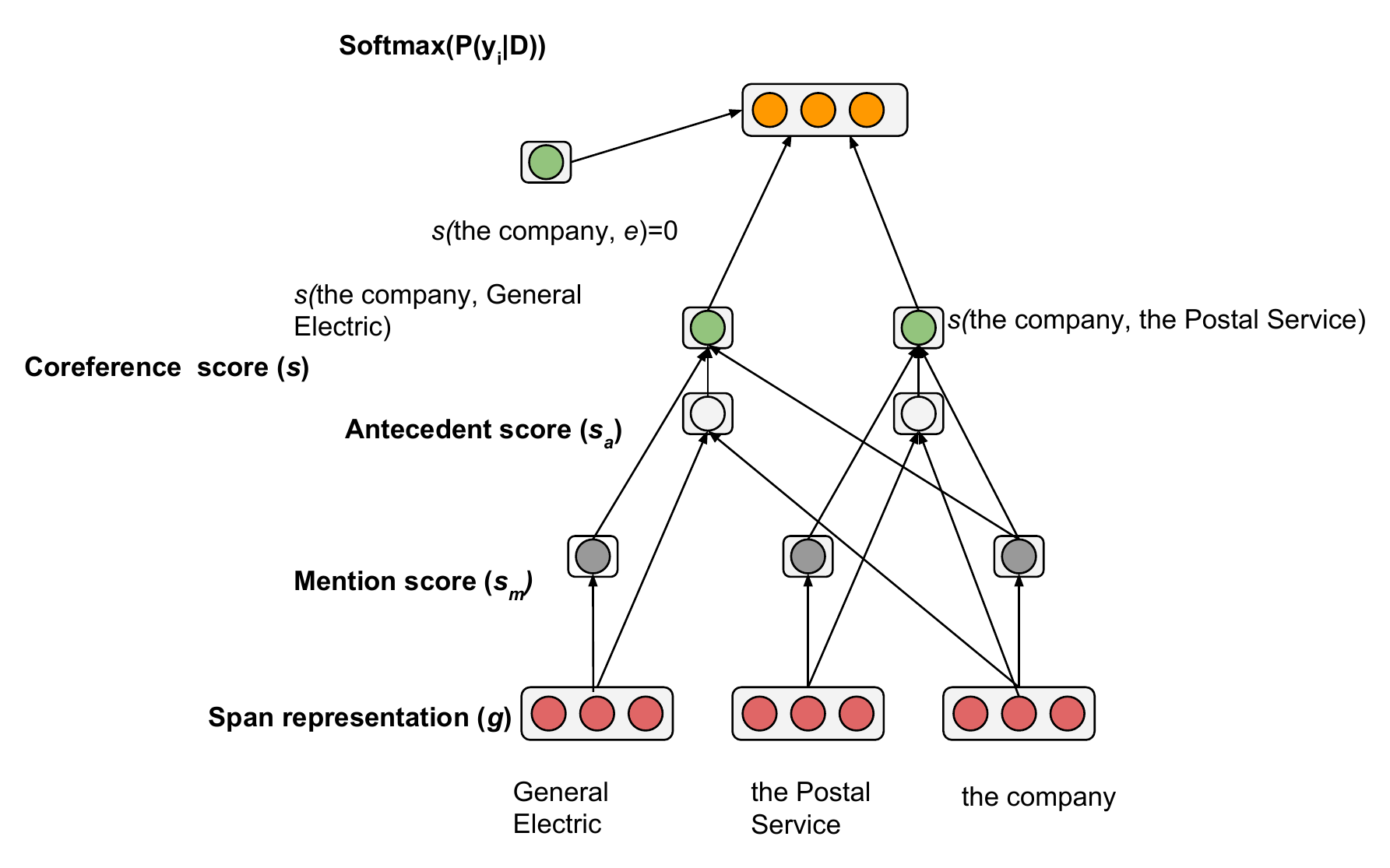}
\end{figure}
 The optimization function used for the model was the marginal log-likelihood of all correct antecedents on basis of the gold-clustering.
\begin{equation}
\log\prod_{i=1}^{N}\sum\limits_{y\sp{\prime}\in\mathcal{Y}\hat{i} GOLD(i)}P(y\sp{\prime})
\end{equation}
During inference the best scoring antecedent was chosen as the most probable antecedent and coreference chains were formed using the property of transitivity.

The authors report the ensembling experiments using five models with different initializations and prune the spans here using average of the mention scores over each model. The proposed approach was extensively evaluated for precision, recall and F1 on the MUC,$B^3$ and CEAF metrics. The authors also provide a quantitative and qualitative analysis of the model for better interpretability.

Challenging aspect of this model was that its high computational time and large number of trainable parameters. This model used a very large deep neural network and, hence, is very difficult to maintain. This creates a challenge for deploying this system as an easy to use off-the-shelf system.
\begin{table}[H]
	\centering
	\caption{Deep learning based entity resolution}
	\label{Table:6}
	\resizebox{\textwidth}{!}{
		\begin{tabular}{llllll}
			\toprule
			Algorithm & \begin{tabular}[c]{@{}c@{}}Neural Network \\ architecture(s) used\end{tabular} & \begin{tabular}[c]{@{}c@{}}Pre-trained Word Embeddings\\      Used\end{tabular} & \begin{tabular}[c]{@{}c@{}}Cluster-level features\\    used (Y/N)\end{tabular} & \begin{tabular}[c]{@{}c@{}}Loss function used for Mention\\         Ranking\end{tabular} & \begin{tabular}[c]{@{}c@{}}External Tools \\    Used\end{tabular} \\ 
			\hline
			~\citep{wiseman2015learning}& FFNN & - & No & \begin{tabular}[c]{@{}l@{}}Heuristic Regularized Slack\\ rescaled latent variable \\ Loss\end{tabular} & \begin{tabular}[c]{@{}l@{}}Berkeley Coreference System\\ for mention extraction and\\ Stanford Coref System's \\Rules for animacy feature\end{tabular} \\ 
			~\citep{wiseman2016learning} & FFNN and RNN & - & Yes & \begin{tabular}[c]{@{}l@{}}Heuristic Slack Rescaled \\ Margin objective\end{tabular} & \begin{tabular}[c]{@{}l@{}}Berkeley Coreference System\\ for mention extraction \\and Stanford deterministic \\system animacy rules\end{tabular} \\ 
			~\citep{clark2016improving}& FFNN & \begin{tabular}[c]{@{}l@{}}English:50d word2vec\\ Chinese:Polyglot 64d\end{tabular} & Yes & \begin{tabular}[c]{@{}l@{}}Heuristic Slack Rescaled \\ Max-margin objective\end{tabular} & \begin{tabular}[c]{@{}l@{}}Stanford Deterministic Coref\\ System rules to extract\\ mentions\end{tabular} \\ 
			~\citep{clark2016deep}& FFNN & \begin{tabular}[c]{@{}l@{}}English:50d word2vec\\ Chinese:Polyglot 64d\end{tabular} & Yes & \begin{tabular}[c]{@{}l@{}}Heuristic Max-margin objective,\\ REINFORCE policy gradient, \\ Reward Rescaling Algorithm\end{tabular} & \begin{tabular}[c]{@{}l@{}}Stanford Deterministic Coref\\ System rules to extract \\mentions\end{tabular} \\ 
			~\citep{lee2017end} & \begin{tabular}[c]{@{}l@{}}FFNN+Bi-LSTM+\\ CNN+Neural Attention\end{tabular} & Glove300d+turian 50d & No & Marginal Log-likelihood & - \\ \bottomrule
		\end{tabular}}
	\end{table}

Deep learning CR systems~\citep{clark2016deep,clark2016improving,lee2017end} represent words using vectors which are known depict semantic relationships between words~\citep{mikolov2013distributed,pennington2014glove}. These models, hence, use less features than the machine learning models. These systems also implicitly capture the dependencies between mentions particularly using RNN and its adaptations like LSTMs and gated recurrent units (GRUs). One disadvantage of these systems is that they are difficult to maintain and often require some amount of genre or domain specific adaptation before use. Amongst the deep learning based CR algorithms discussed earlier we observe that the dependency on features decreased over time. This was mainly because of the pre-trained word embeddings which captured some amount of semantic similarity between the words. Unlike the Stanford deep coref system~\citep{clark2016improving,clark2016deep}, the state-of-the-art system used minimal mention-level and mention-antecedent pair features. This system also did not use any external mention-extractor and proceeds by extracting all possible spans up to a fixed width, greedily. Another advantage of this system was that it did not use any heuristic loss function unlike the other deep learning models~\citep{wiseman2015learning,wiseman2016learning,clark2016improving,clark2016deep} and still managed to beat the earlier models with a very simple log-likelihood loss function. The previous models used heuristic loss functions which were dependent on a mistake specific cost function whose values were set after exhaustive experiments. Though this system is difficult to maintain mainly because of its high-dimensionality, it is a strong evidence of the effectiveness of LSTMs and their ability to capture long term dependencies. Another possible disadvantage of this model is that it is still basically a mention ranking model and chooses the highest scoring antecedent without using any cluster-level information. As posited by many earlier deep learning works which used cluster-level information~\citep{clark2016improving,wiseman2016learning} this information is necessary to avoid linking of incompatible mentions to partially formed coreference chains. In spite of some of the disadvantages of the deep CR systems, future strides in CR can only be achieved by either defining better features to be learnt or by introducing better deep learning architectures for CR. 
\section{An analysis of entity resolution research progress on different datasets}
In previous sections, we have discussed several types of entity resolution algorithms. In this section, we aim at providing an overview of the research progress made in the field of entity resolution over the past few years. Here, we will be analyzing the progress made on mainly three important publicly available datasets: the MUC corpus, the ACE corpora and the CoNLL Shared task (OntoNotes 5.0).

The MUC datasets were the first annotated corpora to be publicly available for CR. Though these datasets were small they have been widely used for evaluation and training. The first system to be evaluated on the MUC dataset was Soon's mention-pair model~\citep{soon2001machine}. This was followed by Ng and Cardie's series of improvements on the dataset~\citep{ng2002improving,ng2002combining}. Followed by these were some mention ranking models have also been attempted on the MUC datasets one of them was Yang's twin candidate model~\citep{yang2008twin} which aimed at capturing competition between between the antecedents. Conditional random fields (CRFs)~\citep{mccallum2005conditional} have also been trained on this dataset to directly model global coreference chains. In addition, some other approaches like Integer Linear Programming~\citep{finkel2008enforcing} and non-parametric Bayesian models~\citep{haghighi2007unsupervised} have also been attempted. The MUC-6 and 7 datasets in spite of being widely popular were quite small in size thus making training on a small corpus very hard. This also meant that some types of references were ignored. The evaluation standards were also not very well defined, hence making comparison of different algorithms a challenge. The ACE and CoNLL datasets aimed at overcoming these disadvantages by providing a much larger training corpus.

When coming to the ACE datasets, we observe a huge disparity in the evaluation standards, train-test splits and metrics used. This was mainly because the test sets of the dataset were not publicly released and, hence, were unavailable to non-participants. This made comparative evaluation with research methodologies which did not participate in this task difficult. Many researchers were hence forced to define their own train-test split~\citep{culotta2007first,bengtson2008understanding}. In addition, the ACE datasets were also released in iterations and phases from 2002 to 2005, thus algorithms tested on newer releases could not be directly compared with the earlier approaches. Multiple algorithms were evaluated on different versions of the ACE datasets like the mention pair models~\citep{bengtson2008understanding}, mention ranking models~\citep{denis2008specialized,rahman2009supervised} and joint inference models~\citep{poon2008joint,haghighi2010coreference}. Some rule-based approaches~\citep{lee2011stanford} were also tested on the ACE datasets mainly with the aim of comparison with past research methodologies, which were not evaluated on the newly introduced CoNLL shared task. 

The best performing rule-based systems on the current version of the CoNLL shared task is the multi-sieve based Stanford deterministic system~\citep{lee2013deterministic}. Most of the early systems which outperformed the rule-based system were machine learning based. There have been multiple variants of the mention-pair models which used structured perceptron models for CR on the CoNLL 2012 dataset~\citep{fernandes2012latent,chang2013constrained,durrett2013easy}. This was followed by a model which jointly modeled Coreference, Named Entity Recognition and Entity Linking using a structured CRF. Models which used cluster-level features were very well known in CR by then and some models~\citep{ma2014prune} also used average across all pairs in clusters involved to incorporate cluster-level features. The entity-centric model~\citep{clark2015entity} which achieved a vast margin of improvement on the earlier systems proposed a novel way of defining cluster-level features. It combined information from involved mention pairs in variety of ways with higher order features produced from scores of the mention pair models. As observed from the table below, since 2015 the best performing systems on CoNLL 2012 datasets have been deep learning systems. These used neural networks to individually model different subtasks like antecedent ranking and cluster ranking or to jointly model mention prediction and CR. Though the most common use of deep neural networks in CR has been for scoring mention-pairs and clusters, some methods~\citep{wiseman2016learning} also used RNN's to sequentially store cluster states, with the aim of modeling cluster-level information. The current best performing system on the CR task is a very deep end to end system which is an amalgamation of LSTM, CNN, FFNN and neural attention. Since deep learning systems are typically hard to maintain some recent systems have also proposed a hybrid of rule-based and machine learning systems~\citep{lee2017scaffolding}. Though this system does not perform at par with the deep learning system, it is easy to maintain and use and even outperforms some of the machine learning systems. Overall, the deep learning trend in CR looks very exciting and future progress could be expected by incorporating better features and using more sophisticated neural network architectures. As stated by many~\citep{lee2017end,durrett2013easy}, this could be modeled by developing an intuitive way to incorporate differences between entailment, equivalence and alteration phenomenon.  

As observed from the table below, until about a few years ago we observe that the CR datasets and mainly the evaluation metrics were not standardized. This made comparison of algorithms very difficult. The early corpora like MUC and ACE did not release very strict evaluation guidelines for the task. Also, there were multiple releases, few of which were publicly available. The test datasets of the ACE corpora were initially not available to non-participants which also created issues with the comparison of algorithms. Hence, most authors often defined train and test splits of their own on the datasets~\citep{culotta2007first,bengtson2008understanding}. Though future approaches tried to stick to the earlier train-test splits for comparative evaluation~\citep{haghighi2009simple}, it was difficult as often the datasets needed for comparison were not freely available. Another issue was with the very definition of the Coreference Task. Some approaches~\citep{luo2005coreference} which reported highest accuracy on the ACE and MUC corpus could not be compared with others because they reported performance on true labelled mentions instead of system predicted mentions. This was different from other approaches which jointly modeled the tasks of mention prediction and CR. This, however, changed with the introduction of CoNLL 2012 shared task~\citep{pradhan2012conll} which defined strict evaluation guidelines for CR. After this, CR research gained momentum and has seen more consistent progress and clearer evaluation standards. 
	\begin{landscape}
		\begin{table}[htbp]
			\centering
			\caption{Dataset-wise comparison of baselines}
			\label{table:7}
			\resizebox{\textwidth}{!}{
				\begin{tabular}{llllllll}
					\toprule
					\multicolumn{1}{c}{\multirow{2}{*}{Dataset}} & \multirow{2}{*}{Release} & \multicolumn{1}{c}{\multirow{2}{*}{Algorithm}} & \multicolumn{4}{c}{Scoring metrics F1 values} & \multicolumn{1}{c}{\multirow{2}{*}{\begin{tabular}[c]{@{}c@{}}Algorithm \\  Type\end{tabular}}} \\
					\hline 
					\multicolumn{1}{c}{} & & \multicolumn{1}{c}{} & MUC & B-cubed & CEAFe & CoNLL & \multicolumn{1}{c}{} \\ 
					\multirow{17}{*}{\begin{tabular}[c]{@{}l@{}}CoNLL Shared Task\\  (OntoNotes 5.0)\end{tabular}} & \multirow{3}{*}{CoNLL 2011} &~\citep{lee2011stanford} & 61.51 & 63.27 & 45.17 & 56.65 & Rule-based \\ 
					& &~\citep{bjorkelund2012data} & 64.71 & 64.73 & 45.35 & 58.26 & \multirow{2}{*}{Machine Learning} \\ 
					& &~\citep{durrett2013easy} & 66.43 & 66.16 & 47.79 & 60.13 & \\ \cline{3-8}
					& \multirow{14}{*}{CoNLL 2012} &~\citep{lee2013deterministic} & 63.72 & 52.08 & 48.65 & 54.82 & Rule-based \\ 
					& &~\citep{chang2013constrained} & 69.48 & 57.44 & 53.07 & 60.00 & Probabilistic \\ 
					& &~\citep{fernandes2012latent} & 70.51 & 57.58 & 53.86 & 60.65 & \multirow{6}{*}{Machine Learning} \\ 
					& &~\citep{durrett2013easy} & 70.51 & 58.33 & 55.36 & 61.40 & \\ 
					& &~\citep{ma2014prune} & 72.84 & 57.94 & 53.91 & 61.56 & \\ 
					& &~\citep{bjorkelund2012data} & 70.72 & 58.58 & 55.61 & 61.63 & \\ 
					& &~\citep{durrett2014joint} & 71.24 & 58.71 & 55.18 & 61.71 & \\ 
					& &~\citep{clark2015entity} & 72.59 & 60.44 & 56.02 & 63.02 & \\ 
					& &~\citep{lee2017scaffolding} & 72.37 & 60.46 & 56.76 & 63.20 & Hybrid=ML+Rules \\ 
					& &~\citep{wiseman2015learning}& 72.6 & 60.52 & 57.05 & 63.39 & \multirow{5}{*}{Deep Learning} \\ 
					& &~\citep{wiseman2016learning}& 73.42 & 61.50 & 57.7 & 64.21 & \\ 
					& &~\citep{clark2016improving} & 74.06 & 62.86 & 58.96 & 65.29 & \\ 
					& &~\citep{clark2016deep} & 74.56 & 63.40 & 59.23 & 65.73 & \\ 
					& &~\citep{lee2017end}& 77.20 & 66.60 & 62.60 & 68.80 & \\ \hline
					\multirow{21}{*}{Automatic Content Extraction} & \multirow{7}{*}{ACE 2004 Culotta Test} &~\citep{stoyanov2010coreference} & 62.0 & 76.5 & - & - & Machine Learning \\
					& & \begin{tabular}[c]{@{}l@{}}~\citep{haghighi2009simple}\\ (true)\end{tabular} & 79.60 & 79.00 & - & - & Rule-based \\ 
					& & \begin{tabular}[c]{@{}l@{}}~\citep{haghighi2009simple}\\ (system)\end{tabular} & 64.4 & 73.2 & - & - & Rule-based \\ 
					& &~\citep{haghighi2010coreference} & 67.0 & 77.0 & - & - & Machine Learning \\ 
					& &~\citep{culotta2007first} & - & 79.30 & - & - & Probabilistic \\ 
					& &~\citep{bengtson2008understanding} & 75.80 & 80.80 & - & - & Machine Learning \\ 
					& &~\citep{lee2013deterministic} & 75.90 & 81.00 & - & - & Rule-based \\ \cline{3-8}
					& \multirow{5}{*}{\begin{tabular}[c]{@{}c@{}}ACE 2004 -training datasets\\ BNEWS,NWIRE\end{tabular}} &~\citep{haghighi2007unsupervised} & 62.3,64.2 & - & - & - & Rule-based \\ 
					& &~\citep{finkel2008enforcing} & 67.1,61.1 & 74.5,73.1 & - & - & \begin{tabular}[c]{@{}l@{}}ML+ Integer Linear \\ Programming\end{tabular} \\ 
					& &~\citep{poon2008joint} & 70.90,67.3 & - & - & - & Machine Learning \\ 
					& & \begin{tabular}[c]{@{}l@{}}~\citep{haghighi2009simple}\end{tabular} & 76.50,- & 76.90,- & - & - & Rule-based \\ 
					& &~\citep{lee2013deterministic} & 79.60,- & 80.20,- & - & - & Rule-based \\ \cline{3-8} 
					& \multirow{3}{*}{\begin{tabular}[c]{@{}c@{}}ACE -Phase 2 Test sets\\ BNEWS,NWIRE,NPAPER\end{tabular}} &~\citep{ng2005machine} & 64.9,54.7,69.3 & 65.6,66.4,66.4 & -& -& \multirow{4}{*}{Machine Learning} \\ 
					& &~\citep{denis2009global} & 69.2,67.5,72.5 & - & -& -& \\ 
					& -&~\citep{poon2008joint} & 67.4,67.4,70.4 & 67.7,71.6,68.2 &-&-& \\ \cline{3-8}
					& \multirow{3}{*}{ACE 2005 Stoyanov Test} &~\citep{stoyanov2010coreference} & 67.4 & 73.7 & -& -& \\ 
					& &~\citep{haghighi2009simple} & 65.2 & 71.8 & -& -& Rule-based \\ 
					& &~\citep{haghighi2010coreference} & 68.1 & 75.1 & -& -& \multirow{2}{*}{Machine Learning} \\ 
					& \multirow{3}{*}{ACE 2005 Rahman and Ng} &~\citep{rahman2009supervised}& 69.3 & 61.4 & - & -& \\ 
					& &~\citep{haghighi2009simple} & 67.0 & 60.6 & -& -& Rule-based \\ 
					& &~\citep{haghighi2010coreference}& 71.6 & 62.7 & -& -& Machine Learning \\ \hline 
					\multirow{15}{*}{\begin{tabular}[c]{@{}l@{}}Message \\ Understanding\\ Conference\end{tabular}} & \multirow{11}{*}{MUC 6} &~\citep{soon2001machine} & 62.6 & - & - & - & \multirow{6}{*}{Machine Learning} \\
					& &~\citep{ng2002combining} & 69.5 & - & - & - & \\ 
					& &~\citep{ng2002improving}& 70.4 & - & - & - & \\ 
					& &~\citep{yang2003coreference}& 71.3 & -& & -& \\ 
					& &~\citep{mccallum2005conditional} & 73.4 & -& -& -& \\ 
					&&~\citep{haghighi2007unsupervised} & 63.9 & -& -& -& \\ 
					& &~\citep{finkel2008enforcing} & 68.30 & 64.30 & - & - & \begin{tabular}[c]{@{}l@{}}ML+ Integer Linear \\ Programming\end{tabular} \\ 
					& &~\citep{stoyanov2010coreference} & 68.5 & 70.88 & -& -& \multirow{2}{*}{Machine Learning} \\
					& &~\citep{poon2008joint}& 79.20 & - & - & - & \\ 
					& &~\citep{haghighi2009simple} & 81.90 & 75.0 & - & - & \multirow{2}{*}{Rule-based} \\ 
					& &~\citep{lee2013deterministic}& 78.40 & 74.40 & - & - & \\ \cline{3-8}
					& \multirow{4}{*}{MUC 7} &~\citep{soon2001machine} & 60.4 & -& -& -& \multirow{4}{*}{Machine Learning} \\ 
					& &~\citep{ng2002improving} & 63.4 & -& -& -& \\ 
					& &~\citep{yang2003coreference} & 60.2 & -&- & -& \\ 
					& &~\citep{stoyanov2010coreference} & 62.8 & 65.86 & -& -& \\ 
					\bottomrule
				\end{tabular}}
			\end{table}
		\end{landscape}
 \section{Open source tools}
 
 Computer science has now stumbled upon an era wherein sharing research output is both a demand and necessity. On the one hand, it helps the researchers to think about possible improvements from peer suggestions and, on the other hand, it allows researchers mainly interested in its application to pick an off-the-shelf model. Given the wide range of applications of CR, practical tools to tackle this issue are a necessity. These tools may deploy a specific approach like~\citep{mitkov2002new} and or others like Reconcile~\citep{stoyanov2010coreference} could be a combination of many research methodologies.
 
 The GuiTAR tool~\citep{poesio2004general} aimed at making an open source tool available for researchers mainly interested in applying AR to NLP applications like question answering, summarization, sentiment analysis and information retrieval. This tool has primarily been developed for the tasks of segmentation and summarization. This is a domain dependent AR tool.
 Stanford coref toolkit provides 3 models which were pioneered by the Stanford NLP group. These three algorithms are Deterministic~\citep{lee2013deterministic, recasens2013life,raghunathan2010multi}, Statistical~\citep{clark2015entity} and Neural~\citep{clark2016deep,clark2016improving}. BART~\citep{versley2008bart} is one of the few highly modular toolkit for CR that
 supports the statistical approaches. BART is multilingual and is available for
 German, English and Italian. BART relies on a maximum entropy model for
 classification of mention-pairs. Currently, it supports a novel Semantic Tree based approach (for English).
 As mentioned earlier, coreference constraints and coreference types varies from language to language and BART aims at separation of linguistic and machine learning aspects of the problem. BART proceeds by converting input document into a set of linguistic layers represented by separate XML layers. They are used to extract mentions, assign syntactic properties and define pairwise features for the mention. A decoder generates training examples through sample selection and learns pairwise classifier. The encoder generates testing examples through sample selections and partitions them based on trained coreference chains. This toolkit aimed at combining the best state-of-the-art models into a modular
 toolkit which has been widely used for broader applications of AR. ARKref~\citep{o2013arkerf} is a tool for NP CR that is based on system described by Haghighi and Klein~\citep{haghighi2009simple}. ARKref is deterministic, rule-based that uses syntactic information from a constituency parser, semantic information from an entity recognition component to constraint the set of possible antecedent candidate that could be referred by a given mention. It was trained and tested on CoNLL shared task~\citep{pradhan2012conll}. 
The Reconcile System~\citep{stoyanov2010coreference} solved a problem of comparison of various CR algorithms. This problem mainly arises due to high cost of implementing a complete end to end CR system, thus giving way to inconsistent and often unrealistic evaluation scenarios. Reconcile is an infrastructure for development of learning based NP CR system Reconcile can be considered an amalgamation of rapid creation of CR systems, easy implementation of new feature sets and approaches to CR and empirical evaluation of CR across a variety of benchmark datasets and scoring metrics.
 Reconcile is evaluated on six most commonly used CR datasets (MUC-6, MUC-7 ACE, etc.). Performance is evaluated according to B\textsuperscript{3} and MUC scoring metrics. Reconcile aims to address one of the issues in AR~\citep{mitkov2001outstanding}, which is the huge disparity in the evaluation standards used. It further makes an attempts to reduce the labelling standards disparity too.
		\begin{landscape}
			
			\setlength\tabcolsep{1pt}
			\footnotesize
			\begin{table}[htbp]
				\caption{Off-the-shelf entity resolution systems}
				\label{table:6}
				\begin{minipage}{\columnwidth}
					\begin{center}
						\begin{tabular}{lllll}
							\toprule
							Toolkit & Algorithm & \begin{tabular}[c]{@{}l@{}}Development and\\ Evaluation Corpus\end{tabular} & Languages & Type \\ 
							\hline
							\begin{tabular}[c]{@{}l@{}}GuiTAR\\~\citep{poesio2004general}\end{tabular} & \begin{tabular}[c]{@{}l@{}}General toolkit which \\ incorporates many \\ algorithms like\\~\citep{mitkov2002new},\\~\citep{vieira2005coreference}, etc. and \\can be extented to include\\ other algorithms\end{tabular} & GNOME corpus & English & \begin{tabular}[c]{@{}l@{}}Hybrid: Rule-based+\\ Machine Learning\end{tabular} \\ 
							\begin{tabular}[c]{@{}l@{}}BART \\~\citep{versley2008bart}\end{tabular} & \begin{tabular}[c]{@{}l@{}}\\Primarily~\citep{soon2001machine}\\ and some other \\ machine learning \\ approaches to CR\end{tabular} & ACE-2 corpora & English & Machine Learning \\ 
							\begin{tabular}[c]{@{}l@{}}PARKref \\~\citep{o2013arkerf}\end{tabular} & \begin{tabular}[c]{@{}l@{}}PHaghini and Klein Model \\~\citep{haghighi2009simple}\end{tabular}& \begin{tabular}[c]{@{}l@{}}ACE2004-ROTH-DEV\\ ACE2004-CULOTTA\\ -TEST\end{tabular} & \begin{tabular}[c]{@{}l@{}}German, \\English and\\ Italian \end{tabular}& Rule-based \\ 
							Reconcile~\citep{stoyanov2010coreference} & \begin{tabular}[c]{@{}l@{}}Abstracts the basic \\ architecture of most \\ contemporary \\ supervised learning-\\ based coreference \\ resolution systems\\ e.g.,~\citep{soon2001machine},\\~\citep{ng2002improving},\\~\citep{bengtson2008understanding}\end{tabular} & \begin{tabular}[c]{@{}l@{}}2 MUC datasets (MUC-6 \\ and MUC-7) \\4 ACE datasets\end{tabular} & English & \begin{tabular}[c]{@{}l@{}}Supervised Machine\\ Learning Classifiers\end{tabular} \\ 
							\begin{tabular}[c]{@{}l@{}}Stanford CoreNLP deterministic,\\ rule-based system\end{tabular} &~\citep{lee2013deterministic}& \begin{tabular}[c]{@{}l@{}}CoNLL 2011 (OntoNotes \\ Corpus), ACE2004-\\Culotta- Test, ACE \\ 2004-nwire, MUC6-Test\end{tabular} & \begin{tabular}[c]{@{}l@{}}Chinese \\ English\end{tabular} & Rule-based \\ 
							\begin{tabular}[c]{@{}l@{}}Stanford Core NLP Statistical\\ System\end{tabular} &~\citep{clark2015entity} & CoNLL 2012 & English & Statistical \\ 
							\begin{tabular}[c]{@{}l@{}}Stanford CoreNLP Neural\\ Coreference Resolution\end{tabular} &~\citep{clark2016improving,clark2016deep}& CoNLL 2012 & English & Deep Neural Network \\ \bottomrule
						\end{tabular}
					\end{center}
				\end{minipage}
			\end{table}
		\end{landscape}
		
\section{Reference Resolution in sentiment analysis}
Being one of the core component of NLP, CR has many potential downstream applications in NLP like machine translation~\citep{hardmeier2010modelling,werlen2017using,bawden2017evaluating}, paraphrase detection~\citep{recasens2010paraphrase,regneri2012using}, summarization~\citep{bergler2003using,witte2003fuzzy}, question answering~\citep{vicedo2000importance,weston2015towards,youaug}, sentiment analysis~\citep{camnt5,valcon}, etc. The application of most interest to us is sentiment analysis. Though AR is said to be one of the most commonly faced challenge in sentiment analysis, we observe that there is scarcity of research work targeting the question of how could CR be effectively incorporated into a sentiment analysis system and which issues will it most likely solve. In this section, we provide a background of the scenarios in sentiment analysis which necessitate entity resolution. We also discuss some of the prominent approaches in the intersection of these two fields. 

On analyzing the use of AR in sentiment analysis, we come across two main scenarios where AR can prove beneficial to sentiment analysis. The first place where this could prove beneficial is for \textquotedblleft global entity resolution\textquotedblright{}. An often observed phenomenon in reviews which are used for sentiment analysis is that the reviews are often centered around one particular entity which is trivial. Hence, most reviewers do not explicitly specify the entity that they are reviewing. Some others do specify this entity. Thus, this cross-review information can be exploited effectively to resolve the pronominal references to the global entity. An example of this taken from the SemEval aspect-based sentiment analysis dataset is depicted in the figure below. In the example below, multiple reviews could be used to chain the references to a global entity (i.e., HP Pavillion Laptop). Global entity resolution can aid the process of extracting the sentiment associated with the general entity.
\begin{figure}[H]
		\centering
		\includegraphics[width=0.8\linewidth,height=4.5cm]{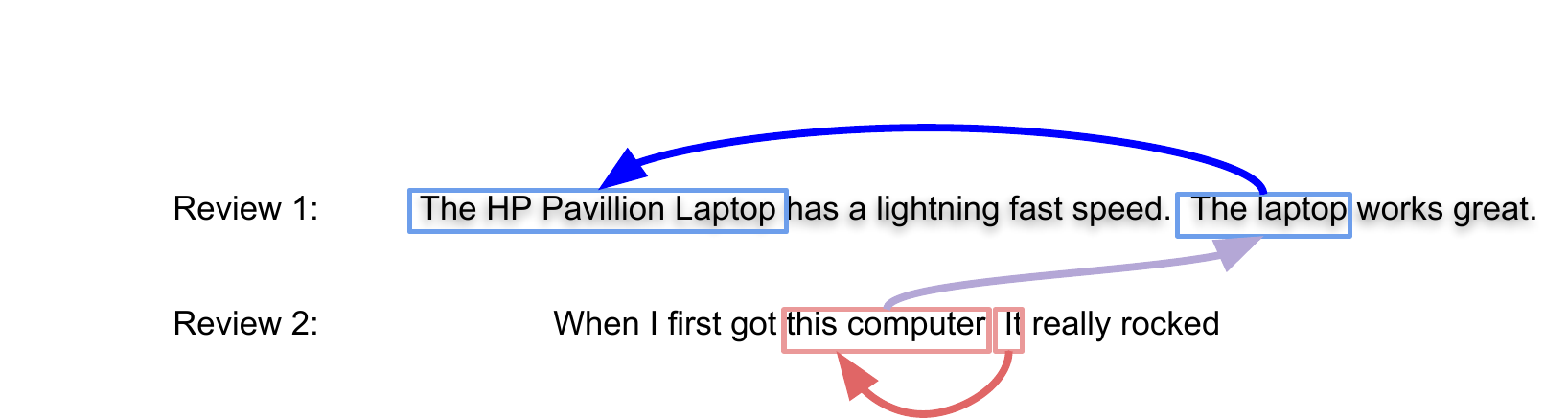}
		\caption{Global Entity Resolution (\href{https://huggingface.co/coref/}{Neural Coref Api})}
		\label{fig:7}
\end{figure}
Another possible use of AR in sentiment analysis is mainly for fine-grained aspect-based sentiment analysis~\citep{maatar}. AR can help infer multiple pronominal references to a particular aspect of the product.This in turn can help extract the opinion associated with that particular aspect. An example of this can be seen in the figure below. In the example below, the resolution of the pronouns to the aspects as depicted by the links between them could aid in the procedure of extraction of fine grained opinion on the aspect. These two images were the resolved references returned by the \href{https://huggingface.co/coref/}{hugging face api } which deploys the Stanford Deep Coref System~\citep{clark2016deep}. 
\begin{figure}[H]
		\centering
		\includegraphics[width=0.7\linewidth,height=4.5cm]{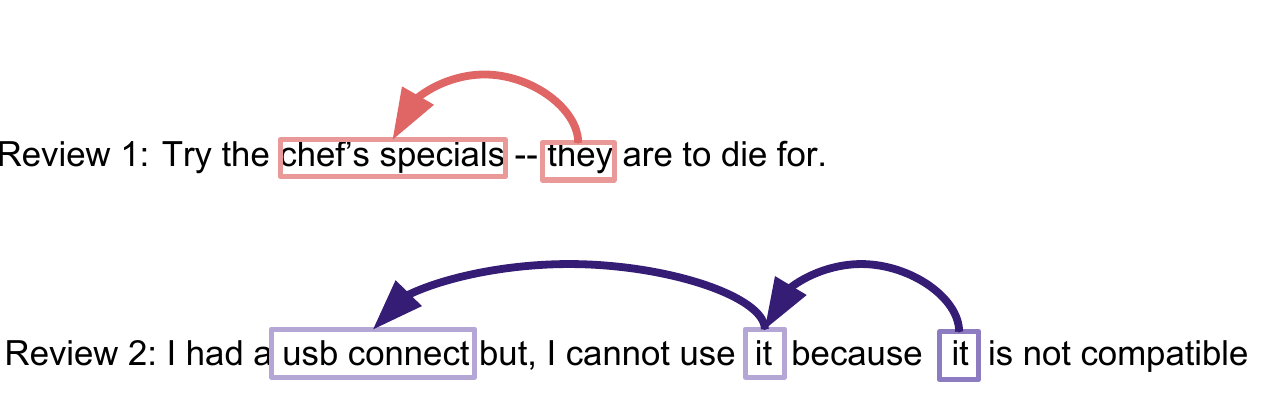}
		\caption{Fine-grained aspect resolution (\href{https://huggingface.co/coref/}{Neural Coref Api})}
		\label{fig:8}
\end{figure} 
Now that we have established the importance of AR for sentiment analysis, we will be providing an overview of the approaches which have worked at the intersection of these two fields. The importance of AR in sentiment analysis has been delineated in many significant research works which consider sentiment analysis as a suitcase research problem~\citep{camsui}. 

AR and CR enables sentiment analysis to break beyond the sentence-level opinion mining task. A recent approach which targets this~\citep{le2016sentiment} addresses the problem of aspect extraction which is a crucial task for aspect-based sentiment analysis. AR in sentiment analysis aids the task of extracting aspect-opinion word pair. This approach to aspect extraction proceeds by construction of sentiment ontology knowledge base. This is followed by lightweight NLP on the text. There is Ontological resolution engine constructed which discovers the implied by and mentioned by relations in aspect-based sentiment ontology. The sentiment rating engine (i.e., assigns rating (polarity) to the aspect extracted.

Another paper~\citep{nicolov2008sentiment} aimed at investigating whether a performance boost is obtained on taking coreference information into account in sentiment analysis. Take, for example, \emph{The canon G3 power shot has impressed me. This camera combines amazing picture quality with ease of use.} As human annotators it is easy for us to understand that the term camera here co-refers with canon G3 power shot. However, this task is a major challenge faced by most algorithms. The sentiment analysis algorithm introduced here is proximity-based for focused sentiment identification. If first calculates the anchor-level sentiment by considering sentiment window of 8 tokens before and after a phrase using distance weighting approach. The anchor weighted scores are aggregated and sentiment phrases are created. Finally, the co-referring entities are identified and the algorithm is evaluated over an opinionated corpus. The percentage improvement obtained over baseline CR modules is on an average 10\% and varies over different datasets used for evaluation.

Another algorithm~\citep{jakob2010using} aimed at tackling the issue of extracting opinion targets expressed by anaphoric pronouns. Opinion word and target pair extraction can benefit from AR to a great extent. The algorithm presented by~\citep{zhuang2006movie} is used as a baseline for the experiment using opinion target and opinion word extraction. A modified version of CogNIAC~\citep{baldwin1997cogniac} is used for CR. The best configuration of this algorithm reaches approximately 50\% of the improvements which are theoretically possible with perfect AR.
	
Another recent interesting work~\citep{ding2010resolving} posits that object and attribute co-reference is important because without solving it a great deal of opinion information will be lost and opinions may be assigned to wrong entities. Major loss encountered in case pronouns are not resolved is of opinion words. The paper elicits the importance of the issue with an example:
\emph{I bought the cannon S500 camera yesterday. It looked beautiful. I took a few photos last night. They were amazing.} In the example, the last two sentences express opinions but it is difficult to specify the target at which the opinion is aimed. Target extraction becomes meaningless if the association between the target and the opinion word is not captured appropriately or is obscure due to co-referent phrases. The paper describes two basic entities object and attribute, e.g., camera (object) and Picture quality (attribute). The pairwise learning approach followed is supervised model based on~\citep{soon2001machine} CR feature model and the annotation is performed as per the MUC-7 standards. The datasets used are blog conversations on products of myriad categories like dvd, cars, TV, lcd, etc. The algorithm first pre-processes the text and then constructs features in a way similar to~\citep{soon2001machine} with addition of some other features like sentiment consistency, comparative sentences and entity-opinion word pair association. Pre-processing POS, NP finder. A decision tree is trained on these features and the algorithm is tested on an un-annotated dataset.

As observed from the research methodologies discussed earlier, the amalgamation of entity resolution systems into a sentiment analysis systems is a challenging task. This is further accentuated by the fact that current entity resolution systems are themselves not perfect and resolving references before sentiment analysis could in fact prove detrimental to the performance of sentiment analysis systems if not incorporated correctly. Future research methodologies in this area should focus on more exhaustive evaluations on standard sentiment analysis datasets.

\section{Reference Resolution: Issues and Controversies}
In this section, we will be discussing the major issues and controversies spanning the area of entity resolution. Upon a thorough investigation of entity resolution research there have been three main areas of debate in this field: the evaluation metrics used, the scope of the datasets used and the idea of commonsense knowledge induction in entity resolution. We will be providing an overview of these issues and the progress made in addressing them. 

The issues with the evaluation metrics to be used for CR have been delineated by many prominent researchers~\citep{ng2010supervised,mitkov2001outstanding}. We have now progresses from evaluation using simple metrics like \textquotedblleft Hobb's Accuracy Metric\textquotedblright{} to much more advanced metrics like MUC~\citep{vilain1995model}, $B^3$~\citep{bagga1998algorithms} and CEAF~\citep{luo2005coreference} which capture the entirety of the CR task. In spite of the progress made over the years, as pointed out by many researchers, the metric currently used for CR~\citep{pradhan2012conll} is still fundamentally the average of three faulty metrics (i.e., MUC, $B^3$ and CEAF). Recently, there have been metrics proposed to circumvent the issues faced by the earlier metrics. Some of these include modifications on existing metrics~\citep{cai2010evaluation} and the other the new LEA metric~\citep{moosavi2016coreference}. We encourage researchers to evaluate their models on these recently proposed metrics in addition to the earlier standard metrics.\\
Another area pertaining to CR research is whether the standard datasets for the task address different types of references that exist in natural language. As discussed earlier the field of entity resolution is composed of different types of references. Some of these references are rare and some types are not labelled by the current CR datasets~\citep{zeldes2016annotation}. This has led to proliferation of research targeting specific types of references like multi-antecedent references~\citep{vala2016more}, Abstract Anaphora~\citep{marasovic2017mention} and One Anaphora~\citep{goldberg2017one}. We suggest that to address this issue future datasets released clearly specify the types of references considered for labelling and the ones not. We also encourage future CR models to carry out cross-domain evaluations on other datasets which are also annotated in CoNLL format like the Character Identification dataset~\citep{moosavi2016coreference}. This will mainly aid in the process of identifying the types of references which still pose a challenge to the state-of-the-art CR algorithms.

Since the inception of the field of CR it has been known that some type of references are extremely hard to resolve for a machine mainly because it requires some amount of external world knowledge. Though the usefulness of world knowledge for a coreference system has been known since the late nineties, early mention pair models~\citep{soon2001machine,ng2002improving,yang2003coreference} did not incorporate any form of world knowledge into the system. As knowledge resources became less noisy and exhaustive, some CR researchers started deploying world knowledge to CR. The two main questions to be answered were whether world knowledge offered complementary benefits and whether the noisy nature of world knowledge would effect the performance of the model negatively. Several researchers have deployed world knowledge in the form of web-based encyclopaedias~\citep{uryupina2012disambiguation}, un-annotated data~\citep{daume2005large}, coreference annotated data~\citep{bengtson2008understanding}, and knowledge bases like YAGO, Framenet~\citep{rahman2011coreference} and Wordnet~\citep{durrett2013easy}. World knowledge was mainly incorporated as features into the mention pair models and cluster ranking models. These features were often defined over NPs and verbs. Some initial algorithms like Ng reported an increase in performance up to 4.8\% on inducing world-knowledge features from YAGO and FrameNet on the CR task. While some others~\citep{durrett2013easy} reported only minor performance gains using world-knowledge on system mentions extracted by coreference systems. Some models instead of representing commonsense knowledge as features, also used predicates to encode the commonsense relations~\citep{peng2015solving}. They evaluated their model on hard CR problems that fit the definition of the Winograd Schema Challenge. As posited by Durret and Klein~\citep{durrett2013easy}, the task of modeling complex linguistic constraints into a coreference system remains an uphill battle. 

\section{Conclusion}
Our survey presents an exhaustive overview of the entity resolution field, which forms a core component of natural language processing research. In this survey, we put forth a detailed account of the types of references and the important constraints for entity resolution with the aim of establishing the bread scope of the task. We also clarify the boundaries between the tasks of coreference resolution and anaphora resolution for more focussed research progress in the future. In addition, we also attempt to compare the predominantly used evaluation metrics. We observe that though there are multiple datasets available, the state-of-the-art methods have not been evaluated on them. With the spirit of encouraging more exhaustive evaluations, we also provide an account on the datasets released for the task. 

Entity resolution research has seen a shift from rule-based methods to deep learning methods. To this end, we provide an analysis of the types of algorithms used with special focus on recent deep learning methods. Anaphora resolution is a very important component of the suitcase research problem of sentiment analysis. As the research in the intersection of these two fields is scarce, we also establish a background for the inter-dependency between the two tasks. Finally, we also state the outstanding issues in this task requiring attention, thus laying a firm cornerstone for future researchers to build on. 

 \bibliography{bibtex_review.bib}

\begin{thebibliography}{160}
\providecommand{\natexlab}[1]{#1}
\providecommand{\url}[1]{{#1}}
\providecommand{\urlprefix}{URL }
\expandafter\ifx\csname urlstyle\endcsname\relax
  \providecommand{\doi}[1]{DOI~\discretionary{}{}{}#1}\else
  \providecommand{\doi}{DOI~\discretionary{}{}{}\begingroup
  \urlstyle{rm}\Url}\fi
\providecommand{\eprint}[2][]{\url{#2}}

\bibitem[{Aone and Bennett(1995)}]{aone1995evaluating}
Aone C, Bennett SW (1995) Evaluating automated and manual acquisition of
  anaphora resolution strategies. In: Proceedings of the 33rd annual meeting on
  Association for Computational Linguistics, Association for Computational
  Linguistics, pp 122--129

\bibitem[{Atlas and Levinson(1981)}]{atlas1981clefts}
Atlas JD, Levinson SC (1981) It-clefts, informativeness and logical form:
  Radical pragmatics (revised standard version). In: Radical pragmatics,
  Academic Press, pp 1--62

\bibitem[{Bagga and Baldwin(1998)}]{bagga1998algorithms}
Bagga A, Baldwin B (1998) Algorithms for scoring coreference chains. In: The
  first international conference on language resources and evaluation workshop
  on linguistics coreference, Granada, vol~1, pp 563--566

\bibitem[{Baldwin(1997)}]{baldwin1997cogniac}
Baldwin B (1997) Cogniac: high precision coreference with limited knowledge and
  linguistic resources. In: Proceedings of a Workshop on Operational Factors in
  Practical, Robust Anaphora Resolution for Unrestricted Texts, Association for
  Computational Linguistics, pp 38--45

\bibitem[{Batista-Navarro and Ananiadou(2011)}]{batista2011building}
Batista-Navarro RT, Ananiadou S (2011) Building a coreference-annotated corpus
  from the domain of biochemistry. In: Proceedings of BioNLP 2011 Workshop,
  Association for Computational Linguistics, pp 83--91

\bibitem[{Bawden et~al(2017)Bawden, Sennrich, Birch, and
  Haddow}]{bawden2017evaluating}
Bawden R, Sennrich R, Birch A, Haddow B (2017) Evaluating discourse phenomena
  in neural machine translation. arXiv preprint arXiv:171100513

\bibitem[{Bean and Riloff(2004)}]{bean2004unsupervised}
Bean D, Riloff E (2004) Unsupervised learning of contextual role knowledge for
  coreference resolution. In: Proceedings of the Human Language Technology
  Conference of the North American Chapter of the Association for Computational
  Linguistics: HLT-NAACL 2004

\bibitem[{Bengtson and Roth(2008)}]{bengtson2008understanding}
Bengtson E, Roth D (2008) Understanding the value of features for coreference
  resolution. In: Proceedings of the Conference on Empirical Methods in Natural
  Language Processing, Association for Computational Linguistics, pp 294--303

\bibitem[{Berger et~al(1996)Berger, Pietra, and Pietra}]{berger1996maximum}
Berger AL, Pietra VJD, Pietra SAD (1996) A maximum entropy approach to natural
  language processing. Computational linguistics 22(1):39--71

\bibitem[{Bergler et~al(2003)Bergler, Witte, Khalife, Li, and
  Rudzicz}]{bergler2003using}
Bergler S, Witte R, Khalife M, Li Z, Rudzicz F (2003) Using knowledge-poor
  coreference resolution for text summarization. In: Proceedings of DUC, vol~3

\bibitem[{Bj{\"o}rkelund and Farkas(2012)}]{bjorkelund2012data}
Bj{\"o}rkelund A, Farkas R (2012) Data-driven multilingual coreference
  resolution using resolver stacking. In: Joint Conference on EMNLP and
  CoNLL-Shared Task, Association for Computational Linguistics, pp 49--55

\bibitem[{Brennan et~al(1987)Brennan, Friedman, and
  Pollard}]{brennan1987centering}
Brennan SE, Friedman MW, Pollard CJ (1987) A centering approach to pronouns.
  In: Proceedings of the 25th annual meeting on Association for Computational
  Linguistics, Association for Computational Linguistics, pp 155--162

\bibitem[{Cai and Strube(2010)}]{cai2010evaluation}
Cai J, Strube M (2010) Evaluation metrics for end-to-end coreference resolution
  systems. In: Proceedings of the 11th Annual Meeting of the Special Interest
  Group on Discourse and Dialogue, Association for Computational Linguistics,
  pp 28--36

\bibitem[{Cambria(2016)}]{camacsa}
Cambria E (2016) Affective computing and sentiment analysis. {IEEE} Intelligent
  Systems 31(2):102--107

\bibitem[{Cambria et~al(2017)Cambria, Poria, Gelbukh, and Thelwall}]{camsui}
Cambria E, Poria S, Gelbukh A, Thelwall M (2017) Sentiment analysis is a big
  suitcase. {IEEE} Intelligent Systems 32(6):74--80

\bibitem[{Cambria et~al(2018)Cambria, Poria, Hazarika, and Kwok}]{camnt5}
Cambria E, Poria S, Hazarika D, Kwok K (2018) {SenticNet} 5: Discovering
  conceptual primitives for sentiment analysis by means of context embeddings.
  In: {AAAI}, pp 1795--1802

\bibitem[{Carbonell and Brown(1988)}]{carbonell1988anaphora}
Carbonell JG, Brown RD (1988) Anaphora resolution: a multi-strategy approach.
  In: Proceedings of the 12th conference on Computational linguistics-Volume 1,
  Association for Computational Linguistics, pp 96--101

\bibitem[{Carlson et~al(2003)Carlson, Marcu, and
  Okurowski}]{carlson2003building}
Carlson L, Marcu D, Okurowski ME (2003) Building a discourse-tagged corpus in
  the framework of rhetorical structure theory. In: Current and new directions
  in discourse and dialogue, Springer, pp 85--112

\bibitem[{Castagnola(2002)}]{castagnola2002anaphora}
Castagnola L (2002) Anaphora resolution for question answering. PhD thesis,
  Massachusetts Institute of Technology

\bibitem[{Chang et~al(2013)Chang, Samdani, and Roth}]{chang2013constrained}
Chang KW, Samdani R, Roth D (2013) A constrained latent variable model for
  coreference resolution. In: Proceedings of the 2013 Conference on Empirical
  Methods in Natural Language Processing, pp 601--612

\bibitem[{Chaturvedi et~al(2018)Chaturvedi, Cambria, Welsch, and
  Herrera}]{chadis}
Chaturvedi I, Cambria E, Welsch R, Herrera F (2018) Distinguishing between
  facts and opinions for sentiment analysis: Survey and challenges. Information
  Fusion 44:65--77

\bibitem[{Chen and Manning(2014)}]{chen2014fast}
Chen D, Manning C (2014) A fast and accurate dependency parser using neural
  networks. In: Proceedings of the 2014 conference on empirical methods in
  natural language processing (EMNLP), pp 740--750

\bibitem[{Chen and Choi(2016)}]{chen2016character}
Chen YH, Choi JD (2016) Character identification on multiparty conversation:
  Identifying mentions of characters in tv shows. In: Proceedings of the 17th
  Annual Meeting of the Special Interest Group on Discourse and Dialogue, pp
  90--100

\bibitem[{Chinchor(1998)}]{chinchor1998overview}
Chinchor NA (1998) Overview of muc-7/met-2. Tech. rep., SCIENCE APPLICATIONS
  INTERNATIONAL CORP SAN DIEGO CA

\bibitem[{Clark and Manning(2015)}]{clark2015entity}
Clark K, Manning CD (2015) Entity-centric coreference resolution with model
  stacking. In: Proceedings of the 53rd Annual Meeting of the Association for
  Computational Linguistics and the 7th International Joint Conference on
  Natural Language Processing (Volume 1: Long Papers), vol~1, pp 1405--1415

\bibitem[{Clark and Manning(2016{\natexlab{a}})}]{clark2016deep}
Clark K, Manning CD (2016{\natexlab{a}}) Deep reinforcement learning for
  mention-ranking coreference models. arXiv preprint arXiv:160908667

\bibitem[{Clark and Manning(2016{\natexlab{b}})}]{clark2016improving}
Clark K, Manning CD (2016{\natexlab{b}}) Improving coreference resolution by
  learning entity-level distributed representations. arXiv preprint
  arXiv:160601323

\bibitem[{Cohen et~al(2010)Cohen, Johnson, Verspoor, Roeder, and
  Hunter}]{cohen2010structural}
Cohen KB, Johnson HL, Verspoor K, Roeder C, Hunter LE (2010) The structural and
  content aspects of abstracts versus bodies of full text journal articles are
  different. BMC bioinformatics 11(1):492

\bibitem[{Cohen and Singer(1999)}]{cohen1999simple}
Cohen WW, Singer Y (1999) A simple, fast, and effective rule learner. AAAI/IAAI
  99:335--342

\bibitem[{Collobert and Weston(2008)}]{collobert2008unified}
Collobert R, Weston J (2008) A unified architecture for natural language
  processing: Deep neural networks with multitask learning. In: Proceedings of
  the 25th international conference on Machine learning, ACM, pp 160--167

\bibitem[{Culotta et~al(2007)Culotta, Wick, and McCallum}]{culotta2007first}
Culotta A, Wick M, McCallum A (2007) First-order probabilistic models for
  coreference resolution. In: Human Language Technologies 2007: The Conference
  of the North American Chapter of the Association for Computational
  Linguistics; Proceedings of the Main Conference, pp 81--88

\bibitem[{Cybulska and Vossen(2014)}]{cybulska2014guidelines}
Cybulska A, Vossen P (2014) Guidelines for ecb+ annotation of events and their
  coreference. Tech. rep., Technical Report NWR-2014-1, VU University Amsterdam

\bibitem[{Daelemans et~al(2004)Daelemans, Zavrel, Van Der~Sloot, and Van~den
  Bosch}]{daelemans2004timbl}
Daelemans W, Zavrel J, Van Der~Sloot K, Van~den Bosch A (2004) Timbl: Tilburg
  memory-based learner. Tilburg University

\bibitem[{Daum{\'e}~III and Marcu(2005)}]{daume2005large}
Daum{\'e}~III H, Marcu D (2005) A large-scale exploration of effective global
  features for a joint entity detection and tracking model. In: Proceedings of
  the conference on Human Language Technology and Empirical Methods in Natural
  Language Processing, Association for Computational Linguistics, pp 97--104

\bibitem[{Denis and Baldridge(2007)}]{denis2007joint}
Denis P, Baldridge J (2007) Joint determination of anaphoricity and coreference
  resolution using integer programming. In: Human Language Technologies 2007:
  The Conference of the North American Chapter of the Association for
  Computational Linguistics; Proceedings of the Main Conference, pp 236--243

\bibitem[{Denis and Baldridge(2008)}]{denis2008specialized}
Denis P, Baldridge J (2008) Specialized models and ranking for coreference
  resolution. In: Proceedings of the Conference on Empirical Methods in Natural
  Language Processing, Association for Computational Linguistics, pp 660--669

\bibitem[{Denis and Baldridge(2009)}]{denis2009global}
Denis P, Baldridge J (2009) Global joint models for coreference resolution and
  named entity classification. Procesamiento del Lenguaje Natural 42

\bibitem[{Ding and Liu(2010)}]{ding2010resolving}
Ding X, Liu B (2010) Resolving object and attribute coreference in opinion
  mining. In: Proceedings of the 23rd International Conference on Computational
  Linguistics, Association for Computational Linguistics, pp 268--276

\bibitem[{Dixon(2003)}]{dixon2003demonstratives}
Dixon RM (2003) Demonstratives: A cross-linguistic typology. Studies in
  Language International Journal sponsored by the Foundation ``Foundations of
  Language'' 27(1):61--112

\bibitem[{Doddington et~al(2004)Doddington, Mitchell, Przybocki, Ramshaw,
  Strassel, and Weischedel}]{doddington2004automatic}
Doddington GR, Mitchell A, Przybocki MA, Ramshaw LA, Strassel S, Weischedel RM
  (2004) The automatic content extraction (ace) program-tasks, data, and
  evaluation. In: LREC, vol~2, p~1

\bibitem[{Durrett and Klein(2013)}]{durrett2013easy}
Durrett G, Klein D (2013) Easy victories and uphill battles in coreference
  resolution. In: Proceedings of the 2013 Conference on Empirical Methods in
  Natural Language Processing, pp 1971--1982

\bibitem[{Durrett and Klein(2014)}]{durrett2014joint}
Durrett G, Klein D (2014) A joint model for entity analysis: Coreference,
  typing, and linking. Transactions of the Association for Computational
  Linguistics 2:477--490

\bibitem[{Fernandes et~al(2012)Fernandes, Dos~Santos, and
  Milidi{\'u}}]{fernandes2012latent}
Fernandes ER, Dos~Santos CN, Milidi{\'u} RL (2012) Latent structure perceptron
  with feature induction for unrestricted coreference resolution. In: Joint
  Conference on EMNLP and CoNLL-Shared Task, Association for Computational
  Linguistics, pp 41--48

\bibitem[{Fillmore(1986)}]{fillmore1986pragmatically}
Fillmore CJ (1986) Pragmatically controlled zero anaphora. In: Annual Meeting
  of the Berkeley Linguistics Society, vol~12, pp 95--107

\bibitem[{Finkel and Manning(2008)}]{finkel2008enforcing}
Finkel JR, Manning CD (2008) Enforcing transitivity in coreference resolution.
  In: Proceedings of the 46th Annual Meeting of the Association for
  Computational Linguistics on Human Language Technologies: Short Papers,
  Association for Computational Linguistics, pp 45--48

\bibitem[{Gasperin and Briscoe(2008)}]{gasperin2008statistical}
Gasperin C, Briscoe T (2008) Statistical anaphora resolution in biomedical
  texts. In: Proceedings of the 22nd International Conference on Computational
  Linguistics-Volume 1, Association for Computational Linguistics, pp 257--264

\bibitem[{Ge et~al(1998)Ge, Hale, and Charniak}]{ge1998statistical}
Ge N, Hale J, Charniak E (1998) A statistical approach to anaphora resolution.
  In: Sixth Workshop on Very Large Corpora

\bibitem[{Ghaddar and Langlais(2016)}]{ghaddar2016wikicoref}
Ghaddar A, Langlais P (2016) Wikicoref: An english coreference-annotated corpus
  of wikipedia articles. In: LREC

\bibitem[{Goldberg and Michaelis(2017)}]{goldberg2017one}
Goldberg AE, Michaelis LA (2017) One among many: Anaphoric one and its
  relationship with numeral one. Cognitive science 41(S2):233--258

\bibitem[{Grishman and Sundheim(1996)}]{grishman1996message}
Grishman R, Sundheim B (1996) Message understanding conference-6: A brief
  history. In: COLING 1996 Volume 1: The 16th International Conference on
  Computational Linguistics, vol~1

\bibitem[{Gross et~al(1993)Gross, Allen, and Traum}]{gross1993trains}
Gross D, Allen J, Traum D (1993) The trains 91 dialogues (trains tech. note
  92--1). Rochester, NY: University of Rochester, Department of Computer
  Science

\bibitem[{Grosz et~al(1995)Grosz, Weinstein, and Joshi}]{grosz1995centering}
Grosz BJ, Weinstein S, Joshi AK (1995) Centering: A framework for modeling the
  local coherence of discourse. Computational linguistics 21(2):203--225

\bibitem[{Guillou et~al(2014)Guillou, Hardmeier, Smith, Tiedemann, and
  Webber}]{guillou2014parcor}
Guillou L, Hardmeier C, Smith A, Tiedemann J, Webber B (2014) Parcor 1.0: A
  parallel pronoun-coreference corpus to support statistical mt. In: 9th
  International Conference on Language Resources and Evaluation (LREC), MAY
  26-31, 2014, Reykjavik, ICELAND, European Language Resources Association, pp
  3191--3198

\bibitem[{Gundel et~al(2005)Gundel, Hedberg, and
  Zacharski}]{gundel2005pronouns}
Gundel J, Hedberg N, Zacharski R (2005) Pronouns without np antecedents: How do
  we know when a pronoun is referential. Anaphora processing: linguistic,
  cognitive and computational modelling pp 351--364

\bibitem[{Haghighi and Klein(2007)}]{haghighi2007unsupervised}
Haghighi A, Klein D (2007) Unsupervised coreference resolution in a
  nonparametric bayesian model. In: Proceedings of the 45th annual meeting of
  the association of computational linguistics, pp 848--855

\bibitem[{Haghighi and Klein(2009)}]{haghighi2009simple}
Haghighi A, Klein D (2009) Simple coreference resolution with rich syntactic
  and semantic features. In: Proceedings of the 2009 Conference on Empirical
  Methods in Natural Language Processing: Volume 3-Volume 3, Association for
  Computational Linguistics, pp 1152--1161

\bibitem[{Haghighi and Klein(2010)}]{haghighi2010coreference}
Haghighi A, Klein D (2010) Coreference resolution in a modular, entity-centered
  model. In: Human Language Technologies: The 2010 Annual Conference of the
  North American Chapter of the Association for Computational Linguistics,
  Association for Computational Linguistics, pp 385--393

\bibitem[{Harabagiu and Maiorano(1999)}]{harabagiu1999knowledge}
Harabagiu SM, Maiorano SJ (1999) Knowledge-lean coreference resolution and its
  relation to textual cohesion and coherence. The Relation of
  Discourse/Dialogue Structure and Reference

\bibitem[{Harabagiu et~al(2001)Harabagiu, Bunescu, and
  Maiorano}]{harabagiu2001text}
Harabagiu SM, Bunescu RC, Maiorano SJ (2001) Text and knowledge mining for
  coreference resolution. In: Proceedings of the second meeting of the North
  American Chapter of the Association for Computational Linguistics on Language
  technologies, Association for Computational Linguistics, pp 1--8

\bibitem[{Hardmeier and Federico(2010)}]{hardmeier2010modelling}
Hardmeier C, Federico M (2010) Modelling pronominal anaphora in statistical
  machine translation. In: IWSLT (International Workshop on Spoken Language
  Translation); Paris, France; December 2nd and 3rd, 2010., pp 283--289

\bibitem[{Hasler and Orasan(2009)}]{hasler2009coreferential}
Hasler L, Orasan C (2009) Do coreferential arguments make event mentions
  coreferential. In: Proc. the 7th Discourse Anaphora and Anaphor Resolution
  Colloquium (DAARC 2009), Citeseer

\bibitem[{Heeman and Allen(1995)}]{heeman1995trains}
Heeman PA, Allen JF (1995) The trains 93 dialogues. Tech. rep., ROCHESTER UNIV
  NY DEPT OF COMPUTER SCIENCE

\bibitem[{Heim(1982)}]{heim1982semantics}
Heim I (1982) The semantics of definite and indefinite nps. University of
  Massachusetts at Amherst dissertation

\bibitem[{Hobbs(1978)}]{hobbs1978resolving}
Hobbs JR (1978) Resolving pronoun references. Lingua 44(4):311--338

\bibitem[{Hou et~al(2013)Hou, Markert, and Strube}]{hou2013global}
Hou Y, Markert K, Strube M (2013) Global inference for bridging anaphora
  resolution. In: Proceedings of the 2013 Conference of the North American
  Chapter of the Association for Computational Linguistics: Human Language
  Technologies, pp 907--917

\bibitem[{Jakob and Gurevych(2010)}]{jakob2010using}
Jakob N, Gurevych I (2010) Using anaphora resolution to improve opinion target
  identification in movie reviews. In: Proceedings of the ACL 2010 Conference
  Short Papers, Association for Computational Linguistics, pp 263--268

\bibitem[{Kennedy and Boguraev(1996)}]{kennedy1996anaphora}
Kennedy C, Boguraev B (1996) Anaphora for everyone: pronominal anaphora
  resoluation without a parser. In: Proceedings of the 16th conference on
  Computational linguistics-Volume 1, Association for Computational
  Linguistics, pp 113--118

\bibitem[{Kibble(2001)}]{kibble2001reformulation}
Kibble R (2001) A reformulation of rule 2 of centering theory. Computational
  Linguistics 27(4):579--587

\bibitem[{Kim et~al(2003)Kim, Ohta, Tateisi, and Tsujii}]{kim2003genia}
Kim JD, Ohta T, Tateisi Y, Tsujii J (2003) Genia corpus---a semantically
  annotated corpus for bio-textmining. Bioinformatics 19(suppl\_1):i180--i182

\bibitem[{Kim et~al(2008)Kim, Ohta, and Tsujii}]{kim2008corpus}
Kim JD, Ohta T, Tsujii J (2008) Corpus annotation for mining biomedical events
  from literature. BMC bioinformatics 9(1):10

\bibitem[{Kim et~al(2011)Kim, Pyysalo, Ohta, Bossy, Nguyen, and
  Tsujii}]{kim2011overview}
Kim JD, Pyysalo S, Ohta T, Bossy R, Nguyen N, Tsujii J (2011) Overview of
  bionlp shared task 2011. In: Proceedings of the BioNLP shared task 2011
  workshop, Association for Computational Linguistics, pp 1--6

\bibitem[{Lappin and Leass(1994)}]{lappin1994algorithm}
Lappin S, Leass HJ (1994) An algorithm for pronominal anaphora resolution.
  Computational linguistics 20(4):535--561

\bibitem[{Le et~al(2016)Le, Vo, Mai, Quan, and Phan}]{le2016sentiment}
Le TT, Vo TH, Mai DT, Quan TT, Phan TT (2016) Sentiment analysis using
  anaphoric coreference resolution and ontology inference. In: International
  Workshop on Multi-disciplinary Trends in Artificial Intelligence, Springer,
  pp 297--303

\bibitem[{Lee et~al(2011)Lee, Peirsman, Chang, Chambers, Surdeanu, and
  Jurafsky}]{lee2011stanford}
Lee H, Peirsman Y, Chang A, Chambers N, Surdeanu M, Jurafsky D (2011)
  Stanford's multi-pass sieve coreference resolution system at the conll-2011
  shared task. In: Proceedings of the fifteenth conference on computational
  natural language learning: Shared task, Association for Computational
  Linguistics, pp 28--34

\bibitem[{Lee et~al(2013)Lee, Chang, Peirsman, Chambers, Surdeanu, and
  Jurafsky}]{lee2013deterministic}
Lee H, Chang A, Peirsman Y, Chambers N, Surdeanu M, Jurafsky D (2013)
  Deterministic coreference resolution based on entity-centric,
  precision-ranked rules. Computational Linguistics 39(4):885--916

\bibitem[{Lee et~al(2017{\natexlab{a}})Lee, Surdeanu, and
  Jurafsky}]{lee2017scaffolding}
Lee H, Surdeanu M, Jurafsky D (2017{\natexlab{a}}) A scaffolding approach to
  coreference resolution integrating statistical and rule-based models. Natural
  Language Engineering 23(5):733--762

\bibitem[{Lee et~al(2017{\natexlab{b}})Lee, He, Lewis, and
  Zettlemoyer}]{lee2017end}
Lee K, He L, Lewis M, Zettlemoyer L (2017{\natexlab{b}}) End-to-end neural
  coreference resolution. arXiv preprint arXiv:170707045

\bibitem[{Levesque et~al(2011)Levesque, Davis, and
  Morgenstern}]{levesque2011winograd}
Levesque HJ, Davis E, Morgenstern L (2011) The winograd schema challenge. In:
  Aaai spring symposium: Logical formalizations of commonsense reasoning,
  vol~46, p~47

\bibitem[{Liang and Wu(2004)}]{liang2004automatic}
Liang T, Wu DS (2004) Automatic pronominal anaphora resolution in english
  texts. International Journal of Computational Linguistics \& Chinese Language
  Processing, Volume 9, Number 1, February 2004: Special Issue on Selected
  Papers from ROCLING XV 9(1):21--40

\bibitem[{Luo(2005)}]{luo2005coreference}
Luo X (2005) On coreference resolution performance metrics. In: Proceedings of
  the conference on human language technology and empirical methods in natural
  language processing, Association for Computational Linguistics, pp 25--32

\bibitem[{Ma et~al(2014)Ma, Doppa, Orr, Mannem, Fern, Dietterich, and
  Tadepalli}]{ma2014prune}
Ma C, Doppa JR, Orr JW, Mannem P, Fern X, Dietterich T, Tadepalli P (2014)
  Prune-and-score: Learning for greedy coreference resolution. In: Proceedings
  of the 2014 Conference on Empirical Methods in Natural Language Processing
  (EMNLP), pp 2115--2126

\bibitem[{Ma et~al(2018)Ma, Peng, and Cambria}]{maatar}
Ma Y, Peng H, Cambria E (2018) Targeted aspect-based sentiment analysis via
  embedding commonsense knowledge into an attentive {LSTM}. In: {AAAI}, pp
  5876--5883

\bibitem[{Marasovi{\'c} et~al(2017)Marasovi{\'c}, Born, Opitz, and
  Frank}]{marasovic2017mention}
Marasovi{\'c} A, Born L, Opitz J, Frank A (2017) A mention-ranking model for
  abstract anaphora resolution. arXiv preprint arXiv:170602256

\bibitem[{McCallum and Wellner(2003)}]{mccallum2003object}
McCallum A, Wellner B (2003) Object consolidation by graph partitioning with a
  conditionally-trained distance metric. In: KDD Workshop on Data Cleaning,
  Record Linkage and Object Consolidation, Citeseer

\bibitem[{McCallum and Wellner(2005)}]{mccallum2005conditional}
McCallum A, Wellner B (2005) Conditional models of identity uncertainty with
  application to noun coreference. In: Advances in neural information
  processing systems, pp 905--912

\bibitem[{McCarthy and Lehnert(1995)}]{mccarthy1995using}
McCarthy JF, Lehnert WG (1995) Using decision trees for coreference resolution.
  arXiv preprint cmp-lg/9505043

\bibitem[{Mikolov et~al(2013)Mikolov, Sutskever, Chen, Corrado, and
  Dean}]{mikolov2013distributed}
Mikolov T, Sutskever I, Chen K, Corrado GS, Dean J (2013) Distributed
  representations of words and phrases and their compositionality. In: Advances
  in neural information processing systems, pp 3111--3119

\bibitem[{Mitkov(1998)}]{mitkov1998robust}
Mitkov R (1998) Robust pronoun resolution with limited knowledge. In:
  Proceedings of the 36th Annual Meeting of the Association for Computational
  Linguistics and 17th International Conference on Computational
  Linguistics-Volume 2, Association for Computational Linguistics, pp 869--875

\bibitem[{Mitkov(1999)}]{mitkov1999anaphora}
Mitkov R (1999) Anaphora resolution: the state of the art. Citeseer

\bibitem[{Mitkov(2001{\natexlab{a}})}]{mitkov2001outstanding}
Mitkov R (2001{\natexlab{a}}) Outstanding issues in anaphora resolution. In:
  International Conference on Intelligent Text Processing and Computational
  Linguistics, Springer, pp 110--125

\bibitem[{Mitkov(2001{\natexlab{b}})}]{mitkov2001towards}
Mitkov R (2001{\natexlab{b}}) Towards a more consistent and comprehensive
  evaluation of anaphora resolution algorithms and systems. Applied Artificial
  Intelligence 15(3):253--276

\bibitem[{Mitkov(2014)}]{mitkov2014anaphora}
Mitkov R (2014) Anaphora resolution. Routledge

\bibitem[{Mitkov et~al(2002)Mitkov, Evans, and Orasan}]{mitkov2002new}
Mitkov R, Evans R, Orasan C (2002) A new, fully automatic version of mitkov's
  knowledge-poor pronoun resolution method. In: International Conference on
  Intelligent Text Processing and Computational Linguistics, Springer, pp
  168--186

\bibitem[{Moosavi and Strube(2016)}]{moosavi2016coreference}
Moosavi NS, Strube M (2016) Which coreference evaluation metric do you trust? a
  proposal for a link-based entity aware metric. In: Proceedings of the 54th
  Annual Meeting of the Association for Computational Linguistics (Volume 1:
  Long Papers), vol~1, pp 632--642

\bibitem[{Moosavi and Strube(2017)}]{moosavi2017lexical}
Moosavi NS, Strube M (2017) Lexical features in coreference resolution: To be
  used with caution. arXiv preprint arXiv:170406779

\bibitem[{Ng(2005)}]{ng2005machine}
Ng V (2005) Machine learning for coreference resolution: From local
  classification to global ranking. In: Proceedings of the 43rd Annual Meeting
  on Association for Computational Linguistics, Association for Computational
  Linguistics, pp 157--164

\bibitem[{Ng(2010)}]{ng2010supervised}
Ng V (2010) Supervised noun phrase coreference research: The first fifteen
  years. In: Proceedings of the 48th annual meeting of the association for
  computational linguistics, Association for Computational Linguistics, pp
  1396--1411

\bibitem[{Ng and Cardie(2002{\natexlab{a}})}]{ng2002combining}
Ng V, Cardie C (2002{\natexlab{a}}) Combining sample selection and error-driven
  pruning for machine learning of coreference rules. In: Proceedings of the
  ACL-02 conference on Empirical methods in natural language processing-Volume
  10, Association for Computational Linguistics, pp 55--62

\bibitem[{Ng and Cardie(2002{\natexlab{b}})}]{ng2002improving}
Ng V, Cardie C (2002{\natexlab{b}}) Improving machine learning approaches to
  coreference resolution. In: Proceedings of the 40th annual meeting on
  association for computational linguistics, Association for Computational
  Linguistics, pp 104--111

\bibitem[{Nicolae and Nicolae(2006)}]{nicolae2006bestcut}
Nicolae C, Nicolae G (2006) Bestcut: A graph algorithm for coreference
  resolution. In: Proceedings of the 2006 conference on empirical methods in
  natural language processing, Association for Computational Linguistics, pp
  275--283

\bibitem[{Nicolov et~al(2008)Nicolov, Salvetti, and
  Ivanova}]{nicolov2008sentiment}
Nicolov N, Salvetti F, Ivanova S (2008) Sentiment analysis: Does coreference
  matter. In: AISB 2008 Convention Communication, Interaction and Social
  Intelligence, vol~1, p~37

\bibitem[{O'Connor and Heilman(2013)}]{o2013arkerf}
O'Connor B, Heilman M (2013) Arkref: A rule-based coreference resolution
  system. arXiv preprint arXiv:13101975

\bibitem[{Oneto et~al(2016)Oneto, Bisio, Cambria, and Anguita}]{onesta}
Oneto L, Bisio F, Cambria E, Anguita D (2016) Statistical learning theory and
  {ELM} for big social data analysis. {IEEE} Computational Intelligence
  Magazine 11(3):45--55

\bibitem[{Peng et~al(2015)Peng, Khashabi, and Roth}]{peng2015solving}
Peng H, Khashabi D, Roth D (2015) Solving hard coreference problems. In:
  Proceedings of the 2015 Conference of the North American Chapter of the
  Association for Computational Linguistics: Human Language Technologies, pp
  809--819

\bibitem[{Pennington et~al(2014)Pennington, Socher, and
  Manning}]{pennington2014glove}
Pennington J, Socher R, Manning C (2014) Glove: Global vectors for word
  representation. In: Proceedings of the 2014 conference on empirical methods
  in natural language processing (EMNLP), pp 1532--1543

\bibitem[{Poesio(2004)}]{poesio2004discourse}
Poesio M (2004) Discourse annotation and semantic annotation in the gnome
  corpus. In: Proceedings of the 2004 ACL Workshop on Discourse Annotation,
  Association for Computational Linguistics, pp 72--79

\bibitem[{Poesio and Kabadjov(2004)}]{poesio2004general}
Poesio M, Kabadjov MA (2004) A general-purpose, off-the-shelf anaphora
  resolution module: Implementation and preliminary evaluation. In: LREC

\bibitem[{Poesio et~al(2008)Poesio, Artstein et~al}]{poesio2008anaphoric}
Poesio M, Artstein R, et~al (2008) Anaphoric annotation in the arrau corpus.
  In: LREC

\bibitem[{Poon and Domingos(2008)}]{poon2008joint}
Poon H, Domingos P (2008) Joint unsupervised coreference resolution with markov
  logic. In: Proceedings of the conference on empirical methods in natural
  language processing, Association for Computational Linguistics, pp 650--659

\bibitem[{Pradhan et~al(2011)Pradhan, Ramshaw, Marcus, Palmer, Weischedel, and
  Xue}]{pradhan2011conll}
Pradhan S, Ramshaw L, Marcus M, Palmer M, Weischedel R, Xue N (2011) Conll-2011
  shared task: Modeling unrestricted coreference in ontonotes. In: Proceedings
  of the Fifteenth Conference on Computational Natural Language Learning:
  Shared Task, Association for Computational Linguistics, pp 1--27

\bibitem[{Pradhan et~al(2012)Pradhan, Moschitti, Xue, Uryupina, and
  Zhang}]{pradhan2012conll}
Pradhan S, Moschitti A, Xue N, Uryupina O, Zhang Y (2012) Conll-2012 shared
  task: Modeling multilingual unrestricted coreference in ontonotes. In: Joint
  Conference on EMNLP and CoNLL-Shared Task, Association for Computational
  Linguistics, pp 1--40

\bibitem[{Pradhan et~al(2014)Pradhan, Luo, Recasens, Hovy, Ng, and
  Strube}]{pradhan2014scoring}
Pradhan S, Luo X, Recasens M, Hovy E, Ng V, Strube M (2014) Scoring coreference
  partitions of predicted mentions: A reference implementation. In: Proceedings
  of the conference. Association for Computational Linguistics. Meeting, NIH
  Public Access, vol 2014, p~30

\bibitem[{Preuss(1992)}]{preuss1992anaphora}
Preuss S (1992) Anaphora resolution in machine translation. TU, Fachbereich 20,
  Projektgruppe KIT

\bibitem[{Pustejovsky et~al(2002)Pustejovsky, Castano, Sauri, Rumshinsky,
  Zhang, and Luo}]{pustejovsky2002medstract}
Pustejovsky J, Castano J, Sauri R, Rumshinsky A, Zhang J, Luo W (2002)
  Medstract: creating large-scale information servers for biomedical libraries.
  In: Proceedings of the ACL-02 workshop on Natural language processing in the
  biomedical domain-Volume 3, Association for Computational Linguistics, pp
  85--92

\bibitem[{Quinlan(1986)}]{quinlan1986induction}
Quinlan JR (1986) Induction of decision trees. Machine learning 1(1):81--106

\bibitem[{Raghunathan et~al(2010)Raghunathan, Lee, Rangarajan, Chambers,
  Surdeanu, Jurafsky, and Manning}]{raghunathan2010multi}
Raghunathan K, Lee H, Rangarajan S, Chambers N, Surdeanu M, Jurafsky D, Manning
  C (2010) A multi-pass sieve for coreference resolution. In: Proceedings of
  2010 Conference on Empirical Methods in NAturalLanguage Processing,
  Association for Computational Linguistics, pp 492--501

\bibitem[{Rahman and Ng(2009)}]{rahman2009supervised}
Rahman A, Ng V (2009) Supervised models for coreference resolution. In:
  Proceedings of the 2009 Conference on Empirical Methods in Natural Language
  Processing: Volume 2-Volume 2, Association for Computational Linguistics, pp
  968--977

\bibitem[{Rahman and Ng(2011)}]{rahman2011coreference}
Rahman A, Ng V (2011) Coreference resolution with world knowledge. In:
  Proceedings of the 49th Annual Meeting of the Association for Computational
  Linguistics: Human Language Technologies-Volume 1, Association for
  Computational Linguistics, pp 814--824

\bibitem[{Rand(1971)}]{rand1971objective}
Rand WM (1971) Objective criteria for the evaluation of clustering methods.
  Journal of the American Statistical association 66(336):846--850

\bibitem[{Recasens and Hovy(2011)}]{recasens2011blanc}
Recasens M, Hovy E (2011) Blanc: Implementing the rand index for coreference
  evaluation. Natural Language Engineering 17(4):485--510

\bibitem[{Recasens and Mart{\'\i}(2010)}]{recasens2010ancora}
Recasens M, Mart{\'\i} MA (2010) Ancora-co: Coreferentially annotated corpora
  for spanish and catalan. Language resources and evaluation 44(4):315--345

\bibitem[{Recasens and Vila(2010)}]{recasens2010paraphrase}
Recasens M, Vila M (2010) On paraphrase and coreference. Computational
  Linguistics 36(4):639--647

\bibitem[{Recasens et~al(2010)Recasens, M{\`a}rquez, Sapena, Mart{\'\i},
  Taul{\'e}, Hoste, Poesio, and Versley}]{recasens2010semeval}
Recasens M, M{\`a}rquez L, Sapena E, Mart{\'\i} MA, Taul{\'e} M, Hoste V,
  Poesio M, Versley Y (2010) Semeval-2010 task 1: Coreference resolution in
  multiple languages. In: Proceedings of the 5th International Workshop on
  Semantic Evaluation, Association for Computational Linguistics, pp 1--8

\bibitem[{Recasens et~al(2013)Recasens, de~Marneffe, and
  Potts}]{recasens2013life}
Recasens M, de~Marneffe MC, Potts C (2013) The life and death of discourse
  entities: Identifying singleton mentions. In: Proceedings of the 2013
  Conference of the North American Chapter of the Association of Computational
  Linguistics:Human Language Technologies, pp 627--633

\bibitem[{Regneri and Wang(2012)}]{regneri2012using}
Regneri M, Wang R (2012) Using discourse information for paraphrase extraction.
  In: Proceedings of the 2012 Joint Conference on Empirical Methods in Natural
  Language Processing and Computational Natural Language Learning, Association
  for Computational Linguistics, pp 916--927

\bibitem[{Roberts(1989)}]{roberts1989modal}
Roberts C (1989) Modal subordination and pronominal anaphora in discourse.
  Linguistics and philosophy 12(6):683--721

\bibitem[{Van~der Sandt(1992)}]{van1992presupposition}
Van~der Sandt RA (1992) Presupposition projection as anaphora resolution.
  Journal of semantics 9(4):333--377

\bibitem[{Segura-Bedmar et~al(2009)Segura-Bedmar, Crespo, de~Pablo, and
  Mart{\'\i}nez}]{segura2009drugnerar}
Segura-Bedmar I, Crespo M, de~Pablo C, Mart{\'\i}nez P (2009) Drugnerar:
  linguistic rule-based anaphora resolver for drug-drug interaction extraction
  in pharmacological documents. In: Proceedings of the third international
  workshop on Data and text mining in bioinformatics, ACM, pp 19--26

\bibitem[{Soon et~al(2001)Soon, Ng, and Lim}]{soon2001machine}
Soon WM, Ng HT, Lim DCY (2001) A machine learning approach to coreference
  resolution of noun phrases. Computational linguistics 27(4):521--544

\bibitem[{Soricut and Marcu(2003)}]{soricut2003sentence}
Soricut R, Marcu D (2003) Sentence level discourse parsing using syntactic and
  lexical information. In: Proceedings of the 2003 Conference of the North
  American Chapter of the Association for Computational Linguistics on Human
  Language Technology-Volume 1, Association for Computational Linguistics, pp
  149--156

\bibitem[{Steinberger et~al(2007)Steinberger, Poesio, Kabadjov, and
  Je{\v{z}}ek}]{steinberger2007two}
Steinberger J, Poesio M, Kabadjov MA, Je{\v{z}}ek K (2007) Two uses of anaphora
  resolution in summarization. Information Processing \& Management
  43(6):1663--1680

\bibitem[{Stoyanov et~al(2010)Stoyanov, Cardie, Gilbert, Riloff, Buttler, and
  Hysom}]{stoyanov2010coreference}
Stoyanov V, Cardie C, Gilbert N, Riloff E, Buttler D, Hysom D (2010)
  Coreference resolution with reconcile. In: Proceedings of the ACL 2010
  Conference Short Papers, Association for Computational Linguistics, pp
  156--161

\bibitem[{Strube et~al(2002)Strube, Rapp, and M{\"u}ller}]{strube2002influence}
Strube M, Rapp S, M{\"u}ller C (2002) The influence of minimum edit distance on
  reference resolution. In: Proceedings of the ACL-02 conference on Empirical
  methods in natural language processing-Volume 10, Association for
  Computational Linguistics, pp 312--319

\bibitem[{Su et~al(2008)Su, Yang, Hong, Tateisi, and
  Tsujii}]{su2008coreference}
Su J, Yang X, Hong H, Tateisi Y, Tsujii J (2008) Coreference resolution in
  biomedical texts: a machine learning approach. In: Dagstuhl Seminar
  Proceedings, Schloss Dagstuhl-Leibniz-Zentrum f{\"u}r Informatik

\bibitem[{Tateisi et~al(2005)Tateisi, Yakushiji, Ohta, and
  Tsujii}]{tateisi2005syntax}
Tateisi Y, Yakushiji A, Ohta T, Tsujii J (2005) Syntax annotation for the genia
  corpus. In: Companion Volume to the Proceedings of Conference including
  Posters/Demos and tutorial abstracts

\bibitem[{Tetreault(1999)}]{tetreault1999analysis}
Tetreault JR (1999) Analysis of syntax-based pronoun resolution methods. In:
  Proceedings of the 37th annual meeting of the Association for Computational
  Linguistics on Computational Linguistics, Association for Computational
  Linguistics, pp 602--605

\bibitem[{Tetreault(2001)}]{tetreault2001corpus}
Tetreault JR (2001) A corpus-based evaluation of centering and pronoun
  resolution. Computational Linguistics 27(4):507--520

\bibitem[{Uryupina et~al(2012)Uryupina, Poesio, Giuliano, and
  Tymoshenko}]{uryupina2012disambiguation}
Uryupina O, Poesio M, Giuliano C, Tymoshenko K (2012) Disambiguation and
  filtering methods in using web knowledge for coreference resolution. In:
  Cross-Disciplinary Advances in Applied Natural Language Processing: Issues
  and Approaches, IGI Global, pp 185--201

\bibitem[{Vala et~al(2016)Vala, Piper, and Ruths}]{vala2016more}
Vala H, Piper A, Ruths D (2016) The more antecedents, the merrier: Resolving
  multi-antecedent anaphors. In: Proceedings of the 54th Annual Meeting of the
  Association for Computational Linguistics (Volume 1: Long Papers), vol~1, pp
  2287--2296

\bibitem[{Valdivia et~al(2018)Valdivia, Luz{\'o}n, Cambria, and
  Herrera}]{valcon}
Valdivia A, Luz{\'o}n V, Cambria E, Herrera F (2018) Consensus vote models for
  detecting and filtering neutrality in sentiment analysis. Information Fusion
  44:126--135

\bibitem[{Versley et~al(2008)Versley, Pozetto, Poesio, Eidlman, Jern, Smith,
  Yang, and Moschitti}]{versley2008bart}
Versley Y, Pozetto SP, Poesio M, Eidlman V, Jern A, Smith J, Yang X, Moschitti
  A (2008) Bart: A modular toolkit for coreference resolution. In: Proceedings
  of the 46th Annual MEeting of Association for Computational Linguistics on
  Human Language Technologies: Demo Session, Association for Computational
  Linguistics, pp 9--12

\bibitem[{Vicedo and Ferr{\'a}ndez(2000)}]{vicedo2000importance}
Vicedo JL, Ferr{\'a}ndez A (2000) Importance of pronominal anaphora resolution
  in question answering systems. In: Proceedings of the 38th Annual Meeting on
  Association for Computational Linguistics, Association for Computational
  Linguistics, pp 555--562

\bibitem[{Vieira et~al(2005)Vieira, Salmon-Alt, Gasperin, Schang, and
  Othero}]{vieira2005coreference}
Vieira R, Salmon-Alt S, Gasperin C, Schang E, Othero G (2005) Coreference and
  anaphoric relations of demonstrative noun phrases in multilingual corpus.
  Anaphora Processing: linguistic, cognitive and computational modeling pp
  385--403

\bibitem[{Vilain et~al(1995)Vilain, Burger, Aberdeen, Connolly, and
  Hirschman}]{vilain1995model}
Vilain M, Burger J, Aberdeen J, Connolly D, Hirschman L (1995) A
  model-theoretic coreference scoring scheme. In: Proceedings of the 6th
  conference on Message understanding, Association for Computational
  Linguistics, pp 45--52

\bibitem[{Walker(1989)}]{walker1989evaluating}
Walker MA (1989) Evaluating discourse processing algorithms. In: Proceedings of
  the 27th annual meeting on Association for Computational Linguistics,
  Association for Computational Linguistics, pp 251--261

\bibitem[{Walker et~al(1998)Walker, Joshi, and Prince}]{walker1998centering}
Walker MA, Joshi AK, Prince EF (1998) Centering theory in discourse. Oxford
  University Press

\bibitem[{Watson-Gegeo(1981)}]{watson1981pear}
Watson-Gegeo KA (1981) The pear stories: Cognitive, cultural, and linguistic
  aspects of narrative production

\bibitem[{Werlen and Popescu-Belis(2017)}]{werlen2017using}
Werlen LM, Popescu-Belis A (2017) Using coreference links to improve
  spanish-to-english machine translation. In: Proceedings of the 2nd Workshop
  on Coreference Resolution Beyond OntoNotes (CORBON 2017), pp 30--40

\bibitem[{Weston et~al(2015)Weston, Bordes, Chopra, Rush, van Merri{\"e}nboer,
  Joulin, and Mikolov}]{weston2015towards}
Weston J, Bordes A, Chopra S, Rush AM, van Merri{\"e}nboer B, Joulin A, Mikolov
  T (2015) Towards ai-complete question answering: A set of prerequisite toy
  tasks. arXiv preprint arXiv:150205698

\bibitem[{Wiseman et~al(2015)Wiseman, Rush, Shieber, and
  Weston}]{wiseman2015learning}
Wiseman S, Rush AM, Shieber S, Weston J (2015) Learning anaphoricity and
  antecedent ranking features for coreference resolution. In: Proceedings of
  the 53rd Annual Meeting of the Association for Computational Linguistics and
  the 7th International Joint Conference on Natural Language Processing (Volume
  1: Long Papers), vol~1, pp 1416--1426

\bibitem[{Wiseman et~al(2016)Wiseman, Rush, and Shieber}]{wiseman2016learning}
Wiseman S, Rush AM, Shieber SM (2016) Learning global features for coreference
  resolution. arXiv preprint arXiv:160403035

\bibitem[{Witte and Bergler(2003)}]{witte2003fuzzy}
Witte R, Bergler S (2003) Fuzzy coreference resolution for summarization. In:
  Proceedings of 2003 International Symposium on Reference Resolution and Its
  Applications to Question Answering and Summarization (ARQAS), pp 43--50

\bibitem[{Yang et~al(2003)Yang, Zhou, Su, and Tan}]{yang2003coreference}
Yang X, Zhou G, Su J, Tan CL (2003) Coreference resolution using competition
  learning approach. In: Proceedings of the 41st Annual Meeting on Association
  for Computational Linguistics-Volume 1, Association for Computational
  Linguistics, pp 176--183

\bibitem[{Yang et~al(2004)Yang, Zhou, Su, and Tan}]{yang2004improving}
Yang X, Zhou G, Su J, Tan CL (2004) Improving noun phrase coreference
  resolution by matching strings. In: International Conference on Natural
  Language Processing, Springer, pp 22--31

\bibitem[{Yang et~al(2008)Yang, Su, and Tan}]{yang2008twin}
Yang X, Su J, Tan CL (2008) A twin-candidate model for learning-based anaphora
  resolution. Computational Linguistics 34(3):327--356

\bibitem[{Young et~al(2018{\natexlab{a}})Young, Cambria, Chaturvedi, Zhou,
  Biswas, and Huang}]{youaug}
Young T, Cambria E, Chaturvedi I, Zhou H, Biswas S, Huang M
  (2018{\natexlab{a}}) Augmenting end-to-end dialogue systems with commonsense
  knowledge. In: {AAAI}, pp 4970--4977

\bibitem[{Young et~al(2018{\natexlab{b}})Young, Hazarika, Poria, and
  Cambria}]{yourec}
Young T, Hazarika D, Poria S, Cambria E (2018{\natexlab{b}}) Recent trends in
  deep learning based natural language processing. IEEE Computational
  Intelligence Magazine 13(3)

\bibitem[{Zeldes(2017)}]{zeldes2017gum}
Zeldes A (2017) The gum corpus: creating multilayer resources in the classroom.
  Language Resources and Evaluation 51(3):581--612

\bibitem[{Zeldes and Zhang(2016)}]{zeldes2016annotation}
Zeldes A, Zhang S (2016) When annotation schemes change rules help: A
  configurable approach to coreference resolution beyond ontonotes. In:
  Proceedings of the Workshop on Coreference Resolution Beyond OntoNotes
  (CORBON 2016), pp 92--101

\bibitem[{Zhuang et~al(2006)Zhuang, Jing, and Zhu}]{zhuang2006movie}
Zhuang L, Jing F, Zhu XY (2006) Movie review mining and summarization. In:
  Proceedings of the 15th ACM international conference on Information and
  knowledge management, ACM, pp 43--50

\end{thebibliography}
\end{document}